\documentclass[10pt]{article} 
\usepackage[preprint]{tmlr}

\usepackage{amsmath,amsfonts,bm}

\def\eqref#1{equation~\ref{#1}}

\def\1{\bm{1}}

\DeclareMathAlphabet{\mathsfit}{\encodingdefault}{\sfdefault}{m}{sl}
\SetMathAlphabet{\mathsfit}{bold}{\encodingdefault}{\sfdefault}{bx}{n}

\usepackage{hyperref}
\usepackage{url}

\usepackage{amsthm}
\usepackage{booktabs}
\usepackage{algorithm}
\usepackage{algpseudocode}
\usepackage{multirow}
\usepackage{graphicx} 
\usepackage{amsmath}
\usepackage{amssymb}
\usepackage{bm}
\usepackage{wrapfig}
\newtheorem{theorem}{Theorem} 
\newtheorem{proposition}{Proposition}
\newcommand{\AS}{\mathrm{AS}}
\newcommand{\SDS}{\mathrm{SDS}}
\newcommand{\MAR}{\mathrm{MAR}}

\title{Post-Hoc Critic Comparison and Adaptive Routing via Split-Point Analysis for Offline Reinforcement Learning}

 \author{\name Zhizhong Zhao \email zhizhong.zhao@manchester.ac.uk\\
       \addr Department of Computer Science\\
       University of Manchester
       \AND
       \name Ke Chen \email ke.chen@manchester.ac.uk\\
       \addr Department of Computer Science\\
       University of Manchester}

\begin{document}

\maketitle

\begin{abstract}
Critics are central to policy extraction in offline reinforcement learning, especially in IQL-style methods, whose performance is sensitive to the implicit maximization behavior induced by the critic. In this work, we investigate the post-hoc diagnosis of a bank of pretrained critics. Building on the generalized IQL perspective that policy extraction is governed by the critic-induced advantage field, we introduce split-point analysis (SPA) to characterize critic residual structures in terms of selectivity and structural credibility. Using a lightweight neural module built on top of each critic, SPA provides diagnostic signals for critic comparison, ranking, and adaptive state-dependent routing. We further develop a multi-critic aggregation strategy based on this proxy, which integrates seamlessly with existing IQL-style pipelines. Experiments across diverse offline RL tasks show that the proposed proxy provides an effective critic-comparison signal and the aggregation improves downstream policy performance for IQL-style instantiations under a unified protocol. 
\end{abstract}

\section{Introduction}
Offline reinforcement learning (offline RL) learns policies from static datasets without further environment interaction~\citep{levine2020offline}, making it attractive for applications where online exploration is costly or unsafe. A central challenge is distributional shift between the offline dataset and the learned policy: under limited data coverage, policy improvement may induce out-of-distribution (OOD) actions whose values are difficult to estimate reliably~\citep{fujimoto2019off}. Existing approaches address this challenge through policy constraints that restrict deviation from the behavior distribution~\citep{kumar2019stabilizing, fujimoto2021minimalist, wu2022supported, ran2023policy}, value regularization that suppresses overestimated OOD actions~\citep{kumar2020cql, yu2021combo, yang2022rorl}, and in-sample learning that avoids querying unseen actions during critic training~\citep{kostrikov2022iql, garg2022extreme, xu2023offline, hansenestruch2023idql}.


Among these approaches, Implicit Q-Learning (IQL)~\citep{kostrikov2022iql} has become a representative in-sample method because it trains its critic independently of policy extraction and avoids explicit maximization over unseen actions, while achieving strong empirical performance~\citep{PrudencioMC24}. However, IQL introduces a different source of sensitivity: its value function is learned through expectile regression, where the expectile level $\tau$ controls the degree of implicit maximization toward upper-value actions. Since $\tau$ also influences Q-function learning through bootstrapped value targets, different expectile regimes can yield critics with substantially different downstream policy performance. Consequently, comparing pretrained critics for downstream policy extraction remains challenging without online evaluation. Existing approaches primarily address expectile sensitivity during critic training, for example by adapting the expectile objective~\citep{upadhyay2026uniq, han2025projection} or improving critic robustness through uncertainty-aware mechanisms~\citep{yang2023towards, chen2025active}. In contrast, post-hoc comparison and aggregation of independently pretrained IQL-style critics remain underexplored. This raises a practical question: \emph{How can pretrained IQL-style critics be directly compared and aggregated for downstream policy extraction when online evaluation is unavailable?}

\begin{figure}
    \centering
    \includegraphics[width=0.9\linewidth]{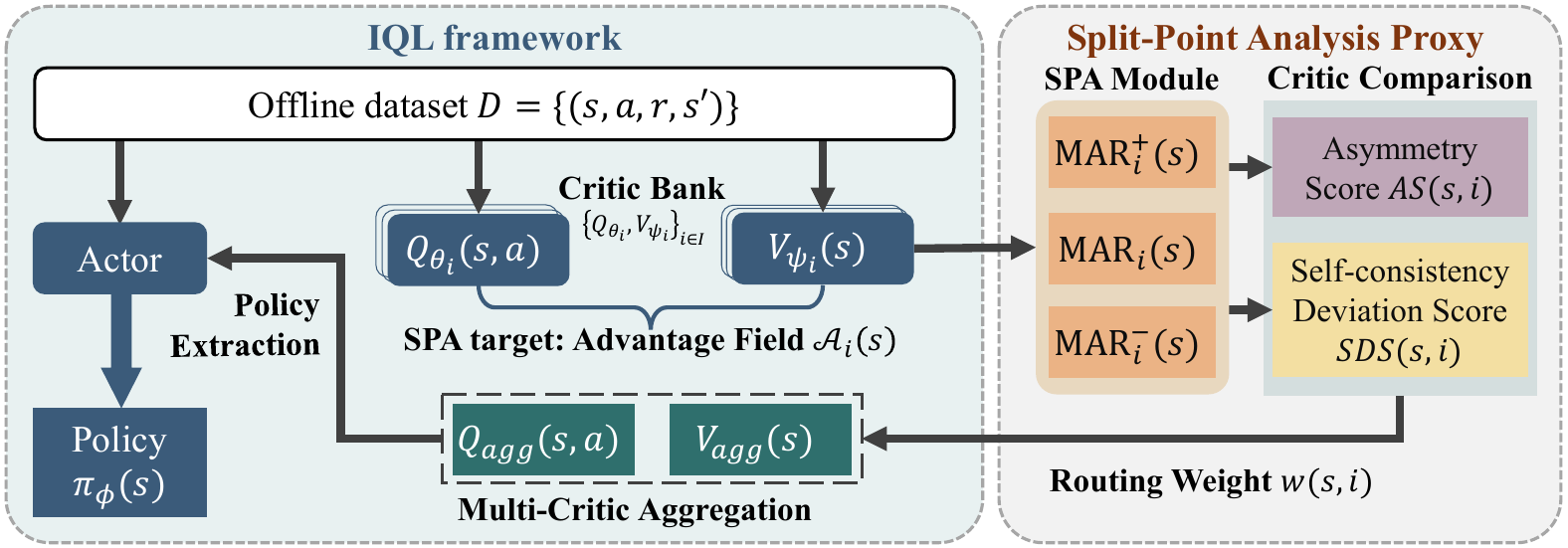}
    \caption{Overview of the proposed post-hoc critic diagnostic and routing framework.}
    \label{fig:selection_overview}
\end{figure}

In this paper, we propose a post-hoc framework for diagnosing pretrained IQL-style critics across different expectile regimes and aggregating their complementary information through a mixture-of-experts (MoE) mechanism, aiming to improve policy performance while reducing sensitivity to critic configuration. As illustrated in Figure~\ref{fig:selection_overview}, we apply \emph{split-point analysis} (SPA)~\citep{zhao2025post} to critic-induced advantage fields and derive two quantities: an \emph{asymmetry score} (AS), which measures split-point selectivity, and a \emph{self-consistency deviation score} (SDS), which assesses the structural credibility of SPA. Together, AS and SDS provide a credibility-aware diagnostic proxy that guides state-dependent aggregation of pretrained critics. Building on this proxy, we introduce a MoE-based critic aggregation framework for IQL-style offline RL, which can be integrated into existing pipelines through a lightweight SPA module once a critic bank is available. Rather than guaranteeing critic correctness in training, our method uses interpretable structural diagnostics to aggregate complementary critic information for downstream policy extraction.

The contributions are threefold:
(i) we introduce a theoretically motivated and credibility-aware diagnostic proxy for post-hoc comparison of IQL-style critics;
(ii) we propose a multi-critic aggregation framework for IQL-style offline RL pipelines;
and (iii) we provide empirical evidence that the SPA proxy enables effective critic comparison and that multi-critic aggregation improves downstream policy performance without task-specific hyperparameter selection.

\section{Related Work}
\textbf{IQL-style offline RL.}
IQL-style methods avoid explicitly querying unseen actions during critic learning by replacing action-space maximization with in-sample upper-value estimation through expectile regression~\citep{kostrikov2022iql, xu2023offline, hansenestruch2023idql}. The expectile objective controls the degree of implicit maximization over in-dataset actions, while the separation between critic learning and policy extraction enables integration with expressive policy classes such as diffusion or flow-based models~\citep{hansenestruch2023idql, fqlpark2025}. Our work focuses specifically on this critic structure and studies how pretrained IQL-style critics obtained under different expectile regimes can be compared and combined under fully offline conditions.

\textbf{Offline critic selection and adaptation.}
To diagnose models and select configurations under fully offline conditions, prior work evaluates policies or value functions using offline criteria based on logged data~\citep{lepaine2020hparamselect}, value-function comparison~\citep{xie2021batch, zhang2021bvftPolicySelection}, or Bellman-error-based measures~\citep{lee2022oracle, zitovsky2023revisiting}.
Within IQL-style learning, critic configuration is closely related to sensitivity to the expectile objective.
PIQL~\citep{han2025projection} replaces the fixed expectile coefficient with an adaptive parameter determined by projecting the behavior policy onto the learned policy, while UNIQ~\citep{upadhyay2026uniq} uses conformally calibrated uncertainty from a multi-expectile value ensemble to adapt the expectile coefficient on a state-wise basis.
While offline selection methods typically select a single candidate globally, these adaptive IQL methods adjust the expectile objective during critic training.
In contrast, we perform post-hoc diagnosis of independently pretrained IQL-style critics across different expectile regimes, without retraining critics or coupling their diagnosis to actor optimization.

\textbf{Ensemble-based offline RL.}
Ensembles have been widely used in offline RL to estimate epistemic uncertainty~\citep{lockwood2022review} and mitigate value estimation errors through pessimistic aggregation~\citep{an2021uncertainty, Bai0YDG0W22, ghasemipour2022so}.
Beyond pessimistic aggregation, GRAM-QT~\citep{shin2027gramqt} introduces a MoE framework that routes Q-function experts according to operating regimes.
Within IQL-style learning, ACTIVE~\citep{chen2025active} employs an in-sample value ensemble to stabilize critic training, while UNIQ~\citep{upadhyay2026uniq} uses conformally calibrated ensemble disagreement to adapt the expectile regression.
However, MoE-based aggregation of independently pretrained IQL-style critics across different expectile regimes for downstream policy extraction remains unexplored.
We address this gap through multi-critic aggregation guided by SPA-based structural diagnostics.

\textbf{Post-hoc split-point analysis.}
Our method builds on post-hoc split-point analysis (SPA) and self-consistency verification developed for supervised uncertainty quantification~\citep{zhao2025post}.
We extend SPA to critic-induced advantage distributions in IQL-style offline RL, using split-point asymmetry to identify critics with more informative advantage structures and self-consistency deviations to assess the structural credibility of the resulting diagnostics.

\section{Preliminaries}
\subsection{Offline Reinforcement Learning}
We consider a Markov decision process (MDP) defined by a tuple $\mathcal{M} = (\mathcal{S}, \mathcal{A}, p, r, \gamma, d_0)$, where $\mathcal{S}$ and $\mathcal{A}$ denote the state and action spaces, $p(s'|s,a)$ the transition dynamics, $r(s,a)$ the reward function, $\gamma \in (0,1)$ the discount factor, and $d_0$ the initial state distribution. The action-value function (Q-function) of a policy $\pi$ is defined as
\(
Q^\pi(s_t, a_t) = \mathbb{E}_{\pi} \left[ \sum_{j=0}^{\infty} \gamma^j r(s_{t+j}, a_{t+j}) \right],
\)
and the corresponding value function is
\(
V^\pi(s) = \mathbb{E}_{a \sim \pi(\cdot|s)} \big[ Q^\pi(s,a) \big].
\)
The objective is to learn a policy that maximizes the expected discounted return
\(
J(\pi) = \mathbb{E}_{s \sim d_0} \big[ V^\pi(s) \big].
\)
In offline reinforcement learning, the agent learns solely from a fixed dataset $\mathcal{D} = \{(s, a, r, s')\}$ collected by a behavior policy $\mu(a|s)$, without further interaction with the environment.

\subsection{Implicit Q-Learning}
Implicit Q-Learning (IQL) \citep{kostrikov2022iql} learns a value function $V_\psi$ that approximates an upper expectile of in-sample Q-values, and uses it to guide Q-function learning without evaluating OOD actions. Specifically, the value function is trained via expectile regression:
\begin{equation}
\mathcal{L}_V(\psi)
=
\mathbb{E}_{(s,a)\sim\mathcal{D}}
\left[
\mathcal{L}_\tau^2
\bigl(Q_{\theta}(s,a)-V_\psi(s)\bigr)
\right],
\end{equation}
where \(\mathcal{L}_\tau^2(u)=|\tau-\mathbf{1}(u<0)|u^2\). The expectile level \(\tau\) controls the degree of implicit maximization, and the value of \(\tau\) that is most beneficial for policy extraction often depends on the task and dataset. As a result, tuning \(\tau\) becomes a key challenge under fully offline conditions.

During training, an online Q-network $Q_{\tilde{\theta}}$ is updated by bootstrapping from $V_\psi$:
\begin{equation}
\mathcal{L}_Q(\tilde{\theta})
=
\mathbb{E}_{(s,a,s')\sim\mathcal{D}}
\left[
\bigl(
r(s,a)+\gamma V_\psi(s')-Q_{\tilde{\theta}}(s,a)
\bigr)^2
\right].
\end{equation}
A target Q-network $Q_{\theta}$ is maintained as a slowly updated copy of $Q_{\tilde{\theta}}$ and provides stable Q-estimates for value-function regression. After training, $Q_\theta$ is retained as the learned Q-function and forms the critic pair $(Q_\theta,V_\psi)$ with $V_\psi$. This in-sample learning scheme avoids querying $\max_a Q(s',a)$ over unseen actions.

The policy $\pi_\phi$ is extracted via advantage-weighted regression (AWR) \citep{peng2019advantage} by maximizing
\begin{equation}
\mathcal{J}_{\pi}(\phi)
=
\mathbb{E}_{(s,a)\sim\mathcal{D}}
\left[
\exp\left(
t\left(Q_\theta(s,a)-V_\psi(s)\right)
\right)
\log \pi_\phi(a|s)
\right].
\end{equation}
Here, inverse temperature parameter $t$ balances exploitation of the learned critic with behavior cloning.

\subsection{Advantage Field under Generalized IQL}
\label{sec:advantage_field}
To analyze how critic learning shapes downstream policy extraction, we adopt the generalized IQL perspective \citep{hansenestruch2023idql}, which extends the expectile loss with a differentiable convex loss $f$ satisfying $f'(0)=0$:
$
V^*(s)
=
\arg\min_{V(s)}
\mathbb{E}_{a\sim\mu(a|s)}
\left[
f\big(Q(s,a)-V(s)\big)
\right].
$
Under this formulation, the solution induces an implicit actor $\pi_{\mathrm{imp}}(a|s)\propto w(s,a)\mu(a|s)$, where
\begin{equation}
w(s,a)
=
\frac{\left|f'\big(Q(s,a)-V^*(s)\big)\right|}
{\left|Q(s,a)-V^*(s)\right|}.
\label{eq:weight_function}
\end{equation}
 For the expectile loss function $f=\mathcal{L}_\tau^2$, the weights are proportional to 
\(w_\tau(s,a)=|\tau-\mathbf{1}\{Q(s,a)<V_\tau(s)\}|\).

This view highlights that policy extraction is governed by a reweighting of the behavior policy, where the weights depend on both the loss function $f$ (controlling the upper expectile) and the accuracy of the learned value function $V^*(s)$ under function approximation. Both effects are mediated through the residual
\(
Q(s,a)-V(s).
\)
Motivated by this, we study the residual structure induced by the learned critic pair $(Q_\theta,V_\psi)$ and define the advantage field:
\begin{equation}
\mathcal{A}(s)
:=
\left\{Q(s,a)-V(s)\mid a\sim\mu(\cdot|s)\right\}.
\label{eq:advantage_field}
\end{equation}
By analyzing each critic-induced advantage field, we can assess critic selectivity and reliability in a post-hoc manner, without online evaluation or coupling to actor learning.

\section{Split-Point Analysis of the Critic-Induced Advantage Field}
\label{sec:split_point_analysis}
Since a learned critic pair $(Q_\theta, V_\psi)$ induces an advantage distribution that governs policy extraction in IQL-style methods, we approximate this distribution from finite offline data using critic-induced residual samples on the training set, and apply split-point analysis (SPA) to the resulting empirical residual structure.

For each transition $(s,a,r,s') \in \mathcal D$, we define the residual
\(
e_{\theta,\psi}(s,a)=Q_\theta(s,a)-V_\psi(s),
\)
which serves as an empirical sample of the advantage field $\mathcal{A}(s)$. Collecting all such state-residual pairs yields
\(
\mathcal{E}=\{(s,e)\},
\)
which we partition according to the sign of the residual:
\(
\mathcal{E}^+=\{(s,e)\mid e \ge 0\},
\quad
\mathcal{E}^-=\{(s,e)\mid e < 0\}.
\)
Here, $\mathcal{E}^+$ corresponds to nonnegative-advantage actions with $Q_\theta(s,a)\ge V_\psi(s)$, whereas $\mathcal{E}^-$ corresponds to negative-advantage actions.
Based on this partition, we define the total, upper-side, and lower-side mean absolute residuals (MARs):
\begin{equation}
\MAR(s) = \mathbb{E}[|e| \mid s], \quad
\MAR^{+}(s) = \mathbb{E}[|e| \mid s, e \ge 0], \quad
\MAR^{-}(s) = \mathbb{E}[|e| \mid s, e < 0].
\label{eq:reg-MARs}
\end{equation}
These quantities summarize the local residual structure around the critic-induced split-point \(V_\psi(s)\). In practice, MAR estimates are obtained empirically. Since exact repetitions of a continuous state are rare in offline control datasets, we learn representation-conditioned surrogates with a neural SPA module, using the critic feature \(h_\psi(s)\) rather than the raw state \(s\). This approximation relies on a local smoothness assumption: the conditional residual distribution does not change abruptly with \(s\)~\citep{chung2021beyond}, together with the regularity induced by neural function approximation.

On top of this decomposition, we develop a post-hoc diagnostic with two components: \emph{asymmetry characterization}, which describes how selectively the split-point partitions the advantage distribution, and \emph{self-consistency verification}, which evaluates whether this split-point analysis remains structurally credible.

\subsection{Asymmetry Characterization}
\label{sec:asymmetry_characterization}
We posit that a selective critic should induce a split-point in a stable upper region of the state-conditional in-sample value distribution, thereby yielding an advantage field that better separates actions useful for policy extraction. To characterize this selectivity, we define the state-wise asymmetry score (AS), which measures how the critic-induced split-point $V_\psi(s)$ asymmetrically partitions the advantage field:
\begin{equation}
\label{eq:asymmetry_score}
\AS(s)
=
\frac{\MAR^{-}(s)}{\MAR(s)}.
\end{equation}
$\AS(s)$ measures the relative lower-side residual magnitude compared with the total residual scale. In the case of expectile regression, this quantity admits a closed-form characterization along the expectile path:
\begin{proposition}[Closed form of asymmetry score]
\label{prop:as_closed_form}
Fix a state $s$, and let $Y=Q(s,A)$ with $A\sim\mu(\cdot\mid s)$. Assume that $Y$ is non-degenerate with $\mathbb{E}[Y^2]<\infty$. For $\tau\in(0,1)$, let $t_\tau$ denote the $\tau$-expectile of $Y$, and assume $\mathbb{P}(Y<t_\tau\mid s)>0$. Then
\begin{equation}
\AS(s,\tau)
=
\frac{\mathbb{E}[t_\tau-Y\mid s, Y<t_\tau]}
{\mathbb{E}[|Y-t_\tau|\mid s]}
=
\frac{\tau}{\mathbb{P}(Y<t_\tau\mid s)}.
\end{equation}
\end{proposition}
Proposition~\ref{prop:as_closed_form} shows that $\AS(s,\tau)$ connects the expectile level $\tau$ to the cumulative probability mass below its corresponding split-point $t_\tau$ in the critic-induced residual structure. It therefore characterizes the residual asymmetry along the expectile path of a fixed conditional distribution, without implying a performance ordering among independently trained critics.

\begin{wrapfigure}{r}{0.55\textwidth}
    \centering
    \vspace{-10pt}
    \includegraphics[width=\linewidth]{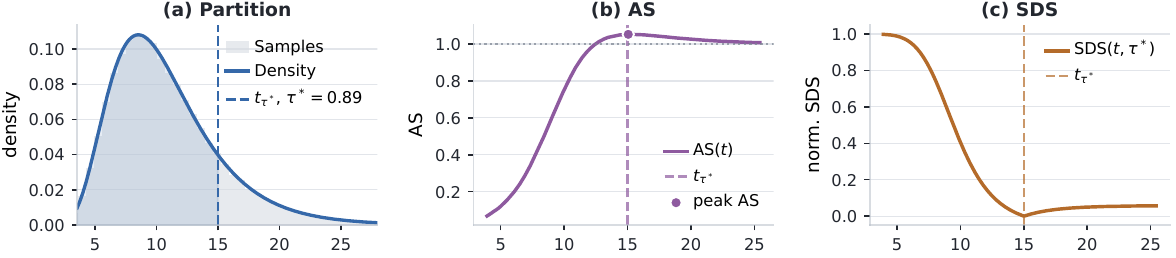}
    \caption{
Illustration of AS and SDS under varying split-point locations in a right-skewed advantage distribution: (a) partition of the distribution at the split-point $t_{\tau^*}$; (b) AS variation with the split-point, attaining an interior maximum at $t_{\tau^*}$; and (c) SDS variation with the split-point, attaining zero at $t_{\tau^*}$.
}
    \label{fig:main_as_sds_illustration}
    \vspace{-10pt}
\end{wrapfigure}
For continuous distributions with a non-dominating upper tail, $\AS(s,\tau)$ can exceed $1$ as $t_\tau$ moves toward the upper-advantage region and exhibit an interior peak before approaching $1$ in the far-right tail, as illustrated in Figure~\ref{fig:main_as_sds_illustration}. This behavior provides a useful diagnostic signal: under a reliable critic-induced residual structure, AS identifies split-points that are selective toward upper-advantage regions while still being tied to the cumulative mass below $t_\tau$. We use it as a post-hoc ranking signal for critic comparison. The proof and further discussion are provided in Appendix~\ref{app:as}.

\subsection{Self-Consistency Verification}
\label{sec:self_consistency_verification}
We next examine whether the residual structure induced by \((Q_\theta,V_\psi)\) is consistent with the structural relation implied by expectile regression. Building on SPC-UQ~\citep{zhao2025post}, we extend its self-consistency principle to the expectile-regression setting and derive a state-wise self-consistency deviation score (SDS) for assessing the credibility of the MAR-based asymmetry characterization.

\begin{theorem}[Expectile self-consistency]
\label{th:expectile_sc}
Let $Y$ be a non-degenerate random variable with $\mathbb{E}[Y^2] < \infty$. Let \(t_\tau \in \mathbb{R}\) denote the unique
\(\tau\)-expectile of \(Y\) for \(\tau\in(0,1)\). Assume $\mathbb{P}(Y<t_\tau)>0$, $\mathbb{P}(Y>t_\tau)>0$, and define the total, upper-side, and lower-side mean absolute deviations by
$\mathrm{MAD}(t_\tau)=\mathbb{E}[|Y-t_\tau|]$, 
$\mathrm{MAD}^{+}(t_\tau)=\mathbb{E}[Y-t_\tau\mid Y\ge t_\tau]$, 
$\mathrm{MAD}^{-}(t_\tau)=\mathbb{E}[t_\tau-Y\mid Y<t_\tau]$.
Then
\begin{equation}
\label{eq:self_consistency}
\frac{1}{\mathrm{MAD}(t_\tau)}
=
\frac{\tau}{\mathrm{MAD}^{-}(t_\tau)}
+
\frac{1-\tau}{\mathrm{MAD}^{+}(t_\tau)}.
\end{equation}
\end{theorem}
In practice, we use the estimated state-conditional residual summaries $\MAR(s)$, $\MAR^{+}(s)$ and $\MAR^{-}(s)$.
Rearranging Eq.~\ref{eq:self_consistency}, we define the normalized self-consistency deviation score:
\begin{equation}
\label{eq:sds_score}
\SDS(s,\tau)
=
\frac{
\left|\tau \MAR(s)(\MAR^{+}(s)-\MAR^{-}(s))-\MAR^{-}(s)(\MAR^{+}(s)-\MAR(s))\right|
}{
\max\left\{\left|\tau \MAR(s)(\MAR^{+}(s)-\MAR^{-}(s))\right|+\left|\MAR^{-}(s)(\MAR^{+}(s)-\MAR(s))\right|,
\epsilon
\right\}
}.
\end{equation}
Here, $\epsilon>0$ is a small constant for numerical stability. SDS assesses whether the independently estimated MARs satisfy the expectile self-consistency relation. A large SDS indicates inaccurate MAR estimation or misalignment of the split-point with the nominal expectile, as illustrated in Figure~\ref{fig:main_as_sds_illustration}, whereas a small SDS suggests that the estimated advantage field approximately satisfies the self-consistency relation. SDS therefore serves as a structural credibility check for SPA-based critic ranking.

However, when $\MAR^{+}(s)=\MAR^{-}(s)=\MAR(s)$, SDS vanishes even if the split-point is misaligned with the nominal $\tau$-expectile. This degeneracy necessarily implies $\AS=1$. Since our aggregation prioritizes critics with high AS, we explicitly restrict eligibility to $\AS>1$, thereby excluding this structural false-zero case. Further analysis and numerical considerations are provided in Appendix~\ref{app:sds}.

Together, $\AS$ and $\SDS$ provide complementary diagnostic signals for pretrained critic comparison, which we use for adaptive routing and aggregation in the next section.

\section{Post-Hoc Critic Routing and Aggregation}
\label{sec:spa_routing}

\subsection{Split-Point Analysis Module}
\label{subsect:SPA-module}
As illustrated in Figure~\ref{fig:selection_overview}, we introduce a lightweight SPA module to estimate the three MAR quantities defined in Eq.~\ref{eq:reg-MARs}. Operating in a post-hoc manner\footnote{\label{fn:model}If no pretrained base model is available, our framework can also be trained from scratch by first learning the base critic and subsequently fitting the SPA module in a stagewise manner, or by optimizing both components jointly.}, the SPA module is built on top of the learned value function
\(
V_{\psi^*}(s)= g\bigl(h(s;\psi^*)\bigr),
\)
where \(h(s;\psi^*)\) denotes the penultimate-layer feature representation and \(g(\cdot)\) is the output mapping. We use \(h(s;\psi^*)\) as a critic-induced state representation, which allows the SPA module to remain lightweight while reusing features already learned for value estimation.

The SPA module is implemented as a multi-head multilayer perceptron (MLP) regressor with parameters \(\omega\):
\[
p_\omega\colon h(\mathcal{S})\to \mathcal{Z}_{\MAR}
\times \mathcal{Z}_{\MAR^+} \times \mathcal{Z}_{\MAR^-},
\]
where
\(
\mathcal{Z}_{\MAR},~
\mathcal{Z}_{\MAR^+},~
\mathcal{Z}_{\MAR^-}\subset\mathbb{R}
\)
denote the ranges of the total, upper-side, and lower-side MARs, respectively.
Its output is:
\[
p_\omega(h)=\bigl(z(h;\omega), z^{+}(h;\omega), z^{-}(h;\omega)\bigr),
\]
where each output head uses a softplus activation to enforce non-negativity of the estimated MAR quantities.

Let $e_j = Q_\theta(s_j,a_j)-V_\psi(s_j)$ and $h_j=h_\psi(s_j)$.
We train $p_\omega(h)$ by regressing its three heads onto critic-induced residual magnitudes:
\begin{equation}
\label{eq:spa_head_training}
\omega^*
=
\arg\min_{\omega}
\Biggl\{
\frac{1}{|\mathcal{D}|}
\sum_{j\in\mathcal{D}}
\bigl(z(h_j;\omega)-|e_j|\bigr)^2
+
\frac{1}{|\mathcal{D}^{+}|}
\sum_{j\in\mathcal{D}^{+}}
\bigl(z^{+}(h_j;\omega)-|e_j|\bigr)^2
+
\frac{1}{|\mathcal{D}^{-}|}
\sum_{j\in\mathcal{D}^{-}}
\bigl(z^{-}(h_j;\omega)-|e_j|\bigr)^2
\Biggr\},
\end{equation}
where $\mathcal{D}^{+}=\{j:e_j\ge 0\}$ and $\mathcal{D}^{-}=\{j:e_j<0\}$.

\subsection{Multi-Critic Aggregation}
\label{sec:multi_critic_aggregation}

Given a set of candidate critics $\mathcal I$, we define each candidate critic $i\in\mathcal I$ as the pair $(Q_{\theta_i}, V_{\psi_i})$ associated with expectile level $\tau_i$. Denoting its state-dependent diagnostic scores from split-point analysis by $\AS(s,i)$ and $\SDS(s,i)$, we define the \emph{Credited AS} routing signal as
\begin{equation}
\label{eq:credited-AS}
u(s,i)
=
\AS(s,i)\cdot
\exp\!\left(-\SDS(s,i)/\beta\right).
\end{equation}
Here, $c(s,i)=\exp\!\left(-\SDS(s,i)/\beta\right)$ serves as a credibility gate that penalizes critics whose residual structures exhibit greater deviations from the expectile-induced self-consistency relation. Since SDS is normalized in Eq.~\ref{eq:sds_score}, it measures relative structural deviation and reduces sensitivity to differences in critic value scales. Consequently, $\beta>0$ primarily controls the credibility penalty strength rather than compensating for scale differences, and we fix $\beta=1$ throughout our experiments without task-specific tuning.

To prioritize critics with stronger split-point selectivity, we restrict aggregation to the eligible critic set $\mathcal I_s=\{i\in\mathcal I:\AS(s,i)>1\}$. This criterion also excludes the false-zero degeneracy when the eligible set is nonempty. For $\mathcal I_s\neq\emptyset$, we standardize the Credited AS values over eligible critics and obtain the routing weights through a softmax:
\begin{equation}
\label{eq:routing_weight}
w(s,i)=
\frac{\exp\!\left(\tilde{u}(s,i)/\alpha\right)}
{\sum_{k\in\mathcal I_s}\exp\!\left(\tilde{u}(s,k)/\alpha\right)},
\qquad i\in\mathcal I_s.
\end{equation}
Here, $\tilde{u}(s,i)=\frac{u(s,i)-\mu_s(u)}{\sigma_s(u)+\epsilon}$, where $\mu_s(u)$ and $\sigma_s(u)$ are the mean and standard deviation
over $\mathcal I_s$, and $\epsilon>0$ ensures numerical stability. Standardization enhances discrimination based on relative diagnostic scores. Critics outside $\mathcal I_s$ receive zero weight; if $\mathcal I_s=\emptyset$, we select the critic with the highest AS. The temperature $\alpha>0$ controls weight concentration, with $\alpha\to0$ favoring the highest Credited AS.

The aggregated critic is then defined as
\begin{equation}
\label{eq:agg_critic}
Q_{\mathrm{agg}}(s,a)
=
\sum_{i\in\mathcal I}
w(s,i)\, Q_{\theta_i}(s,a), \qquad
V_{\mathrm{agg}}(s)
=
\sum_{i\in\mathcal I}
w(s,i)\, V_{\psi_i}(s).
\end{equation}
The resulting pair $(Q_{\mathrm{agg}},V_{\mathrm{agg}})$ is used for downstream policy extraction. This yields a mixture-of-experts-style aggregation that combines pretrained critics across expectile regimes through state-dependent weights derived from split-point analysis. Since the aggregated critic is used only for policy extraction rather than as a bootstrapping target for value learning, Bellman consistency of the aggregated pair is not required.

\subsection{Downstream Policy Extraction}
\label{sec:policy_extraction}

The aggregated $Q_{\mathrm{agg}}$ and $V_{\mathrm{agg}}$ can be used as plug-in critic signals for downstream policy extraction in IQL-style offline RL pipelines. For Gaussian-policy-based methods, we replace the original critic signals in the advantage-weighted regression (AWR) objective with the aggregated critic:
\begin{equation}
\label{eq:policy-awr}
\phi^*
=
\arg\max_{\phi}\,
\mathbb{E}_{(s,a)\sim\mathcal{D}}
\Bigl[\exp\!\left(t\bigl(Q_{\mathrm{agg}}(s,a)-V_{\mathrm{agg}}(s)\bigr)\right) \cdot \log \pi_\phi(a\mid s)
\Bigr].
\end{equation}
For diffusion-policy-based methods, a diffusion model approximating $\mu(\cdot \mid s)$ generates candidate actions, and the policy selects the action with the highest aggregated value:
\begin{equation}
\label{eq:policy-select}
\pi(s)
=
\arg\max_{a_i \sim \mu(\cdot \mid s),\, i=1,\ldots,N}
Q_{\mathrm{agg}}(s,a_i).
\end{equation}
Overall, our method serves as a post-hoc critic routing and aggregation module that can be attached to existing IQL-style offline RL pipelines without modifying critic training or policy extraction.
We instantiate it on Gaussian-policy-based IQL~\citep{kostrikov2022iql} and diffusion-policy-based IDQL~\citep{hansenestruch2023idql}, denoted as \emph{IQL-Agg} and \emph{IDQL-Agg}.
In each case, the original single-expectile critic used for policy extraction is replaced by the aggregated critic.
Algorithm~\ref{alg:spa-main} summarizes the pipeline, with full details and complexity analysis in Appendix~\ref{app:algorithm}.

\begin{algorithm}
\caption{SPA-based critic aggregation}
\label{alg:spa-main}
\begin{algorithmic}[1]
\Require Offline dataset $\mathcal D$, pretrained critics $\{(Q_{\theta_i},V_{\psi_i})\}_{i\in\mathcal I}$ with expectile levels $\{\tau_i\}_{i\in\mathcal I}$
\Ensure Extracted policy $\pi_\phi$

\State Train an SPA module $p_{\omega_i}$ for each candidate critic $i\in\mathcal I$ to estimate $(\MAR_i,\MAR_i^{+},\MAR_i^{-})$

\For{each state-action pair $(s,a)$}
     \For{each candidate critic $i\in\mathcal I$}
        \State Compute $\AS(s,i)$ via Eq.~\ref{eq:asymmetry_score} and $\SDS(s,i)$ via Eq.~\ref{eq:sds_score} from the SPA module output $p_{\omega_i}(h_i)$
    \EndFor

    \State Set eligible critics $\mathcal I_s\gets\{i\in\mathcal I:\AS(s,i)>1\}$

    \State Compute routing weights $w(s,i)$ via Eq.~\ref{eq:credited-AS} and Eq.~\ref{eq:routing_weight}

    \State Aggregate $Q_{\mathrm{agg}}$ and $V_{\mathrm{agg}}$ via Eq.~\ref{eq:agg_critic}
\EndFor

\State Extract policy $\pi_\phi$ using Eq.~\ref{eq:policy-awr} (IQL-Agg) or Eq.~\ref{eq:policy-select} (IDQL-Agg)
\State \Return $\pi_\phi$
\end{algorithmic}
\end{algorithm}

\section{Experiments}
\paragraph{Benchmarks.}
All experiments are conducted on D4RL~\citep{fu2020d4rl}, covering three task domains: MuJoCo for dense-reward locomotion, AntMaze for sparse-reward navigation, and Kitchen for multi-goal manipulation. The corruption analysis follows RIQL~\citep{yang2023towards}.
For critic comparison and structural credibility analysis, each dataset is randomly split into $80\%$ training and $20\%$ held-out transitions for diagnostic evaluation. For downstream policy evaluation with adaptive critic routing, we use the full offline dataset following the standard IQL protocol~\citep{kostrikov2022iql}.

\paragraph{Unified hyperparameter protocol.}
To ensure controlled comparisons without task-specific tuning based on environment returns, we adopt a unified protocol with a fixed hyperparameter configuration within each domain. For IQL-style critics, the expectile is set to $\tau=0.7$ for MuJoCo and Kitchen, and $\tau=0.9$ for AntMaze. The AWR inverse temperature is $t=3.0$ for MuJoCo, $t=10.0$ for AntMaze, and $t=0.5$ for Kitchen. All remaining backbone hyperparameters, including the learning rate, are kept fixed within each domain. The SPA-module training and aggregation settings are shared across all domains, using the critic bank $\mathcal{T}=\{0.6,0.7,0.8,0.9\}$ with $\alpha=\beta=1.0$.

\paragraph{Baselines and implementation.}
We compare against representative offline RL methods, including TD3+BC~\citep{fujimoto2021minimalist}, CQL~\citep{kumar2020cql}, IQL~\citep{kostrikov2022iql}, ACTIVE~\citep{chen2025active}, UNIQ~\citep{upadhyay2026uniq}, and IDQL~\citep{hansenestruch2023idql}. We evaluate all reproduced methods under the unified protocol. The SPA module is a lightweight MLP with a single hidden layer that takes features from the pretrained value network as input. Appendix~\ref{app:experimental_details} provides further details on network architectures, optimization, computational resources, and implementation. Our source code is available at \url{https://github.com/zzz0527/SPA-Critic-Aggregation}.

\subsection{Post-Hoc Critic Diagnosis}
\label{sec:critic_diagnosis}

\subsubsection{Critic Ranking and Selection}
\label{sec:ranking_and_selection}
This experiment evaluates whether the proposed SPA-based diagnostics provide informative post-hoc signals for comparing IQL-style critics when online policy evaluation is unavailable. For each task, we construct a critic bank with five expectile levels, $\tau\in\{0.5,0.6,0.7,0.8,0.9\}$, and compare the ranking induced by each signal with the downstream performance of the corresponding extracted policies. We consider three reference signals: Random, which uses a random critic ordering; Max-$\tau$, which ranks critics directly by their expectile levels; and the expectile regression Loss. These are compared with the proposed AS, SDS, and Credited AS. For the four data-dependent signals (Loss, AS, SDS, and Credited AS), we average their sample-wise values over the held-out validation set to obtain a single score for each critic. We evaluate all signals using Spearman rank correlation~\citep{spearman1961proof} and Top-$K$ selection accuracy.

\begin{figure}[h]
    \centering
    \includegraphics[width=1\linewidth]{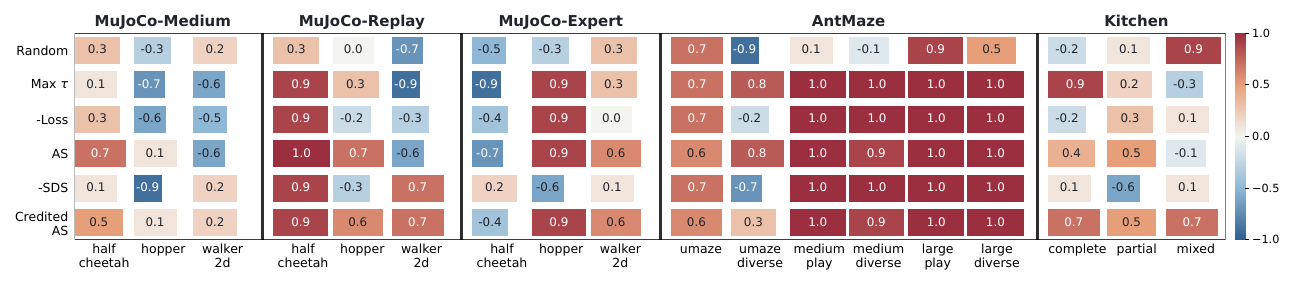}
    \caption{Correlation between critic comparison signals and policy performance. Correlations for Loss and SDS are computed using $-\text{Loss}$ and $-\SDS$ so that higher values consistently indicate better critics.}
    \label{fig:correlation}
\end{figure}

As shown in Figure~\ref{fig:correlation}, Random, Max-$\tau$, and expectile regression Loss achieve average correlations of $0.07$, $0.32$, and $0.27$, respectively. In particular, Max-$\tau$ exhibits substantial domain dependence, ranging from $0.92$ on AntMaze to $-0.40$ on MuJoCo-Medium, while Loss is positively correlated with performance on only 10 of the 18 tasks. These results indicate that neither training loss nor a fixed preference for larger expectiles provides a reliable post-hoc criterion for critic comparison.

In contrast, AS achieves an average correlation of $0.46$, with positive correlations on 14 tasks. Incorporating SDS further improves these results to $0.60$ and 17 tasks, respectively. These findings suggest that AS provides a more informative post-hoc critic comparison signal than training loss or nominal expectile ordering, while SDS further improves its empirical alignment with downstream performance. The only negative case for Credited AS is halfcheetah-medium-expert, where performance varies little across expectile levels. Full task-wise results are provided in Appendix~\ref{app:correlation}.

\begin{wraptable}{r}{0.4\columnwidth}
\vspace{-4mm}
\centering
\caption{Top-$K$ critic selection hit rate (\%) across D4RL tasks.}
\label{tab:topk_selection}
\begin{tabular}{lccc}
\toprule
Metric & Top-1 & Top-2 & Top-3 \\
\midrule
Random       & 22.22 & 38.89 & 66.67 \\
Max-$\tau$   & 27.78 & 66.67 & 88.89 \\
Loss         & 27.78 & 55.56 & 77.78 \\
SDS          & 27.78 & 38.89 & 66.67 \\
AS           & 61.11 & 72.22 & 77.78 \\
Credited AS  & \textbf{77.78} & \textbf{88.89} & \textbf{94.44} \\
\bottomrule
\end{tabular}
\vspace{-4mm}
\end{wraptable}

Table~\ref{tab:topk_selection} further assesses the practical utility of each diagnostic for critic selection, measuring whether the Top-$K$ ranked critics include the best-performing critic. For Top-1 selection, the two naive references Max-$\tau$ and Loss achieve only $27.78\%$, close to Random ($22.22\%$), whereas AS improves the hit rate to $61.11\%$. Incorporating SDS further improves the Top-$K$ hit rates, while SDS alone remains ineffective for selection. These results suggest that AS provides the primary ranking signal, with SDS serving as a complementary structural credibility measure.

\subsubsection{Offline Diagnostic Validation}
\label{sec:offline_diagnostic_validation}
We further investigate the statistical properties of SPA diagnostics independently of downstream policy performance. First, a synthetic expectile-regression example illustrates the residual asymmetry associated with different expectile levels and the behavior of AS and SDS under distributional extrapolation. Second, we assess the estimation quality of the auxiliary SPA module through a held-out data calibration test. Detailed analyses are provided in Appendices~\ref{app:offline_diagnostic_validation1} and~\ref{app:offline_diagnostic_validation2}.

\begin{wrapfigure}{r}{0.5\textwidth}
    \centering
    \vspace{-10pt}
    \includegraphics[width=\linewidth]{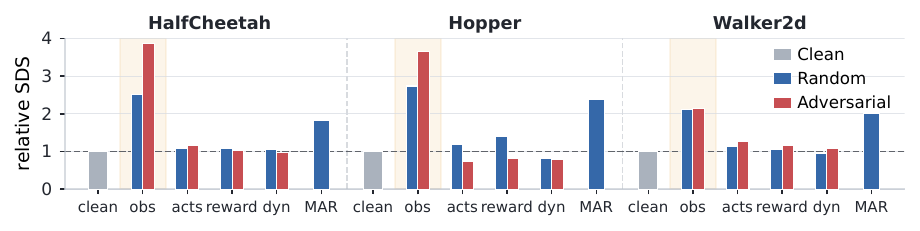}
    \caption{Sensitivity of SDS to data and MAR-output perturbations relative to the uncorrupted baseline.}
    \label{fig:sds_corruption}
    \vspace{-10pt}
\end{wrapfigure}

We next examine the sensitivity of SDS to controlled perturbations of transition data and independently injected noise in the estimated MAR statistics. As shown in Figure~\ref{fig:sds_corruption}, observation corruption and MAR-output perturbations produce pronounced changes in SDS. In contrast, perturbations to actions, rewards, or next states do not directly affect SPA predictions, as the SPA module takes only the state representation as input. These experiments characterize the sensitivity of SDS to post-training perturbations rather than its robustness to corruption during critic training.

Finally, we examine whether improvements in expectile self-consistency are reflected in the AS and SDS estimates through value-function refitting with a fixed Q-function. Results are reported in Appendix~\ref{app:offline_diagnostic_validation3}.

\subsection{Adaptive Critic Routing and Aggregation}
\label{sec:routing-baseline}

\subsubsection{Policy Performance and Baseline Comparison}
This experiment evaluates whether the proposed multi-critic aggregation improves downstream policy performance. Table~\ref{tab:d4rl_result} compares IQL-Agg and IDQL-Agg with representative offline RL baselines across MuJoCo, AntMaze, and Kitchen. To isolate the contribution of critic aggregation from the policy backbone, we focus on matched comparisons between IQL and IQL-Agg, and between IDQL and IDQL-Agg. The values in parentheses indicate signed performance changes relative to the corresponding backbone.

The matched-backbone comparisons show consistent improvements at the domain level. IQL-Agg improves over IQL by $+39.74$, $+32.20$, and $+10.66$ on MuJoCo, AntMaze, and Kitchen, respectively. Similarly, IDQL-Agg improves over IDQL by $+2.68$, $+11.05$, and $+4.53$. These gains are obtained without modifying the underlying critic-training objective or policy parameterization, indicating that post-hoc aggregation can effectively exploit complementary information from critics trained under different expectile regimes.

Compared with existing adaptive critic learning methods, UNIQ employs conformally calibrated uncertainty estimates for adaptive expectile selection and improves upon IQL on MuJoCo. However, its reproduced performance degrades on AntMaze, which was not evaluated in the original study. This highlights the need to further investigate Bellman bootstrapping stability in adaptive critic learning. In contrast, our post-hoc aggregation operates directly on pretrained critics without altering their training process. ACTIVE achieves performance comparable to IQL-Agg through an in-sample value ensemble but retains a fixed expectile parameter, leaving expectile sensitivity unaddressed. Given the challenges of reproducing baseline methods under a unified protocol, we provide additional comparisons with originally reported results in Appendix~\ref{app:additional_baseline}.

Among all methods under the unified protocol, IQL-style methods remain highly competitive, with IDQL providing the strongest single-backbone baseline by combining in-sample critic learning with a more expressive diffusion-based policy extractor. Applying SPA-based aggregation on top of IDQL further improves all three domain totals, with IDQL-Agg achieving the strongest overall performance. This result highlights the generality of the proposed multi-critic aggregation framework, which accommodates both Gaussian and diffusion-based policies without modifying their parameterizations. Although ACTIVE and UNIQ also pursue ensemble-based or adaptive-parameter improvements to standard IQL, their compatibility with more expressive policy extractors has not been systematically established.

\begin{table*}[t]
\centering
\caption{
Average normalized scores on the D4RL tasks.
Results marked with $^{\dagger}$ are taken from prior work; all other methods are reproduced under our unified protocol and reported as mean $\pm$ standard deviation over five seeds.
For IQL-Agg and IDQL-Agg, the values in parentheses denote the signed score change relative to their matched backbones.
The highest mean score in each row is shown in bold.
}
\label{tab:d4rl_result}
\resizebox{\textwidth}{!}{
\begin{tabular}{lrrrrrrrr}
\hline
\multicolumn{1}{c}{Dataset}
& \multicolumn{1}{c}{TD3+BC$^{\dagger}$}
& \multicolumn{1}{c}{CQL$^{\dagger}$}
& \multicolumn{1}{c}{IQL}
& \multicolumn{1}{c}{UNIQ}
& \multicolumn{1}{c}{ACTIVE}
& \multicolumn{1}{c}{IDQL}
& \multicolumn{1}{c}{IQL-Agg}
& \multicolumn{1}{c}{IDQL-Agg} \\
\hline

halfcheetah-medium
& 48.3 & 44.0 & 47.39 $\pm$ 0.17 & 46.01 $\pm$ 0.19 & 48.94 $\pm$ 0.29 & 51.09 $\pm$ 0.19 & 47.87 $\pm$ 0.22 $(+0.48)$ & \textbf{52.01} $\pm$ 0.19 $(+0.92)$ \\

hopper-medium
& 59.3 & 58.5 & 71.33 $\pm$ 1.95 & 72.10 $\pm$ 4.47 & 71.98 $\pm$ 3.15 & 60.48 $\pm$ 8.64 & \textbf{72.44} $\pm$ 4.62 $(+1.11)$ & 70.52 $\pm$ 6.63 $(+10.04)$ \\

walker2d-medium
& 83.7 & 72.5 & 82.34 $\pm$ 2.10 & 81.25 $\pm$ 1.75 & 82.93 $\pm$ 2.25 & \textbf{85.83} $\pm$ 1.28 & 82.19 $\pm$ 2.35 $(-0.15)$ & 78.55 $\pm$ 1.09 $(-7.28)$ \\

halfcheetah-medium-replay
& 44.6 & 45.5 & 43.57 $\pm$ 1.28 & 43.06 $\pm$ 0.79 & 44.85 $\pm$ 0.23 & 45.60 $\pm$ 0.19 & 44.71 $\pm$ 0.52 $(+1.14)$ & \textbf{45.96} $\pm$ 0.15 $(+0.36)$ \\

hopper-medium-replay
& 60.9 & 95.0 & 80.63 $\pm$ 15.22 & 89.46 $\pm$ 4.26 & 94.23 $\pm$ 5.63 & 96.24 $\pm$ 1.24 & \textbf{98.51} $\pm$ 2.39 $(+17.88)$ & 96.24 $\pm$ 4.32 $(0.00)$ \\

walker2d-medium-replay
& 81.8 & 77.2 & 73.39 $\pm$ 1.82 & 74.38 $\pm$ 4.29 & 77.94 $\pm$ 7.71 & 84.05 $\pm$ 5.68 & 76.87 $\pm$ 6.58 $(+3.48)$ & \textbf{85.03} $\pm$ 4.08 $(+0.98)$ \\

halfcheetah-medium-expert
& 90.7 & 91.6 & 89.06 $\pm$ 3.40 & 86.68 $\pm$ 1.91 & 92.84 $\pm$ 0.70 & 94.17 $\pm$ 0.15 & 94.94 $\pm$ 1.69 $(+5.88)$ & \textbf{95.52} $\pm$ 1.82 $(+1.35)$ \\

hopper-medium-expert
& 98.0 & 105.4 & 95.34 $\pm$ 17.88 & 96.43 $\pm$ 9.23 & \textbf{109.26} $\pm$ 1.71 & 108.81 $\pm$ 3.18 & 104.77 $\pm$ 6.89 $(+9.43)$ & 104.84 $\pm$ 3.28 $(-3.97)$ \\

walker2d-medium-expert
& 110.1 & 108.8 & 110.05 $\pm$ 1.31 & 110.29 $\pm$ 0.64 & 111.74 $\pm$ 0.42 & 112.39 $\pm$ 0.15 & 110.54 $\pm$ 0.25 $(+0.49)$ & \textbf{112.67} $\pm$ 0.03 $(+0.28)$ \\

\hline
MuJoCo-Total
& 677.4 & 698.5 & 693.10 & 699.66 & 734.70 & 738.66 & 732.84 $(+39.74)$ & \textbf{741.34} $(+2.68)$ \\
\hline

antmaze-umaze
& 78.6 & 74.0 & 86.30 $\pm$ 3.10 & 82.83 $\pm$ 1.95 & 84.74 $\pm$ 3.53 & 94.25 $\pm$ 2.09 & 88.76 $\pm$ 3.41 $(+2.46)$ & \textbf{96.90} $\pm$ 2.08 $(+2.65)$ \\

antmaze-umaze-diverse
& 71.4 & 84.0 & 65.98 $\pm$ 3.42 & 62.09 $\pm$ 4.24 & 73.18 $\pm$ 3.86 & 80.65 $\pm$ 12.12 & 73.23 $\pm$ 5.72 $(+7.25)$ & \textbf{84.81} $\pm$ 16.05 $(+4.16)$ \\

antmaze-medium-play
& 10.6 & 61.2 & 70.70 $\pm$ 1.51 & 3.14 $\pm$ 1.20 & 79.09 $\pm$ 2.18 & \textbf{82.87} $\pm$ 5.01 & 73.54 $\pm$ 3.46 $(+2.84)$ & 81.63 $\pm$ 2.06 $(-1.24)$ \\

antmaze-medium-diverse
& 3.0 & 53.7 & 67.72 $\pm$ 3.62 & 3.06 $\pm$ 0.21 & 77.35 $\pm$ 4.05 & 81.32 $\pm$ 6.85 & 76.59 $\pm$ 4.05 $(+8.87)$ & \textbf{84.21} $\pm$ 0.74 $(+2.89)$ \\

antmaze-large-play
& 0.2 & 15.8 & 43.43 $\pm$ 2.74 & 0.22 $\pm$ 0.17 & 45.25 $\pm$ 5.78 & 68.11 $\pm$ 2.56 & 45.56 $\pm$ 4.27 $(+2.13)$ & \textbf{71.19} $\pm$ 2.47 $(+3.08)$ \\

antmaze-large-diverse
& 0.0 & 14.9 & 37.75 $\pm$ 3.96 & 0.14 $\pm$ 0.24 & 42.57 $\pm$ 0.92 & \textbf{73.85} $\pm$ 5.09 & 46.39 $\pm$ 4.10 $(+8.64)$ & 73.36 $\pm$ 8.82 $(-0.49)$ \\

\hline
AntMaze-Total
& 163.8 & 303.6 & 371.88 & 151.47 & 402.19 & 481.05 & 404.08 $(+32.20)$ & \textbf{492.10} $(+11.05)$ \\
\hline

kitchen-complete
& 61.5 & 43.8 & 68.51 $\pm$ 5.38 & 59.10 $\pm$ 6.96 & 69.23 $\pm$ 2.89 & 70.23 $\pm$ 2.76 & 63.94 $\pm$ 6.74 $(-4.57)$ & \textbf{70.83} $\pm$ 2.69 $(+0.60)$ \\

kitchen-partial
& 52.8 & 49.8 & 55.56 $\pm$ 8.85 & 39.10 $\pm$ 7.01 & \textbf{72.27} $\pm$ 3.35 & 64.16 $\pm$ 5.64 & 58.72 $\pm$ 5.30 $(+3.16)$ & 67.35 $\pm$ 5.44 $(+3.19)$ \\

kitchen-mixed
& 60.8 & 51.0 & 49.47 $\pm$ 4.67 & 48.14 $\pm$ 4.47 & 41.99 $\pm$ 10.65 & 63.03 $\pm$ 1.75 & 61.54 $\pm$ 3.45 $(+12.07)$ & \textbf{63.77} $\pm$ 1.42 $(+0.74)$ \\

\hline
Kitchen-Total
& 175.1 & 144.6 & 173.54 & 146.34 & 183.48 & 197.42 & 184.20 $(+10.66)$ & \textbf{201.95} $(+4.53)$ \\
\hline

\end{tabular}
}
\end{table*}

\subsubsection{Ablation and Diagnostic Studies}
\label{sec:further_analysis}

To further validate the proposed critic routing and aggregation mechanism, we conduct ablation and diagnostic studies based on IQL-Agg to answer questions regarding the design of diagnostic proxies, the benefits of multi-critic aggregation, the behavior of adaptive critic routing, and sensitivity to configuration choices.

\paragraph{How do diagnostic design choices affect critic routing?}
We begin by examining how the design of SPA-based diagnostic signals affects critic routing. To isolate the contribution of SDS-based credibility weighting, we introduce AS-Only, which performs routing using AS alone. As shown in Table~\ref{tab:ablation}, IQL-Agg achieves higher domain-total scores than AS-Only across all three domains, with a particularly pronounced gain on AntMaze ($404.08$ vs.\ $358.08$). By incorporating SDS as a structural credibility check, Credited AS penalizes the diagnostic scores of critics with larger self-consistency discrepancies, thereby improving aggregation performance of IQL-Agg. We further examine two complementary design choices: AS-based eligibility filtering, which prioritizes critics with stronger residual selectivity, and SDS normalization, which aims to improve the comparability of credibility scores across critics. Detailed results are provided in Appendix~\ref{app:diagnostic_design}.

\paragraph{What are the benefits of SPA-based critic routing?}
Having examined the construction of the diagnostic signal, we turn to the benefits of using it for state-dependent critic aggregation. We first consider whether adaptive aggregation can outperform selecting a single favorable critic for each task. To this end, we compare IQL-Agg against Oracle Single, which selects the highest evaluation return among the individual single-critic policies for each task and random seed. Oracle Single serves as an idealized single-critic performance reference rather than a fully offline selection strategy. As shown in Table~\ref{tab:ablation}, IQL-Agg achieves higher domain-total scores than Oracle Single on MuJoCo and AntMaze, while obtaining comparable performance on Kitchen (184.20 vs.\ 184.73). These results suggest that SPA-based aggregation can achieve competitive performance against hindsight single-critic selection, without requiring environment returns.

However, outperforming a single-critic oracle does not establish whether the observed gains arise from multi-critic aggregation, SPA-based weighting, or state-dependent routing. To disentangle these factors, we consider five alternative aggregation strategies using the same pretrained critic bank. Shuffled Agg randomly permutes SPA outputs and their associated expectile levels across critics, disrupting the correspondence between critics and their diagnostic information. Uniform Agg assigns equal weights to all critics, serving as a non-selective aggregation baseline. Static Agg replaces the state-dependent routing weights with their averages over the offline dataset, preserving the average SPA-based weight allocation while removing state dependence. Performance-Weighted Aggregation (PW-Agg) constructs fixed global weights from the evaluation returns of individual critic-induced policies, providing an idealized performance-informed reference for critic aggregation. Finally, Expectile-Balance Mixture of Experts (EB-MoE) employs a learnable state-dependent routing mechanism based on estimated deviations from the expectile-regression optimality condition, providing a generic MoE baseline. Further implementation details are provided in Appendix~\ref{app:aggregation-baseline}.

As shown in Table~\ref{tab:ablation}, Uniform Agg and Shuffled Agg exhibit performance degradation when diagnostic-guided weighting is either removed or misaligned, highlighting the importance of diagnostic-guided selective weighting. Compared with the two fixed-weight strategies, Static Agg and PW-Agg, IQL-Agg achieves higher domain-total mean scores across all three domains. Although fixed weighting performs better on some individual tasks, potentially reflecting consistent critic preferences across states, the observed mean differences suggest task-dependent benefits from state-dependent weighting over the evaluated fixed-weight strategies, particularly on AntMaze and Kitchen. Furthermore, EB-MoE achieves lower domain-total scores despite employing learnable state-dependent routing, suggesting that a generic MoE routing mechanism based solely on expectile-regression consistency may be insufficient for effective critic aggregation. Together, these results indicate that the benefits of IQL-Agg arise not merely from combining multiple critics, but from informative SPA-based diagnostics and their selective, state-dependent use in critic aggregation.


\begin{table*}
\centering
\caption{
Ablation results comparing critic routing and aggregation strategies.
Scores are reported as mean $\pm$ standard deviation over five seeds.
The highest mean score in each row is shown in bold.
}
\label{tab:ablation}
\resizebox{\textwidth}{!}{
\begin{tabular}{lrrrrrrrr}
\hline
\multicolumn{1}{c}{Dataset}
& \multicolumn{1}{c}{IQL-Agg}
& \multicolumn{1}{c}{AS-Only}
& \multicolumn{1}{c}{Oracle Single}
& \multicolumn{1}{c}{Shuffled Agg}
& \multicolumn{1}{c}{Uniform Agg}
& \multicolumn{1}{c}{Static Agg}
& \multicolumn{1}{c}{PW-Agg}
& \multicolumn{1}{c}{EB-MoE} \\
\hline

halfcheetah-medium
& 47.87 $\pm$ 0.22
& 47.86 $\pm$ 0.06
& 47.39 $\pm$ 0.17
& 47.76 $\pm$ 0.06
& 47.95 $\pm$ 0.13
& \textbf{48.00} $\pm$ 0.08
& 47.77 $\pm$ 0.08
& 47.89 $\pm$ 0.12 \\

hopper-medium
& 72.44 $\pm$ 4.62
& 73.01 $\pm$ 5.82
& 73.33 $\pm$ 1.95
& 58.83 $\pm$ 3.35
& 68.81 $\pm$ 4.93
& 71.21 $\pm$ 2.63
& \textbf{79.20} $\pm$ 4.07
& 71.01 $\pm$ 2.17 \\

walker2d-medium
& 82.19 $\pm$ 2.35
& \textbf{84.44} $\pm$ 2.94
& 82.89 $\pm$ 0.39
& 75.24 $\pm$ 3.09
& 82.17 $\pm$ 3.84
& 81.48 $\pm$ 2.06
& 83.84 $\pm$ 0.02
& 83.70 $\pm$ 1.36 \\

halfcheetah-medium-replay
& 44.71 $\pm$ 0.52
& 43.60 $\pm$ 0.89
& 44.69 $\pm$ 0.61
& 41.22 $\pm$ 0.63
& 44.21 $\pm$ 0.46
& 45.20 $\pm$ 0.51
& \textbf{45.93} $\pm$ 0.11
& 42.99 $\pm$ 1.35 \\

hopper-medium-replay
& \textbf{98.51} $\pm$ 2.39
& 98.25 $\pm$ 10.91
& 94.51 $\pm$ 11.14
& 87.03 $\pm$ 6.57
& 97.48 $\pm$ 9.53
& 95.76 $\pm$ 5.60
& 87.31 $\pm$ 4.34
& 95.92 $\pm$ 1.06 \\

walker2d-medium-replay
& 76.87 $\pm$ 6.58
& 70.40 $\pm$ 7.23
& 77.37 $\pm$ 5.21
& 54.75 $\pm$ 19.73
& 71.29 $\pm$ 12.90
& \textbf{79.11} $\pm$ 5.61
& 78.01 $\pm$ 6.17
& 73.75 $\pm$ 6.83 \\

halfcheetah-medium-expert
& 94.94 $\pm$ 1.69
& 87.41 $\pm$ 2.57
& 92.54 $\pm$ 1.44
& 77.70 $\pm$ 3.33
& 83.43 $\pm$ 7.21
& \textbf{96.29} $\pm$ 1.17
& 95.93 $\pm$ 1.64
& 85.32 $\pm$ 4.85 \\

hopper-medium-expert
& \textbf{104.77} $\pm$ 6.89
& 101.49 $\pm$ 6.61
& 102.77 $\pm$ 11.97
& 98.64 $\pm$ 16.57
& 100.02 $\pm$ 6.46
& 100.75 $\pm$ 2.85
& 92.53 $\pm$ 14.41
& 99.06 $\pm$ 13.28 \\

walker2d-medium-expert
& 110.54 $\pm$ 0.25
& 110.15 $\pm$ 1.48
& \textbf{111.49} $\pm$ 0.62
& 108.40 $\pm$ 2.28
& 109.48 $\pm$ 1.61
& 110.44 $\pm$ 0.06
& 110.81 $\pm$ 0.41
& 110.31 $\pm$ 0.38 \\

\hline
MuJoCo-Total
& \textbf{732.84}
& 716.61
& 726.98
& 649.57
& 704.82
& 728.24
& 721.33
& 709.93 \\
\hline

antmaze-umaze
& \textbf{88.76} $\pm$ 3.41
& 85.95 $\pm$ 7.70
& 86.87 $\pm$ 1.58
& 78.25 $\pm$ 5.55
& 86.98 $\pm$ 11.90
& 68.46 $\pm$ 3.12
& 86.87 $\pm$ 2.91
& 85.79 $\pm$ 6.49 \\

antmaze-umaze-diverse
& 73.23 $\pm$ 5.72
& 66.77 $\pm$ 10.52
& 75.98 $\pm$ 3.42
& 59.92 $\pm$ 3.10
& 71.39 $\pm$ 11.86
& 67.75 $\pm$ 3.42
& \textbf{81.79} $\pm$ 1.38
& 69.07 $\pm$ 7.73 \\

antmaze-medium-play
& 73.54 $\pm$ 3.46
& 62.71 $\pm$ 2.55
& 70.70 $\pm$ 1.51
& 13.08 $\pm$ 3.00
& 67.56 $\pm$ 6.08
& \textbf{74.22} $\pm$ 1.85
& 70.43 $\pm$ 0.54
& 69.44 $\pm$ 2.88 \\

antmaze-medium-diverse
& 76.59 $\pm$ 4.05
& 63.39 $\pm$ 1.68
& \textbf{77.72} $\pm$ 3.62
& 9.03 $\pm$ 3.80
& 68.13 $\pm$ 4.45
& 75.89 $\pm$ 3.80
& 76.86 $\pm$ 3.67
& 65.93 $\pm$ 1.89 \\

antmaze-large-play
& 45.56 $\pm$ 4.27
& 34.30 $\pm$ 2.55
& 44.43 $\pm$ 2.74
& 0.65 $\pm$ 0.69
& 36.43 $\pm$ 7.14
& \textbf{46.27} $\pm$ 2.00
& 45.14 $\pm$ 1.28
& 39.68 $\pm$ 5.88 \\

antmaze-large-diverse
& \textbf{46.39} $\pm$ 4.10
& 44.97 $\pm$ 3.81
& 43.75 $\pm$ 3.96
& 6.68 $\pm$ 2.24
& 38.58 $\pm$ 8.82
& 33.31 $\pm$ 1.14
& 37.34 $\pm$ 4.08
& 38.84 $\pm$ 3.03 \\

\hline
AntMaze-Total
& \textbf{404.08}
& 358.08
& 399.45
& 167.62
& 369.08
& 365.91
& 398.44
& 368.76 \\
\hline

kitchen-complete
& 63.94 $\pm$ 6.74
& 64.27 $\pm$ 5.76
& \textbf{70.64} $\pm$ 2.21
& 33.53 $\pm$ 29.68
& 64.92 $\pm$ 7.50
& 66.17 $\pm$ 4.12
& 60.47 $\pm$ 5.87
& 55.85 $\pm$ 17.84 \\

kitchen-partial
& 58.72 $\pm$ 5.30
& 58.30 $\pm$ 15.23
& \textbf{60.13} $\pm$ 4.77
& 49.82 $\pm$ 8.84
& 56.55 $\pm$ 12.18
& 47.70 $\pm$ 6.97
& 55.75 $\pm$ 4.64
& 57.44 $\pm$ 8.35 \\

kitchen-mixed
& 61.54 $\pm$ 3.45
& 52.44 $\pm$ 3.47
& 53.96 $\pm$ 4.38
& 57.06 $\pm$ 10.01
& 56.01 $\pm$ 3.81
& 56.75 $\pm$ 1.97
& \textbf{62.35} $\pm$ 0.53
& 60.36 $\pm$ 6.00 \\

\hline
Kitchen-Total
& 184.20
& 175.01
& \textbf{184.73}
& 140.41
& 177.48
& 170.62
& 178.56
& 173.65 \\
\hline

\end{tabular}
}
\end{table*}

\paragraph{State-dependent Routing Behavior and Dynamics.}
We further investigate how SPA-based routing utilizes the available critic bank by examining state-dependent routing distributions, trajectory-level routing preferences, and expectile dynamics during joint training, with detailed analyses provided in Appendix~\ref{app:aggregation_analysis}. The results show that critics trained with different expectile levels exhibit distinct action preferences, providing complementary information for aggregation. Consistent with this diversity, SPA assigns non-uniform, state-dependent routing weights without consistently favoring a single expectile regime. Moreover, during joint training, the average effective expectile remains relatively stable while the actor's evaluation return generally improves, although this does not establish the stability of individual state-dependent routing weights.

\paragraph{Training and routing configuration sensitivity.}
We finally examine the robustness of SPA-based routing to different training strategies and configuration choices. Routing configuration includes the critic-bank composition, routing temperature $\alpha$, and credibility scale $\beta$. Beyond the default bank, we evaluate a reduced bank $\{0.7,0.9\}$, an extended bank $\{0.5,0.6,0.7,0.8,0.9,0.95\}$, and a duplicated bank containing three critics for each default expectile. We also vary $\alpha$ and $\beta$ over $\{0.25,0.5,1.0,2.0,5.0\}$.

\begin{figure}[h]
    \centering
    \includegraphics[width=1\linewidth]{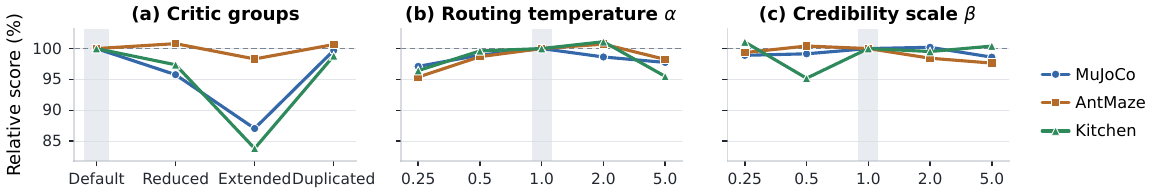}
    \caption{Sensitivity of IQL-Agg to critic-bank configuration, routing temperature $\alpha$, and credibility scale $\beta$. Scores are normalized to the default setting ($100\%$).}
    \label{fig:routing_sensitivity}
\end{figure}

Figure~\ref{fig:routing_sensitivity} summarizes the domain-level sensitivity results. Reducing the critic bank generally degrades performance by limiting the complementary information for aggregation, whereas duplicating critics within existing expectile regimes provides limited benefits without increasing expectile diversity. In contrast, extending the expectile range substantially reduces performance on MuJoCo and Kitchen, potentially reflecting less reliable critic estimates at extreme expectile levels. For $\alpha$ and $\beta$, moderate values within $[0.5,2.0]$ maintain competitive performance, with domain-level performance variations below $5\%$. Overall, the default configuration $\mathcal{T}=\{0.6,0.7,0.8,0.9\}$ and $\alpha=\beta=1.0$ offers a favorable performance trade-off without specific tuning.

In addition, as noted in Footnote~\ref{fn:model}, the SPA module can be trained jointly with the critics from initialization when pretrained critics are unavailable. Our comparison with the default post-hoc strategy demonstrates that joint optimization provides a practical alternative while eliminating the need for a separate SPA estimation stage. Full task-wise results and training comparisons are provided in Appendix~\ref{app:routing-config}.

\section{Conclusion and Discussion}
\label{sec:conclusion}
We proposed a post-hoc critic diagnosis and aggregation framework for IQL-style offline RL based on an auxiliary neural module. By applying split-point analysis to critic-induced residuals, our method derives selectivity and credibility signals for ranking and routing pretrained critics across expectile regimes. Experiments on D4RL show that this diagnostic proxy provides an effective critic-comparison signal and improves aggregate performance over fixed-expectile baselines in the main IQL and IDQL instantiations.

Beyond performance gains, the framework addresses a practical challenge in offline RL: identifying effective critics without online evaluation. While strong performance can often be obtained with extensive evaluation budgets~\citep{tarasov2023revisiting}, such protocols are difficult to deploy in fully offline settings. Since our method is attached after critic training without redesigning the policy optimization objective, it provides a practical post-hoc mechanism for critic selection and aggregation under limited tuning budgets.

Despite these encouraging findings, several limitations remain. 
First, SPA statistics are estimated by an auxiliary module rather than from the true advantage field, which may be inaccurate in sparse-reward or highly heterogeneous datasets. Although SDS assesses the credibility of the resulting split-point analysis, it does not calibrate the SPA module itself; more robust SPA module training is left for future work.

Second, the scoring mechanism is coupled to critic-induced advantage fields and self-bootstrapped critic learning. Thus, it should not be interpreted as a diagnostic of systematic value bias. In particular, a critic family may be internally self-consistent but globally biased, in which case SPA-based routing may still select a biased critic. Addressing such shifts requires diagnostics beyond self-consistency.

Finally, SPA performs critic comparison and routing at the state level, while relying on IQL's in-sample learning mechanism to mitigate OOD action evaluation. Extending SPA to action-level diagnostics and integrating it with more expressive policy learning approaches, such as flow matching~\citep{fqlpark2025} and value gradient flow~\citep{xu2026reinforcement}, are promising directions for future work.

\section*{AI-use Disclosure}
We used several AI models as assistive tools for language editing, technical critique, manuscript revision, and implementation and debugging support. The research ideas and questions originated with the authors; the methodology, theoretical framework and objectives, experimental design and execution, interpretation of results, and scientific conclusions were determined and validated by the authors. The authors critically evaluated the AI-generated suggestions and independently checked all incorporated technical content, code, references and reported experimental results. The authors take full responsibility for the accuracy, originality and integrity of this work.

\bibliography{main}
\bibliographystyle{tmlr}

\newpage
\appendix
\section{Theoretical Analysis of the Asymmetry Characterization}
\label{app:as}
In this appendix, we prove Proposition~\ref{prop:as_closed_form}, interpret the asymmetry score through the relationship between expectile levels and cumulative probability mass, and examine its non-monotonic behavior.

\subsection{Proof of Proposition~1}
Proposition~1 is stated for a fixed state $s$. Throughout the proof, we write $\AS(\tau)$ as shorthand for the asymmetry score $\AS(s,\tau)$ associated with the corresponding state-conditional distribution.

\begin{proof}
Fix a state $s$ and let $t_\tau$ be the $\tau$-expectile of the
state-conditional random variable $Y=Q(s,A)$ with
$A\sim\mu(\cdot\mid s)$. Here, mean absolute deviation (MAD)
denotes the population-level split-point quantity corresponding to
the empirical $\MAR$ notation used in the main text. For
notational simplicity, we suppress the explicit dependence on $s$.

Define
\[
p^-:=\mathbb{P}(Y<t_\tau),
\quad
p^+:=\mathbb{P}(Y\ge t_\tau),
\]
and
\[
\mathrm{MAD}(t_\tau)
:= \mathbb{E}\!\left[|Y-t_\tau|\right], \quad
\mathrm{MAD}^-(t_\tau)
:= \mathbb{E}\!\left[t_\tau-Y \mid Y<t_\tau\right], \quad
\mathrm{MAD}^+(t_\tau)
:= \mathbb{E}\!\left[Y-t_\tau \mid Y\ge t_\tau\right].
\]
Then the side-specific unconditional deviations can be written as
\begin{equation}
\begin{aligned}
\mathbb{E}\!\left[(t_\tau-Y)\mathbf{1}\{Y<t_\tau\}\right]
&= p^-\,\mathrm{MAD}^-(t_\tau),\\
\mathbb{E}\!\left[(Y-t_\tau)\mathbf{1}\{Y\ge t_\tau\}\right]
&= p^+\,\mathrm{MAD}^+(t_\tau).
\end{aligned}
\label{eq:indicator_mad_identities}
\end{equation}

Recall that the $\tau$-expectile $t_\tau$ minimizes
\[
\mathcal{L}(t)
:=
\mathbb{E}\!\left[
(1-\tau)(t-Y)^2\mathbf{1}\{Y<t\}
+
\tau(Y-t)^2\mathbf{1}\{Y>t\}
\right].
\]
Since the contribution at $Y=t_\tau$ is zero, using
$\mathbf{1}\{Y>t_\tau\}$ or $\mathbf{1}\{Y\ge t_\tau\}$ is equivalent.
The first-order optimality condition is therefore
\begin{equation}
(1-\tau)\,
\mathbb{E}\!\left[(t_\tau-Y)\mathbf{1}\{Y<t_\tau\}\right]
=
\tau\,
\mathbb{E}\!\left[(Y-t_\tau)\mathbf{1}\{Y\ge t_\tau\}\right].
\label{eq:expectile_foc_indicator}
\end{equation}
Combining Eq.~\ref{eq:indicator_mad_identities} and Eq.~\ref{eq:expectile_foc_indicator} gives
\begin{equation}
(1-\tau)\,p^-\,\mathrm{MAD}^-(t_\tau)
=
\tau\,p^+\,\mathrm{MAD}^+(t_\tau).
\label{eq:expectile_balance_mad}
\end{equation}

Next, the total mean absolute deviation around $t_\tau$ decomposes as
\begin{equation}
\begin{aligned}
\mathrm{MAD}(t_\tau)
&=
\mathbb{E}\!\left[(Y-t_\tau)\mathbf{1}\{Y\ge t_\tau\}\right]
+
\mathbb{E}\!\left[(t_\tau-Y)\mathbf{1}\{Y<t_\tau\}\right] \\
&=
p^+\,\mathrm{MAD}^+(t_\tau)
+
p^-\,\mathrm{MAD}^-(t_\tau).
\end{aligned}
\label{eq:mad_decomp}
\end{equation}
Using Eq.~\ref{eq:expectile_balance_mad}, we have
\begin{equation}
p^+\,\mathrm{MAD}^+(t_\tau)
=
\frac{1-\tau}{\tau}\,
p^-\,\mathrm{MAD}^-(t_\tau).
\label{eq:upper_to_lower_mad}
\end{equation}
Substituting Eq.~\ref{eq:upper_to_lower_mad} into Eq.~\ref{eq:mad_decomp} yields
\begin{equation}
\begin{aligned}
\mathrm{MAD}(t_\tau)
&=
p^-\,\mathrm{MAD}^-(t_\tau)
+
\frac{1-\tau}{\tau}\,
p^-\,\mathrm{MAD}^-(t_\tau) \\
&=
\frac{p^-\,\mathrm{MAD}^-(t_\tau)}{\tau}.
\end{aligned}
\label{eq:mad_as_lower}
\end{equation}

Therefore,
\[
\begin{aligned}
\AS(\tau)
&=
\frac{\mathrm{MAD}^-(t_\tau)}
{\mathrm{MAD}(t_\tau)}
=
\frac{\mathrm{MAD}^-(t_\tau)}
{\left(p^-\,\mathrm{MAD}^-(t_\tau)\right)/\tau} \\
&=
\frac{\tau}{p^-}
=
\frac{\tau}{\mathbb{P}(Y<t_\tau)}.
\end{aligned}
\]
This completes the proof.
\end{proof}

\subsection{Interpretation via Expectile-Path / Cumulative-Location Separation}
The closed form
\(
\AS(\tau)=\frac{\tau}{\mathbb{P}(Y<t_\tau)}
\)
connects the asymmetry score to two distinct notions of location associated with the same split-point. The first is the \emph{nominal expectile level} $\tau$, which indexes $t_\tau$ through asymmetric squared-loss weighting. The second is the \emph{cumulative location} $\mathbb{P}(Y<t_\tau)$, which records the probability mass below the split-point $t_\tau$.

Unlike quantiles, expectiles are not calibrated by cumulative probability mass. For a continuous random variable, the $\tau$-quantile $q_\tau$ satisfies $\mathbb{P}(Y<q_\tau)=\tau$, so its nominal index directly specifies its cumulative location. By contrast, the $\tau$-expectile $t_\tau$ is determined by an asymmetric squared-loss balance and generally satisfies $\mathbb{P}(Y<t_\tau)\neq\tau$. The ratio $\tau/\mathbb{P}(Y<t_\tau)$ thus characterizes the separation between the nominal expectile level and the cumulative mass induced by its corresponding split-point.

This separation arises because quantiles and expectiles are governed by different principles. 
The cumulative location $\mathbb{P}(Y<t_\tau)$ depends only on the probability mass below the split-point, whereas the expectile level $\tau$ also reflects the deviation magnitudes on both sides. 
Thus, the separation between $\tau$ and $\mathbb{P}(Y<t_\tau)$ captures how magnitude-sensitive asymmetry shifts the expectile away from a purely mass-based partition.

When $\AS(\tau)$ is close to $1$, the nominal expectile level and the induced cumulative location are comparable, indicating a limited separation between magnitude-based and mass-based characterizations of the split-point. A larger $\AS(\tau)$ indicates that the split-point is more strongly influenced by value magnitudes relative to probability mass alone. We interpret this asymmetrical separation as \emph{split-point selectivity}: high-value but low-density outcomes can still influence the split-point despite their limited probability mass. This makes AS a useful adaptive criterion for identifying informative split-points in the upper-value region, rather than relying solely on a predefined expectile level or pushing the split-point toward the extreme upper tail.

\subsection{Non-Monotonicity Analysis and Illustrative Examples}
\label{subsec:as_examples}

We further analyze the variation of AS along the expectile path of a fixed state-conditional value distribution. Recall that $\AS(\tau)=\mathrm{MAD}^{-}(t_\tau)/\mathrm{MAD}(t_\tau)$. Using the MAD decomposition in Eq.~\ref{eq:mad_decomp} and $p^-+p^+=1$, we obtain
\begin{equation}
\begin{aligned}
\AS(\tau)-1
&=\frac{\mathrm{MAD}^{-}(t_\tau)-\mathrm{MAD}(t_\tau)}{\mathrm{MAD}(t_\tau)}\\
&=\frac{\mathrm{MAD}^{-}(t_\tau)-p^-\mathrm{MAD}^{-}(t_\tau)-p^+\mathrm{MAD}^{+}(t_\tau)}{\mathrm{MAD}(t_\tau)}\\
&=\frac{p^+\left(\mathrm{MAD}^{-}(t_\tau)-\mathrm{MAD}^{+}(t_\tau)\right)}{\mathrm{MAD}(t_\tau)}.
\end{aligned}
\end{equation}
For $p^+>0$, this identity establishes that $\AS(\tau)>1$ if and only if $\mathrm{MAD}^{-}(t_\tau)>\mathrm{MAD}^{+}(t_\tau)$. Equivalently, by Proposition~\ref{prop:as_closed_form}, $\AS(\tau)>1$ corresponds to $\tau>\mathbb{P}(Y<t_\tau)$. Thus, AS exceeds $1$ when the average lower-side deviation dominates the upper-side deviation, indicating that the nominal expectile level lies above the cumulative location of its corresponding split-point.

We next examine the behavior of AS as $\tau\to1$. For a continuous state-conditional distribution with finite second moments, the expectile $t_\tau$ approaches the upper endpoint of the distribution as $\tau\to1$. Consequently, $\mathbb{P}(Y<t_\tau)\to1$, and Proposition~\ref{prop:as_closed_form} yields
\[
\lim_{\tau\to1}\AS(\tau)
=\lim_{\tau\to1}\frac{\tau}{\mathbb{P}(Y<t_\tau)}
=1.
\]
Therefore, although AS may exceed $1$ at intermediate expectile levels, it eventually returns toward $1$ as the split-point approaches the upper tail. This behavior allows AS to exhibit an interior maximum rather than monotonically favoring increasingly large expectile levels.

To illustrate this behavior analytically, consider $Y\sim\mathrm{Uniform}(0,1)$. For a split-point $t\in(0,1)$, the conditional mean absolute deviations are
\[
\mathrm{MAD}^{-}(t)=\frac{t}{2},\qquad
\mathrm{MAD}^{+}(t)=\frac{1-t}{2},
\]
and the total deviation is
\[
\mathrm{MAD}(t)=\frac{t^2+(1-t)^2}{2}.
\]
The expectile balance condition gives
\[
\tau=\frac{t_\tau^2}{t_\tau^2+(1-t_\tau)^2}.
\]
Substituting these expressions into the definition of AS yields
\[
\AS(\tau)
=\frac{t_\tau}{t_\tau^2+(1-t_\tau)^2}.
\]
Differentiating with respect to the split-point gives
\[
\frac{d\,\AS}{dt_\tau}
=\frac{1-2t_\tau^2}{\left[t_\tau^2+(1-t_\tau)^2\right]^2}.
\]
Since $t_\tau$ increases with $\tau$, AS increases for $t_\tau<1/\sqrt{2}$ and decreases for $t_\tau>1/\sqrt{2}$. It therefore attains an interior maximum at $t_{\tau^\star}=1/\sqrt{2}$, with $\AS(\tau^\star)=(1+\sqrt{2})/2>1$, before approaching $1$ in the upper tail. This example analytically demonstrates that AS can exhibit non-monotonic behavior, attaining an interior peak before reaching the largest expectile level.

\begin{figure}
    \centering
    \includegraphics[width=0.95\linewidth]{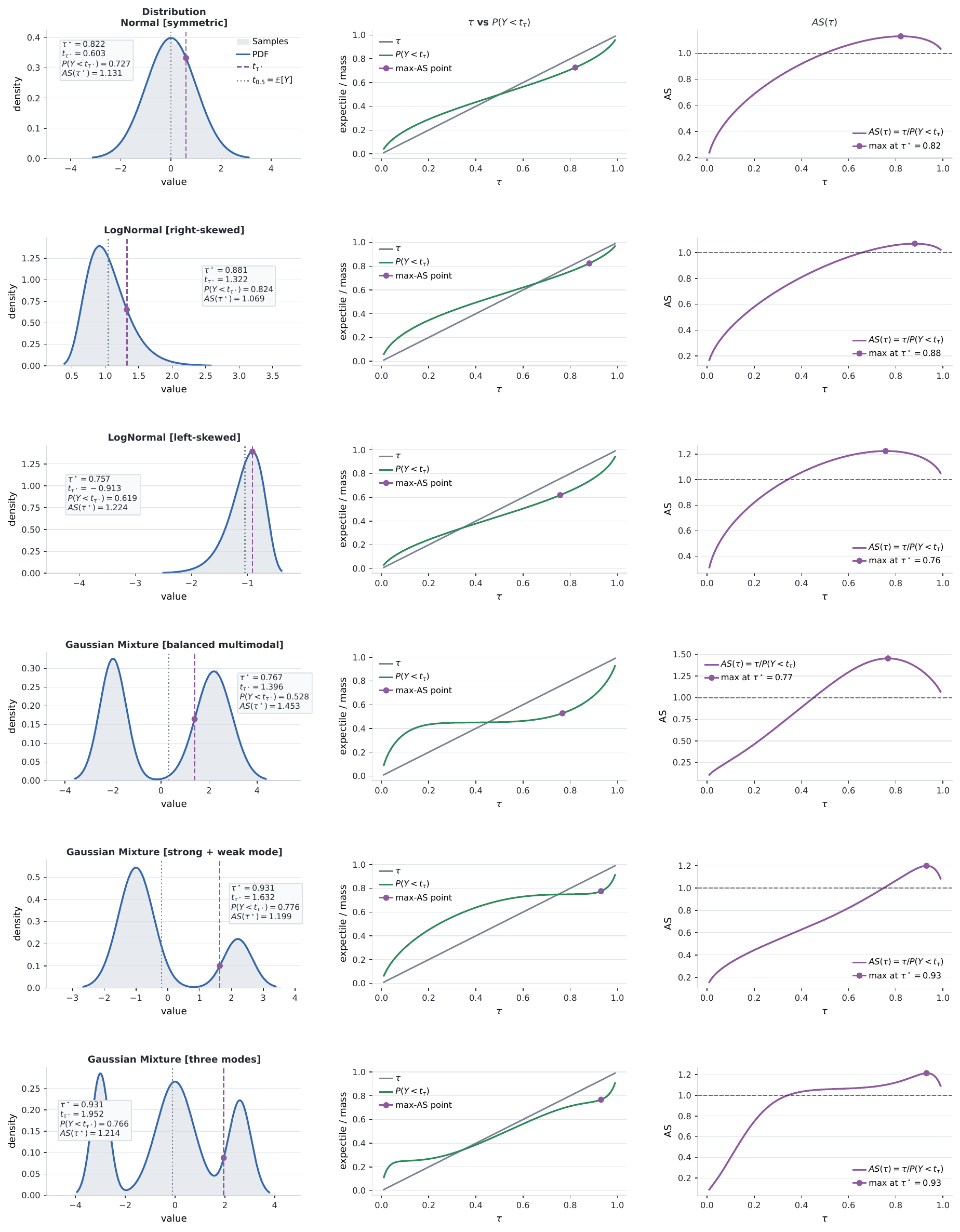}
    \caption{Illustrative examples of the asymmetry score on representative one-dimensional distributions, including symmetric, skewed, and multimodal cases. For each distribution, the left column shows the density with the maximizer-induced split-point $t_{\tau^\star}$ and the reference point $t_{0.5}$; the middle column compares $\tau$ with $\mathbb{P}(Y<t_\tau)$; and the right column plots $\AS(\tau)=\tau/\mathbb{P}(Y<t_\tau)$. The results show that $\AS(\tau)$ identifies aggressive but non-extreme split-points across diverse distributional shapes.}
    \label{fig:as_examples}
\end{figure}

We further illustrate this behavior across representative symmetric, skewed, and multimodal distributions in Figure~\ref{fig:as_examples}. For each distribution, we trace the expectile path $\tau\mapsto t_\tau$, compare the nominal expectile level with its induced cumulative location, and evaluate the corresponding AS. The resulting profiles exhibit distribution-dependent non-monotonic behavior, with interior maxima in the illustrated examples. These observations suggest that AS can serve as an adaptive criterion for identifying selective upper-value split-points without necessarily pushing the split-point toward the extreme upper tail.

Importantly, this analysis concerns the expectile path of a fixed value distribution. In practical critic comparison, independently trained critics may induce different value distributions, so these properties do not establish a universal ordering of critic quality. Instead, they provide a theoretical motivation for AS-based critic ranking, whose structural credibility is further assessed through SDS.

\newpage
\section{Theoretical Analysis of the Self-Consistency Verification}
\label{app:sds}

In this appendix, we prove Theorem~\ref{th:expectile_sc}, derive the normalized self-consistency deviation score, and discuss its scale invariance, structural degeneracy, and role in assessing the credibility of split-point analysis.

\subsection{Proof of Theorem~\ref{th:expectile_sc}}
\begin{proof}
We reuse the notation and identities established in the proof of Proposition~\ref{prop:as_closed_form}. In particular, the expectile balance identity in Eq.~\ref{eq:expectile_balance_mad} gives
\[
(1-\tau)p^-\mathrm{MAD}^-(t_\tau)=\tau p^+\mathrm{MAD}^+(t_\tau).
\]
Hence,
\begin{equation}
\frac{p^+}{p^-}=\frac{(1-\tau)\mathrm{MAD}^-(t_\tau)}{\tau\mathrm{MAD}^+(t_\tau)}.
\label{eq:ratio}
\end{equation}

Let $r:=p^+/p^-$. Since $p^++p^-=1$, we have $p^-=1/(1+r)$ and $p^+=r/(1+r)$. Substituting these expressions into the MAD decomposition in Eq.~\ref{eq:mad_decomp} gives
\[
\mathrm{MAD}(t_\tau)=\frac{r\mathrm{MAD}^+(t_\tau)+\mathrm{MAD}^-(t_\tau)}{1+r}.
\]
Using Eq.~\ref{eq:ratio} to replace $r$, we further obtain
\[
\mathrm{MAD}(t_\tau)=\frac{\mathrm{MAD}^+(t_\tau)\mathrm{MAD}^-(t_\tau)}{\tau\mathrm{MAD}^+(t_\tau)+(1-\tau)\mathrm{MAD}^-(t_\tau)}.
\]
Taking reciprocals yields
\[
\frac{1}{\mathrm{MAD}(t_\tau)}=\frac{\tau}{\mathrm{MAD}^-(t_\tau)}+\frac{1-\tau}{\mathrm{MAD}^+(t_\tau)}.
\]
This completes the proof.
\end{proof}

\subsection{Derivation of the Self-Consistency Deviation Score}
\label{app:sds_derivation}
Starting from Eq.~\ref{eq:self_consistency}, multiplying both sides by $\mathrm{MAD}(t_\tau)\mathrm{MAD}^{+}(t_\tau)\mathrm{MAD}^{-}(t_\tau)$ yields
\[
\mathrm{MAD}^{+}(t_\tau)\mathrm{MAD}^{-}(t_\tau)
=
\tau\,\mathrm{MAD}(t_\tau)\mathrm{MAD}^{+}(t_\tau)
+
(1-\tau)\,\mathrm{MAD}(t_\tau)\mathrm{MAD}^{-}(t_\tau).
\]
Rearranging terms gives
\begin{equation}
\label{eq:mad_stable_identity}
\tau\,\mathrm{MAD}(t_\tau)
\bigl(\mathrm{MAD}^{+}(t_\tau)-\mathrm{MAD}^{-}(t_\tau)\bigr)
=
\mathrm{MAD}^{-}(t_\tau)
\bigl(\mathrm{MAD}^{+}(t_\tau)-\mathrm{MAD}(t_\tau)\bigr).
\end{equation}
This motivates measuring self-consistency through the absolute difference between its two sides. However, this difference scales quadratically with the residual magnitudes, making it sensitive to the value scales of different critics or RL tasks. To reduce sensitivity to residual scales, we normalize the difference by the absolute sum of the two sides, yielding the dimensionless SDS:
\[
\SDS(s,\tau)
=
\frac{
\left|\tau \MAR(s)(\MAR^{+}(s)-\MAR^{-}(s))-\MAR^{-}(s)(\MAR^{+}(s)-\MAR(s))\right|
}{
\max\left\{\left|\tau \MAR(s)(\MAR^{+}(s)-\MAR^{-}(s))\right|+\left|\MAR^{-}(s)(\MAR^{+}(s)-\MAR(s))\right|,
\epsilon
\right\}
}.
\]
A small positive constant \(\epsilon\) is introduced for numerical stability. Under a common positive rescaling of residuals, both the numerator and the unregularized denominator scale quadratically. Thus, SDS is scale-invariant when the numerical floor is inactive. When the floor is active, however, this invariance no longer holds, and a small SDS should not be interpreted as evidence of accurate MAR estimation or expectile alignment.

\subsection{Zero-Point Analysis of SDS}
\label{app:sds_zero_analysis}

To analyze the zero points of SDS, we consider a fixed state-conditional value distribution $Y=Q_\theta(s,A)$ and vary the split-point $v=V_\psi(s)$ while keeping $\tau$ fixed. Extending the previous notation, let $p^-(v)=\mathbb{P}(Y<v)$ and $p^+(v)=\mathbb{P}(Y\geq v)$. By the MAR decomposition, $\MAR(s)=p^+(v)\MAR^{+}(s)+p^-(v)\MAR^{-}(s)$. Substituting this identity into the numerator of SDS and rearranging yields
\begin{equation}
\label{eq:sds_factorization}
\begin{aligned}
&\tau\MAR(s)\bigl(\MAR^{+}(s)-\MAR^{-}(s)\bigr)
-\MAR^{-}(s)\bigl(\MAR^{+}(s)-\MAR(s)\bigr)\\
&=\bigl(\MAR^{+}(s)-\MAR^{-}(s)\bigr)
\bigl(\tau\MAR(s)-p^-(v)\MAR^{-}(s)\bigr)\\
&=\bigl(\MAR^{+}(s)-\MAR^{-}(s)\bigr)
\Bigl(\tau p^+(v)\MAR^{+}(s)-(1-\tau)p^-(v)\MAR^{-}(s)\Bigr).
\end{aligned}
\end{equation}
This factorization identifies two cases in which SDS equals zero. The first satisfies $\tau p^+(v)\MAR^{+}(s)=(1-\tau)p^-(v)\MAR^{-}(s)$, corresponding to the expectile balance condition in Eq.~\ref{eq:expectile_balance_mad} and indicating that the split-point coincides with the nominal $\tau$-expectile. The second occurs when $\MAR^{+}(s)=\MAR^{-}(s)$, which also implies $\MAR^{+}(s)=\MAR^{-}(s)=\MAR(s)$. Unlike the first condition, this equality causes SDS to vanish independently of expectile alignment, introducing a structural false zero. For certain residual distributions, this equality may hold at a split-point that does not coincide with the nominal $\tau$-expectile. Consequently, a zero SDS alone does not necessarily establish expectile alignment and may lead to misleading credibility assessments in critic comparison.

Importantly, under exact population MAR statistics, this degeneracy occurs only when $\AS(s)=1$. To mitigate this issue, our multi-critic aggregation restricts eligibility to critics satisfying $\AS(s)>1$, which implies $\MAR^{-}(s)>\MAR^{+}(s)$ under the population MAR decomposition and thus excludes the structural false zero. Within this eligible region, $\SDS(s,\tau)=0$ is therefore equivalent to expectile alignment. However, when no critic satisfies the eligibility criterion, the fallback selects the critic with the highest AS regardless of whether $\AS(s)>1$. In this case, the structural false zero is not necessarily excluded.

\subsection{Interpretation and Limitations}
SDS assesses whether the three independently estimated MAR statistics satisfy the self-consistency relation implied by expectile regression. A large deviation may arise from inaccurate MAR estimation or misalignment between the critic-induced split-point and its nominal $\tau$-expectile. Conversely, when the individual MAR estimates are accurate and the split-point coincides with the nominal expectile, SDS is expected to be small, reflecting consistency among the estimated residual moments.

However, the structural false-zero degeneracy and errors in MAR estimation can both lead to misleading SDS values. Therefore, SDS is not intended as a definitive measure of critic correctness, but rather as a structural credibility check for the diagnostic information derived from residual statistics.

In fully offline settings, the reliability of critic assessments cannot be directly verified through online policy evaluation. Instead, SDS provides an internal credibility signal by assessing the structural consistency of the residual statistics underlying AS. Incorporating SDS into critic comparison and aggregation allows SPA to penalize AS-based assessments with larger self-consistency discrepancies, reducing their influence on critic routing. Thus, AS characterizes critic selectivity, while SDS provides complementary information about the credibility of this characterization. The empirical benefits of combining these signals are further demonstrated by the ablation studies in Section~\ref{sec:further_analysis}.

\section{Algorithm and Complexity Analysis}
\label{app:algorithm}
This appendix provides pseudocode for the critic diagnosis and aggregation framework as Algorithm~\ref{alg:spa-full} and presents a computational complexity analysis.

\begin{algorithm}
\caption{Critic diagnosis and aggregation framework}
\label{alg:spa-full}
\begin{algorithmic}[1]
\Require Dataset $\mathcal D=\{(s,a,r,s')\}$, critics $\{Q_{\theta_i},V_{\psi_i}\}_{i\in\mathcal I}$
\Require SPA modules $\{p_{\omega_i}\}_{i\in\mathcal I}$, aggregation temperature $\alpha$, credibility scale $\beta$
\Ensure Extracted policy $\pi_\phi$

\Statex
\State \textit{// Step 1: Train SPA modules}

\For{each critic $i\in\mathcal I$}
    \For{each $(s,a)\in\mathcal D$}
        \State $e_i(s,a)\gets Q_{\theta_i}(s,a)-V_{\psi_i}(s)$
    \EndFor

    \State Construct $\mathcal D_{\MAR_i}$, $\mathcal D_{\MAR_i}^{+}$, and $\mathcal D_{\MAR_i}^{-}$ from $\{e_i(s,a)\}$

    \State Train $p_{\omega_i}$ by minimizing
    $\mathcal L_{\mathrm{MSE}_i}(\mathcal D_{\MAR_i},\mathcal D_{\MAR_i}^{+},\mathcal D_{\MAR_i}^{-};\omega_i)$
\EndFor

\Statex
\State \textit{// Step 2: Adaptive routing and aggregation}

\For{each step}
    \State Sample mini-batch
    $\{(s_j,a_j)\}_{j=1}^{B}\sim\mathcal D$

    \For{each $(s_j,a_j)$}
        \For{each critic $i\in\mathcal I$}
            \State $(z_i,z_i^{+},z_i^{-})\gets p_{\omega_i}(h_i(s_j))$
            \State $\AS(s_j,i)\gets z_i^{-}/z_i$
            \State $\SDS(s_j,i)\gets
			\dfrac{\left|\tau_i z_i(z_i^{+}-z_i^{-})-z_i^{-}(z_i^{+}-z_i)\right|}
			{\max\left\{\left|\tau_i z_i(z_i^{+}-z_i^{-})\right|+\left|z_i^{-}(z_i^{+}-z_i)\right|,\epsilon\right\}}$
            \State $u(s_j,i)\gets\AS(s_j,i)\exp\!\left(-\SDS(s_j,i)/\beta\right)$
        \EndFor
        
	    \State $\mathcal I_{s_j}\gets\{i\in\mathcal I:\AS(s_j,i)>1\}$
	    \If{$\mathcal I_{s_j}\neq\emptyset$}
	        \For{each critic $i\in\mathcal I_{s_j}$}
	                \State Normalize the Credited AS:
	                $\tilde u(s_j,i)\gets\dfrac{u(s_j,i)-\mu_{s_j}(u)}{\sigma_{s_j}(u)+\epsilon}$
	                \State Compute routing weights:
	                $w(s_j,i)\gets\dfrac{\exp\!\left(\tilde u(s_j,i)/\alpha\right)}{\sum_{k\in\mathcal I_{s_j}}\exp\!\left(\tilde u(s_j,k)/\alpha\right)}$
	        \EndFor
	        \State $w(s_j,i)\gets0$ for all $i\notin\mathcal I_{s_j}$
	    \Else
	        \State $w(s_j,i)\gets\mathbf{1}\{i=\arg\max_{k\in\mathcal I}\AS(s_j,k)\}$
	    \EndIf
	    \State Compute the aggregated critics:
        	\[Q_{\mathrm{agg}}(s_j,a_j)=\sum_{i\in\mathcal I}w(s_j,i)Q_{\theta_i}(s_j,a_j), 
        	\quad V_{\mathrm{agg}}(s_j)=\sum_{i\in\mathcal I}w(s_j,i)V_{\psi_i}(s_j)\]
    \EndFor

    \Statex
    \State \textit{// Step 3: Downstream policy extraction}
    \If{IQL-Agg}
	    \State Train $\pi_\phi$ using AWR with $Q_{\mathrm{agg}}(s,a)-V_{\mathrm{agg}}(s)$
	    
   \ElsIf{IDQL-Agg}
    	\State Load a pretrained behavior diffusion model $\widehat{\mu}_{\phi}$
    	\State Sample $\{a_j\}_{j=1}^{N}\sim\widehat{\mu}_{\phi}(\cdot\mid s)$ and select $a^*=\arg\max_{a_j}Q_{\mathrm{agg}}(s,a_j)$
    \EndIf
\EndFor

\State \Return $\pi_\phi$

\end{algorithmic}
\end{algorithm}

\subsection{Computational Complexity Analysis}
\label{app:complexity}
We analyze the computational cost of the proposed framework by separating it
into three components: critic-bank training, SPA module training, and adaptive
critic routing. We distinguish between the forward cost
\(\operatorname{Cost}_{\mathrm{fwd}}(\cdot)\) and the training cost
\(\operatorname{Cost}_{\mathrm{train}}(\cdot)\), where the latter includes forward
passes, backward passes, and optimizer updates.

\paragraph{Critic-bank training.}
Our framework operates on a bank of pretrained IQL-style critics with different
expectile levels \(\mathcal{T}=\{\tau_i\}_{i=1}^{K}\). If such a critic bank is not
already available, training the bank requires training \(K\) critics. This cost is
comparable to ensemble-style critic training. For a mini-batch of size \(B\), the
per-update cost of critic-bank training is approximately
\[
\mathcal{O}\!\left(
K B
\left(
\operatorname{Cost}_{\mathrm{train}}(V_{\psi})
+
\operatorname{Cost}_{\mathrm{train}}(Q_{\theta})
\right)
\right).
\]
This cost comes from constructing the candidate critic set rather than from the
SPA module itself. In practice, the critics can be trained independently and in
parallel.

\paragraph{SPA module training.}
For each pretrained critic $i$, training a single SPA module requires constructing the regression targets ${e_i(s,a)}$ and extracting the feature representation $h_{\psi_i}(s)$ from the corresponding value network $V_{\psi_i}$. We instantiate the SPA module as an MLP regressor $p_{\omega_i}$ with $L$ hidden layers that predicts the MAR-related quantities. For a mini-batch of size $B$, each update consists of a forward pass through the frozen critic networks to compute residual targets and features, followed by one optimization step of the SPA module. The resulting per-update cost is
\[
\mathcal{O}\!\left(
B\,(\operatorname{Cost}_{\mathrm{fwd}}(V_{\psi_i}) + \operatorname{Cost}_{\mathrm{fwd}}(Q_{\theta_i}))
\right)
+
\mathcal{O}\!\left(
B\,\operatorname{Cost}_{\mathrm{train}}(p_{\omega_i})
\right).
\]
For an MLP with \(L\) hidden layers, one training iteration of the SPA module includes a forward pass, a backward pass, and a parameter update. Therefore, the SPA training cost can be expressed as \[ \operatorname{Cost}_{\mathrm{train}}(p_{\omega_i}) = \mathcal{O}\!\left( c_{\mathrm{train}} \sum_{m=0}^{L} h_{m}h_{m+1} \right), \] where \(c_{\mathrm{train}}>1\) captures the additional computation introduced by gradient propagation and optimizer updates, while \(h_0\) is the input dimension, \(h_1,\ldots,h_L\) are the hidden-layer widths, and \(h_{L+1}\) is the output dimension. 

For a critic bank of size $K$, the total SPA module training cost scales linearly with
$K$:
\[
\mathcal{O}\!\left(
K B\,(\operatorname{Cost}_{\mathrm{fwd}}(V_{\psi})
+ \operatorname{Cost}_{\mathrm{fwd}}(Q_{\theta})
+ \operatorname{Cost}_{\mathrm{train}}(p_\omega))
\right).
\]
This cost can be amortized when the critic bank is reused for different downstream actors or policy extraction strategies.

\paragraph{Joint training.}
We also consider a joint-training setting, where each SPA module is optimized together with its corresponding critic. In this setting, the critic update already performs the forward passes through $V_{\psi_i}$ and $Q_{\theta_i}$ for value and Bellman training. Therefore, the residual targets and feature representations required by the SPA module can be obtained from the same mini-batch computation, avoiding the additional frozen-critic forward pass used in post-hoc SPA training. The remaining overhead mainly comes from optimizing the SPA regressor $p_{\omega_i}$:
\[
\mathcal{O}\!\left(
K B
\left(
\operatorname{Cost}_{\mathrm{train}}(V_{\psi})
+
\operatorname{Cost}_{\mathrm{train}}(Q_{\theta})
+
\operatorname{Cost}_{\mathrm{train}}(p_\omega)
\right)
\right).
\]
Unlike post-hoc training, joint training optimizes the critic and SPA module simultaneously, reducing total training time by reusing critic-side forward computation and eliminating the separate SPA training stage, despite a slight increase in per-update cost. However, because the SPA module is trained on targets and representations from an evolving critic, its convergence may be affected by critic-side instability.

\paragraph{Adaptive critic routing.}
During actor training or policy extraction, the routing proxy evaluates the SPA-based score for each candidate critic and forms the aggregated critic estimate. In addition to the original critic-side forward computation, routing introduces two extra costs: $K$ SPA module forward passes per mini-batch and the aggregation operations. The additional per-update cost is
\[
\mathcal{O}\!\left(
K B
\left(
\operatorname{Cost}_{\mathrm{fwd}}(p_\omega)
+
\operatorname{Cost}_{\mathrm{agg}}
\right)
\right).
\]
Here, the forward cost of the SPA module is
\[
\operatorname{Cost}_{\mathrm{fwd}}(p_\omega)
=
\mathcal{O}\!\left(
\sum_{m=0}^{L} h_{m}h_{m+1}
\right),
\]
which corresponds to a single forward pass through the MLP regressor with $L$ hidden layers. The aggregation term $\operatorname{Cost}_{\mathrm{agg}}$ includes the computation of AS, SDS, routing weights, and the weighted aggregation of critic outputs. Since these operations only involve scalar scores and linear combinations across the critic bank, we treat $\operatorname{Cost}_{\mathrm{agg}}=\mathcal{O}(1)$. Therefore, once the critic bank and SPA modules are available, adaptive routing is lightweight, relying only on small MLP outputs and linear aggregation; the dominant end-to-end cost remains the construction of the critic bank.

\subsection{Empirical Cost Comparison}
To quantify the computational overhead of SPA-based aggregation, we report both per-update measurements and end-to-end wall-clock training times. The two measurements serve different purposes: Table~\ref{tab:compute_cost} characterizes the cost of individual optimization updates, whereas Table~\ref{tab:compute_cost_total} reports the elapsed time of complete training stages.

All computational measurements were conducted on an NVIDIA L40S GPU. The implementation uses JAX 0.4.30 and Flax 0.8.3. For aggregation methods, the four-critic bank is updated through vectorized parallel computation rather than sequential training. We report two complementary measurements: per-update runtime, which characterizes the computational cost of individual mini-batch updates, and end-to-end wall-clock time, which measures the elapsed training time of the critic, SPA, and actor stages. In joint-training variants, SPA modules are optimized alongside the critics, eliminating the separate post-hoc estimation stage. The per-update measurements and stage-level wall-clock times reflect different measurement scopes and are therefore not directly interchangeable. GPU memory measurements report the observed peak usage during training.

At the per-update level, IQL-Agg increases the critic-step runtime from $1.80$ ms to $2.20$ ms and the actor-step runtime from $0.70$ ms to $0.86$ ms, with an additional $1.26$ ms for post-hoc SPA training. IDQL-Agg exhibits a similar cost profile. Despite using four critics, the vectorized implementation limits the critic-update overhead to approximately $21$--$22\%$ relative to single-critic training, with a moderate increase in the reported GPU memory usage. Joint training introduces only a small additional critic-update cost while eliminating the separate SPA training stage.

The end-to-end measurements in Table~\ref{tab:compute_cost_total} show that IQL-Agg increases total training time from $43.03$ to $71.94$ minutes relative to IQL, while IDQL-Agg increases it from $45.29$ to $76.13$ minutes relative to IDQL, corresponding to overheads of $67.2\%$ and $68.1\%$, respectively. Post-hoc SPA training accounts for $20.53$ and $20.56$ minutes of these totals. Joint training reduces the totals to $52.15$ minutes for IQL-Agg and $56.34$ minutes for IDQL-Agg, saving $19.79$ minutes in each case, or $27.5\%$ and $26.0\%$ relative to their post-hoc variants. The remaining overheads relative to the single-critic backbones are $21.2\%$ and $24.4\%$, respectively. Thus, vectorized critic-bank updates incur a modest per-update overhead, but the separate SPA stage remains a substantial component of the post-hoc training cost; joint optimization removes this stage at a small increase in critic-training time.

Overall, critic-bank construction accounts for a substantial portion of the additional cost. When pretrained critics are available, SPA can be applied post hoc without modifying their training process, and its estimation cost can be amortized across downstream policy extraction strategies. Otherwise, joint training provides an alternative that integrates SPA estimation directly into critic optimization.

\begin{table}[h]
\centering
\caption{
Per-update runtime and peak GPU memory comparison. 
\# Critics denotes the critic-bank size. 
Runtime is reported in milliseconds per mini-batch update for critic training, post-hoc SPA module training, and actor training or policy extraction. 
Memory denotes the peak GPU memory usage during training. 
``J'' denotes joint training without a separate post-hoc SPA stage.
}
\label{tab:compute_cost}
\begin{tabular}{lccccc}
\toprule
Method & \# Critics & Critic Step & SPA Step & Actor Step & Memory \\
\midrule
IQL        & 1 & 1.80 ms & --    & 0.70 ms & 85.96 MB \\
IQL-Agg    & 4 & 2.20 ms & 1.26 ms & 0.86 ms & 101.87 MB \\
IQL-Agg (J)& 4 & 2.25 ms & --    & 0.86 ms & 107.83 MB \\
IDQL       & 1 & 1.82 ms & --    & 0.85 ms & 140.46 MB \\
IDQL-Agg   & 4 & 2.20 ms & 1.25 ms & 1.10 ms & 146.43 MB \\
IDQL-Agg (J)& 4 & 2.23 ms & --   & 1.10 ms & 151.06 MB \\
ACTIVE     & 5 & 1.89 ms & --    & 0.84 ms & 95.35 MB \\
\bottomrule
\end{tabular}
\end{table}

\begin{table}[h]
\centering
\caption{
End-to-end training-time comparison under the unified \(10^6\)-step training budget.
All times are reported in minutes and decomposed into critic training, post-hoc SPA module training, and actor training or policy extraction.
Total Time denotes the overall wall-clock training time.
``J'' denotes joint training without a separate post-hoc SPA stage.
}
\label{tab:compute_cost_total}
\begin{tabular}{lcccc}
\toprule
Method & Critic Train & SPA Train & Actor Train & Total Time \\
\midrule
IQL           & 31.42 min & --       & 11.61 min  & 43.03 min \\
IQL-Agg       & 37.16 min & 20.53 min & 14.25 min  & 71.94 min \\
IQL-Agg (J)   & 37.96 min & --       & 14.19 min  & 52.15 min \\
IDQL          & 30.52 min & --       & 14.77 min  & 45.29 min \\
IDQL-Agg      & 37.35 min & 20.56 min & 18.22 min & 76.13 min \\
IDQL-Agg (J)  & 38.03 min & --       & 18.31 min & 56.34 min \\
ACTIVE        & 32.52 min & --       & 15.43 min  & 47.95 min \\
\bottomrule
\end{tabular}
\end{table}

\newpage
\section{Experimental Details}
\label{app:experimental_details}

\subsection{Experimental Environment}
\label{app:experimental_environment}
We conduct experiments on D4RL~\citep{fu2020d4rl}, covering Gym MuJoCo, AntMaze, and Kitchen tasks, which represent dense-reward locomotion, sparse-reward navigation, and multi-goal manipulation domains, respectively. We use the ``v2'' datasets for Gym MuJoCo and AntMaze, and the ``v0'' datasets for Kitchen.
Protocols for diagnostic analyses and policy-performance evaluation are different. 

For critic comparison analysis, each original D4RL training dataset is randomly split at the transition level into 80\% training transitions and 20\% validation transitions. The SPA module is fitted on the 80\% split, while the held-out 20\% split is used for computing proxy scores. All reported proxy scores in the critic-comparison analysis are obtained by averaging over this held-out validation split. 

The corruption experiments follow the protocol of RIQL~\citep{yang2023towards}. Specifically, critics and SPA modules are trained on the clean training data split, and SDS is then evaluated on both clean and corrupted versions of the same held-out validation split. This ensures that changes in SDS reflect the effect of controlled input corruption rather than changes in the validation samples.

For all policy-evaluation experiments assessing multi-critic aggregation, we use 100\% of the original D4RL training dataset without a validation split, following the standard IQL protocol~\citep{kostrikov2022iql}. Training rewards are preprocessed following the IQL implementation.

Each policy is evaluated over 10 trajectories for MuJoCo locomotion and Kitchen, and over 100 trajectories for AntMaze. For evaluation, episode returns are converted to normalized D4RL scores using D4RL benchmark-specific reference scores. All results are reported as mean $\pm$ standard deviation over five random seeds.

\paragraph{Software.}
We use Python 3.9.24, D4RL 1.1, JAX 0.4.30, Gym 0.23.1, MuJoCo 2.3.7, MuJoCo-py 2.1.2.14, Flax 0.8.3, Optax 0.2.2, NumPy 1.26.4, and SciPy 1.13.1.

\paragraph{Hardware.}
Experiments were run on a Slurm-based compute cluster under Rocky Linux 9.7, using 4 CPU cores, approximately 42 GB of system RAM, and one NVIDIA L40S GPU with 48 GB of GPU memory.

\subsection{Unified Evaluation Protocol}
\label{app:protocol}
To enable a controlled comparison, we evaluate all methods under a unified hyperparameter-selection protocol. Many offline RL benchmarks report results under dataset-specific tuning, where the best configuration is selected separately for each task. While this protocol reflects the best achievable performance after extensive task-level search, it can be overly optimistic for practical offline RL, where online evaluation is unavailable and hyperparameters cannot be selected using environment returns.

We therefore adopt a domain-level selection protocol, using a shared hyperparameter configuration within each domain rather than tuning separately for each dataset. The expectile and AWR settings follow the corresponding IQL and IDQL backbones and were not selected using evaluation returns. This protocol avoids dataset-specific model selection and better aligns with the fully offline motivation of our method.

For IQL-style critics, we use \(\tau=0.7\) for MuJoCo locomotion and Kitchen and \(\tau=0.9\) for AntMaze. The corresponding AWR inverse temperature is \(t=3.0\) for MuJoCo, \(t=10.0\) for AntMaze and \(t=0.5\) for Kitchen. All other hyperparameters are kept fixed within each domain.

\subsection{Baselines}
\label{app:baselines}

We compare our method with both standard offline RL baselines and IQL-style variants. The baselines include TD3+BC, CQL, IQL, IDQL, ACTIVE, PIQL, and UNIQ.

\paragraph{TD3+BC}~\citep{fujimoto2021minimalist} is a policy-constrained actor-critic baseline that augments TD3 with a behavior-cloning regularizer. It constrains the learned policy toward dataset actions while still exploiting the learned Q-function for policy improvement.

\paragraph{CQL}~\citep{kumar2020cql} is a conservative value-learning method that penalizes high Q-values on OOD actions. It serves as a value-regularization baseline for mitigating extrapolation error in offline RL.

\paragraph{IQL}~\citep{kostrikov2022iql} is our primary single-critic baseline. It is a representative in-sample offline RL method and forms the backbone of our method.

\paragraph{IDQL}~\citep{hansenestruch2023idql} replaces the unimodal Gaussian actor in standard IQL with a diffusion-based policy extraction procedure. Since its critic remains IQL-style, IDQL provides a basis for evaluating whether our post-hoc critic diagnosis and aggregation can improve policy extraction when paired with a more expressive actor.

\paragraph{ACTIVE}~\citep{chen2025active} improves IQL by introducing an in-sample value ensemble and adaptive imitation. We treat it as an ensemble-based IQL variant.

\paragraph{PIQL}~\citep{han2025projection} replaces the fixed expectile coefficient in IQL with a projection-based adaptive expectile and introduces a support-constrained policy improvement objective. We include PIQL as an adaptive-expectile baseline, where the expectile regime is coupled with the learned policy during training. 

\paragraph{UNIQ}~\citep{upadhyay2026uniq} introduces state-adaptive expectile regression based on conformally calibrated uncertainty estimates. It uses a multi-expectile value ensemble to adjust conservatism according to local uncertainty. We include UNIQ as another adaptive-expectile baseline for comparison with our post-hoc critic routing method.

\subsection{Hyperparameter Settings}
\label{app:hyperparameters}
The critic and value networks follow the standard IQL parameterization: a two-layer MLP with hidden dimension 256 and ReLU activations. We use Adam with a learning rate of \(3\times 10^{-4}\) for all neural networks. The target critic is updated by exponential moving average with a coefficient of \(0.005\). For actor training, we follow the standard IQL setup for Gaussian policies~\citep{kostrikov2022iql} and the original IDQL setup for diffusion policies~\citep{hansenestruch2023idql}. The main hyperparameters used for IQL-style critic and actor training are summarized in Table~\ref{tab:iql_hyperparameters}.

\begin{table}[h]
\centering
\caption{Default hyperparameters for IQL-style critic and actor training.}
\label{tab:iql_hyperparameters}
\begin{tabular}{ll}
\toprule
Hyperparameter & Value \\
\midrule
Learning rate & \(3\times 10^{-4}\) \\
Critic batch size & 256 \\
Actor batch size & 256 \\
Value / critic network & 2-layer MLP \\
Hidden dimension & 256 \\
Activation & ReLU \\
Discount factor \(\gamma\) & 0.99 \\
Target critic EMA & 0.005 \\
Optimizer & Adam \\
Actor LR schedule & Cosine decay \\
Advantage clipping & 100 \\
\multirow[t]{2}{*}{Expectile level \(\tau\)}
& 0.7 for MuJoCo and Kitchen\\
& 0.9 for AntMaze\\
\multirow[t]{3}{*}{AWR inverse temperature \(t\)}
& 3.0 for MuJoCo \\
& 10.0 for AntMaze \\
& 0.5 for Kitchen \\
Diffusion settings & As in \citet{hansenestruch2023idql} \\
\bottomrule
\end{tabular}
\end{table}

For SPA module training, we extract the penultimate feature representation \(h_{\psi_i}(s)\) from the corresponding value network and train a lightweight MLP with the same optimizer, learning rate, and batch size as the base critic. For critic aggregation, the same routing hyperparameters are used within each domain. The main hyperparameters used for SPA module training are summarized in Table~\ref{tab:spa_hyperparameters}.

\begin{table}[h]
\centering
\caption{Default hyperparameters for the SPA module and multi-critic aggregation.}
\label{tab:spa_hyperparameters}
\begin{tabular}{ll}
\toprule
Hyperparameter & Value \\
\midrule
Input & \(h_{\psi_i}(s)\) \\
Architecture & One-hidden-layer MLP \\
Hidden dimension & 64 \\
Output dimension & 3 \\
Activation & ReLU \\
Optimizer & Adam \\
Learning rate & \(3\times 10^{-4}\) \\
Batch size & 256 \\
Expectile levels \(\mathcal{T}\) & \(\{0.6, 0.7, 0.8, 0.9\}\) \\
Aggregation temperature \(\alpha\) & 1.0 \\
Credibility scale \(\beta\) & 1.0 \\
Numerical stabilizer $\epsilon$ & $10^{-8}$ \\
\bottomrule
\end{tabular}
\end{table}

\newpage
\section{Additional Experiments and Results}
\label{app:additional-experiments}

\subsection{Detailed Critic Comparison Results}
\label{app:correlation}

\begin{figure}[h]
    \centering
    \includegraphics[width=0.88\linewidth]{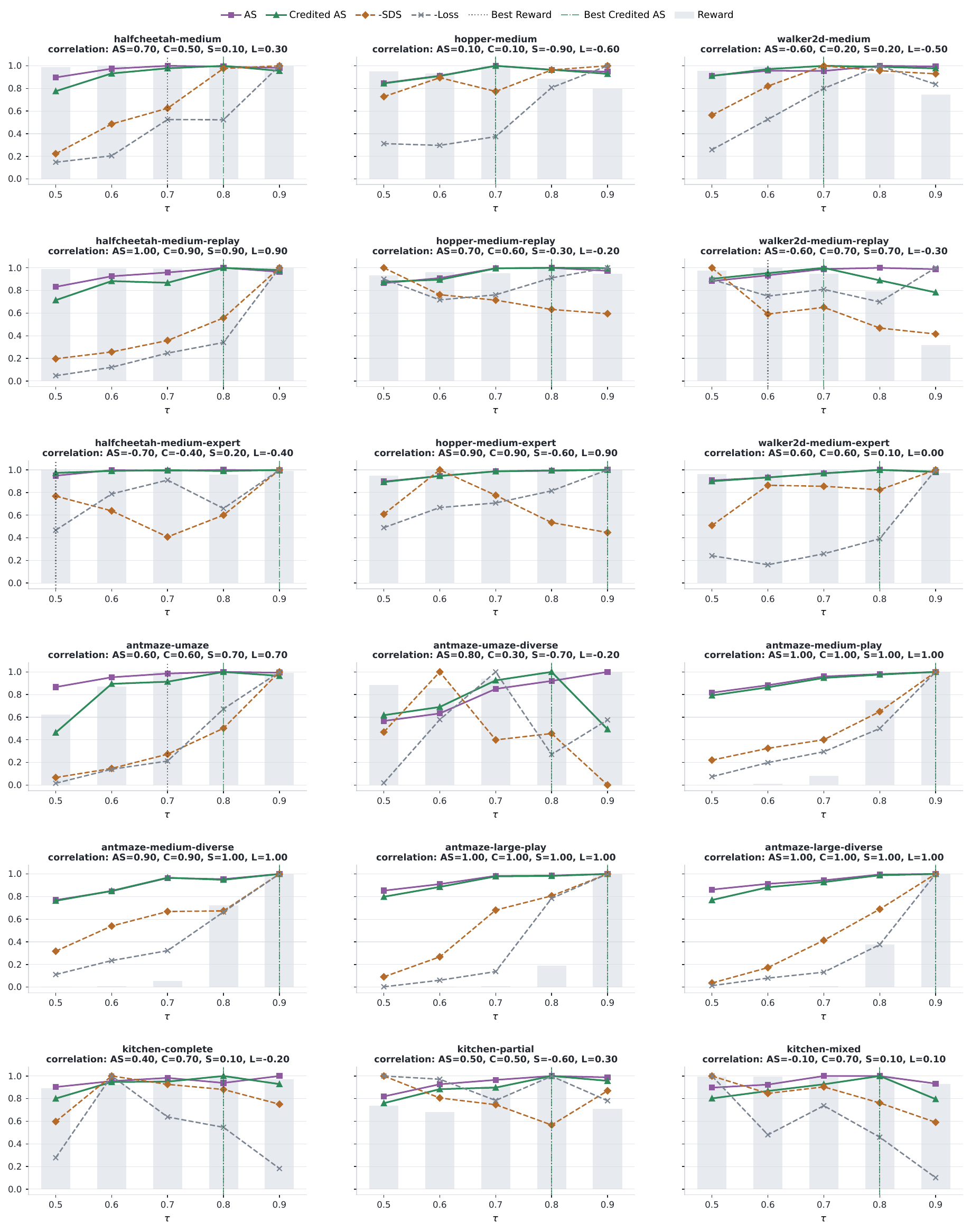}
    \caption{
    Detailed task-wise critic comparison across expectile levels.
    For each task, bars denote the normalized downstream policy return, while lines denote normalized proxy scores, including AS, Credited AS, SDS, and expectile Loss.
    The dotted vertical line marks the best-performing expectile according to policy return, and the dash-dotted vertical line marks the expectile selected by Credited AS.
    }
    \label{fig:app_correlation_detail}
\end{figure}

Figure~\ref{fig:app_correlation_detail} provides the full task-wise visualization supporting the critic-comparison analysis across MuJoCo, AntMaze, and Kitchen. Overall, AS shows a positive correlation with downstream policy performance on 14 out of 18 tasks, suggesting that split-point selectivity is often informative for comparing expectile-trained critics. This trend is particularly clear in AntMaze, where higher-expectile critics frequently yield substantially stronger returns and AS achieves consistently high correlations.

However, raw AS is not uniformly reliable. In several MuJoCo tasks, including walker2d-medium, walker2d-medium-replay, and halfcheetah-medium-expert, stronger split-point asymmetry does not necessarily correspond to better policy performance. A similar failure case appears in kitchen-mixed. These cases indicate that residual asymmetry alone does not consistently predict downstream policy performance. This may reflect either limitations of AS as a utility proxy or errors in estimating the underlying residual statistics.

Although SDS is not intended as a direct ranking criterion, applying SDS as a credibility gate improves the robustness of the proxy. Credited AS yields positive correlations on 17 out of 18 tasks and increases the average Spearman correlation from 0.46 to 0.60. In particular, it corrects the negative correlations on walker2d-medium, walker2d-medium-replay, and kitchen-mixed, indicating that SDS suppresses critics whose residual structure is less self-consistent. The only negative case is halfcheetah-medium-expert, where downstream performance appears less sensitive to the expectile choice. In this case, the returns of different critics are close, which weakens the correlation. This also highlights that our proxy should be interpreted as a diagnostic signal for critic comparison rather than a predictor of final return, since downstream performance can still be affected by the quality of the critic bank and the subsequent actor extraction.

Regarding the other naive proxies, normalized returns do not generally vary monotonically with the expectile level, indicating that the expectile level alone is insufficient for reliable critic comparison. Similarly, the expectile regression loss provides a weaker and less consistent signal. Although lower training losses may be associated with certain expectile regimes, they do not reliably correspond to better downstream policy performance across tasks. These observations suggest that neither the expectile level nor the training loss provides a reliable basis for post-hoc critic comparison, whereas SPA-based scores offer more informative diagnostic signals for ranking and routing pretrained critics.

\subsection{Offline Diagnostic Validation}
\label{app:offline_diagnostic_validation}
This section provides additional analyses of the statistical properties underlying SPA diagnostics, complementing the experiments in Section~\ref{sec:offline_diagnostic_validation}. We examine three aspects: (i) the relationship between expectile levels, residual asymmetry, and structural consistency in a synthetic example; (ii) the estimation quality of the three MAR heads on held-out offline RL data; and (iii) the behavior of AS and SDS as expectile fitting improves under frozen-Q value refitting. Together, these analyses characterize the estimation behavior and structural interpretation of SPA diagnostics independently of downstream policy extraction.

\subsubsection{Synthetic Expectile-Regression Visualization}
\label{app:offline_diagnostic_validation1}

We consider a synthetic regression task with $Y=s^3+\epsilon$, where $\epsilon=30(L-\mathbb{E}[L])$ and $\log L\sim\mathcal{N}(0,0.3^2)$. The resulting noise is zero-mean and right-skewed, with $\mathbb{E}[Y\mid s]=s^3$. This setting provides a simplified analogue of the critic-induced advantage field in IQL-style methods, where action-dependent Q-values are distributed around a state-dependent value estimate. We use a two-hidden-layer MLP with 128 units per layer to learn the expectile and its three associated MAR statistics. The models are trained on $s\in[-4,4]$ and evaluated over $s\in[-6,6]$ for six expectile levels $\tau\in\{0.5,0.6,0.7,0.8,0.9,0.96\}$. Since the noise is state-independent, the population expectile satisfies $t_\tau(s)=s^3+t_\tau(\epsilon)$, yielding an identical residual distribution across states at the true expectile. Thus, this experiment focuses on the estimation and extrapolation behavior of SPA diagnostics rather than heterogeneous state-dependent residual estimation.

\begin{figure}[t]
    \centering
    \includegraphics[width=\linewidth]{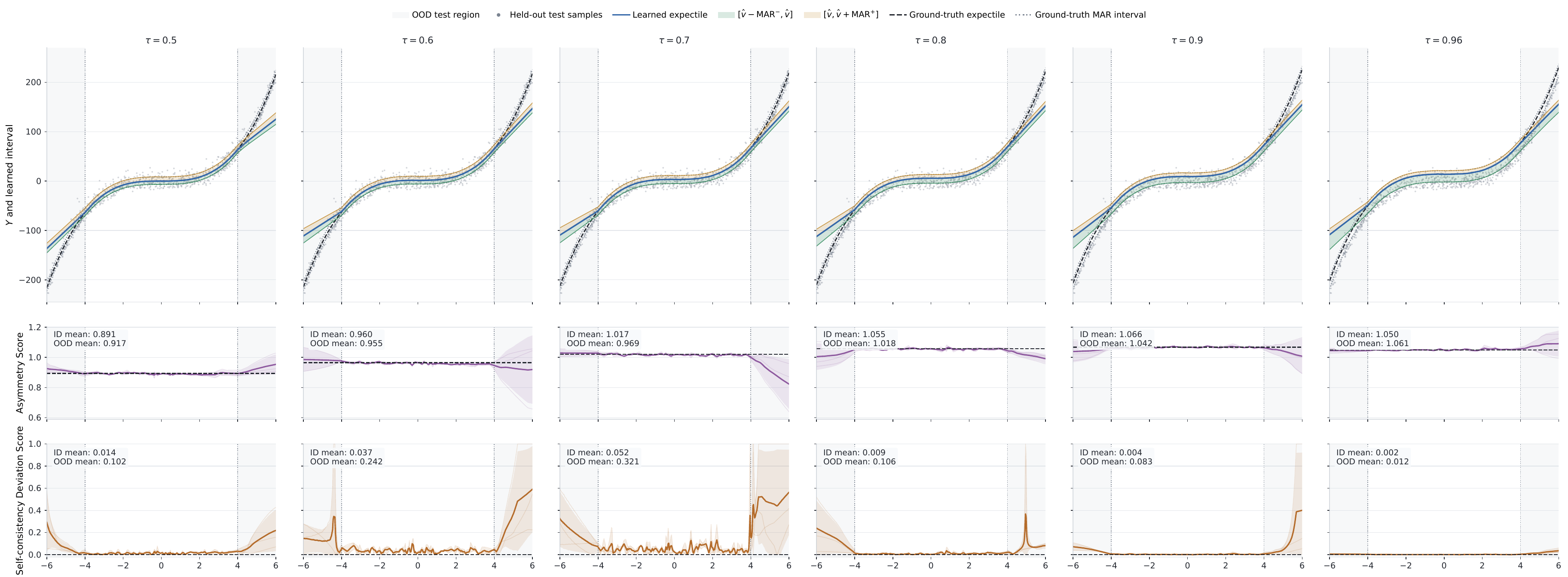}
    \caption{
Synthetic expectile-regression visualization across six expectile levels.
From top to bottom: learned expectiles and MAR intervals, estimated AS compared with population references computed at the true expectiles, and SDS in different regions.
Shaded regions indicate OOD states.
Results are averaged over three random seeds.
}
    \label{fig:synthetic_validation}
\end{figure}

As shown in Figure~\ref{fig:synthetic_validation}, increasing $\tau$ shifts the expectile toward the upper tail of the target distribution, resulting in progressively asymmetric MAR intervals. Correspondingly, AS generally increases with $\tau$, exceeding $1$ at $\tau=0.7$ and reaching its maximum near $\tau=0.9$, before declining slightly at $\tau=0.96$. Within the training region, the estimated AS remains relatively stable across states for each expectile level, accompanied by low mean SDS values ranging from $0.002$ to $0.052$, suggesting greater structural consistency of the underlying MAR estimates. Under OOD extrapolation, the estimated AS exhibits more pronounced variations, accompanied by higher mean SDS values ranging from $0.012$ to $0.321$. These deviations may arise from both expectile-fitting and MAR-estimation errors. Overall, these observations illustrate the complementary roles of AS and SDS: AS characterizes expectile-dependent residual asymmetry, while SDS captures structural inconsistencies that may indicate reduced credibility of the resulting AS estimates.

\subsubsection{Validation of MAR Estimation on Held-Out Data}
\label{app:offline_diagnostic_validation2}
To assess the accuracy of MAR estimation, we conduct held-out validation on three MuJoCo medium-replay datasets: HalfCheetah, Hopper, and Walker2d. Since state-conditional MARs do not have directly observable ground-truth labels in offline RL datasets, we evaluate their estimation quality through binned calibration, comparing predicted MARs with empirical residual magnitudes on held-out data. For each pretrained critic, we train the SPA module on 80\% of the offline transitions and evaluate its predictions on the remaining 20\%, which are excluded from SPA training. We examine four expectile levels $\tau\in\{0.6,0.7,0.8,0.9\}$ across five random seeds.

For each MAR head, held-out samples are divided into 10 equal-frequency bins according to their predictions. We compare the mean predicted MAR with the mean observed residual magnitude within each bin, using the corresponding positive or negative residual subsets for the two side-specific heads. Perfect group-level agreement corresponds to the diagonal $y=x$. This procedure evaluates the agreement between predicted and empirical residual statistics at the group level without requiring repeated samples at individual states.

\begin{figure}
    \centering
    \includegraphics[width=0.67\linewidth]{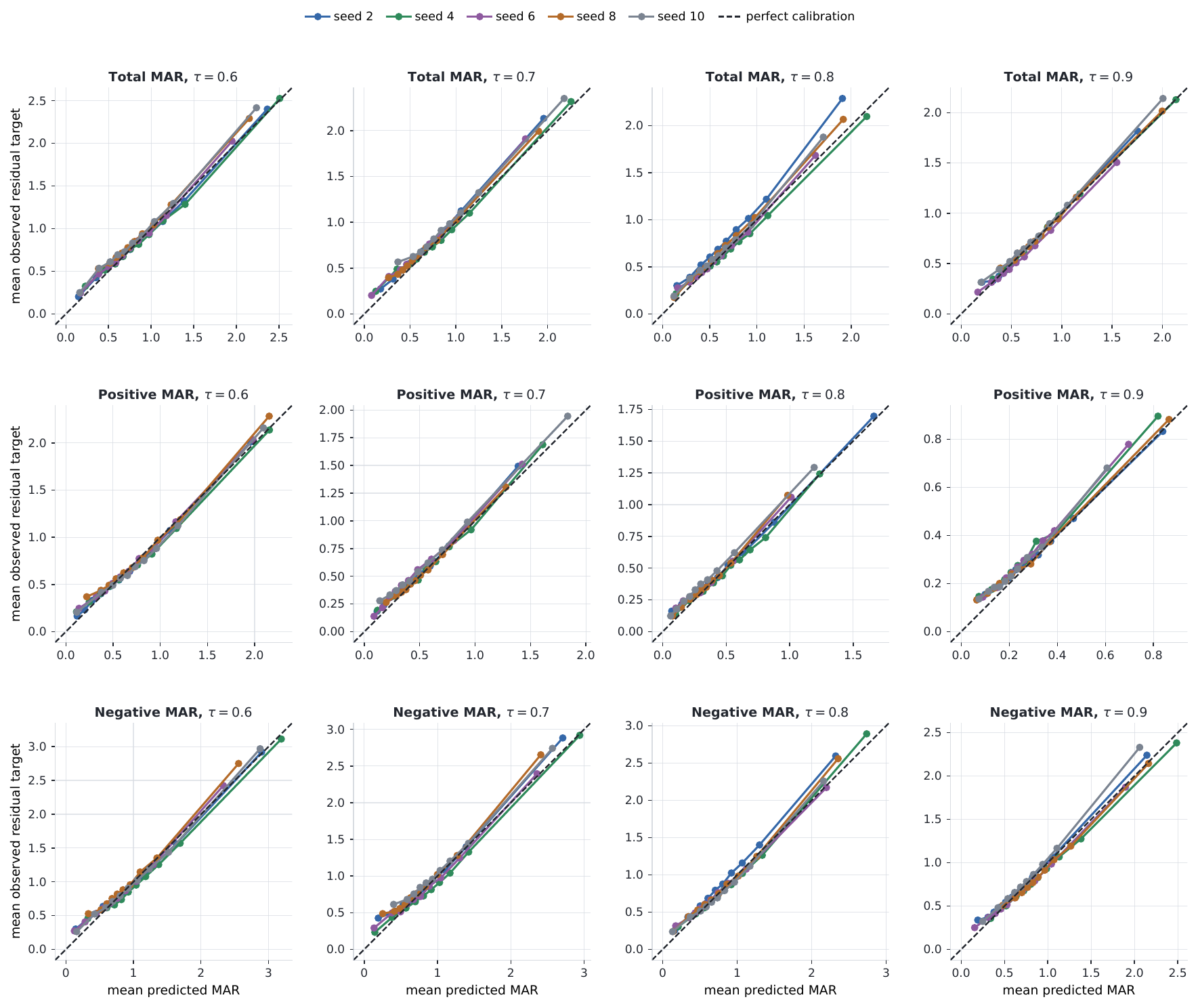}
    \caption{Held-out MAR calibration on halfcheetah-medium-replay across four expectile levels and five random seeds. The dashed diagonal indicates perfect calibration.}
    \label{fig:mar_calibration_halfcheetah}
\end{figure}

\begin{figure}
    \centering
    \includegraphics[width=0.67\linewidth]{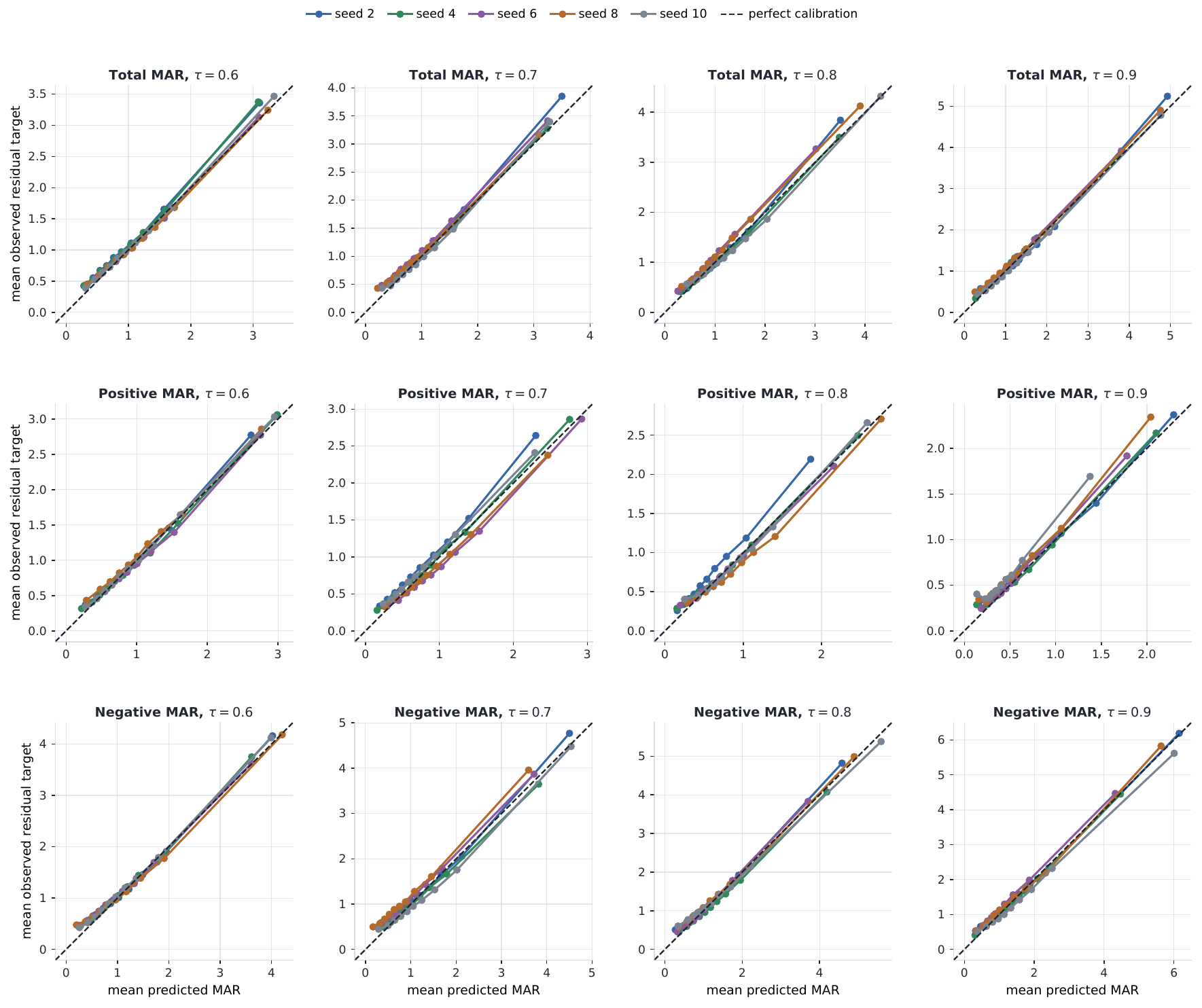}
    \caption{Held-out MAR calibration on hopper-medium-replay.}
    \label{fig:mar_calibration_hopper}
\end{figure}

\begin{figure}
    \centering
    \includegraphics[width=0.67\linewidth]{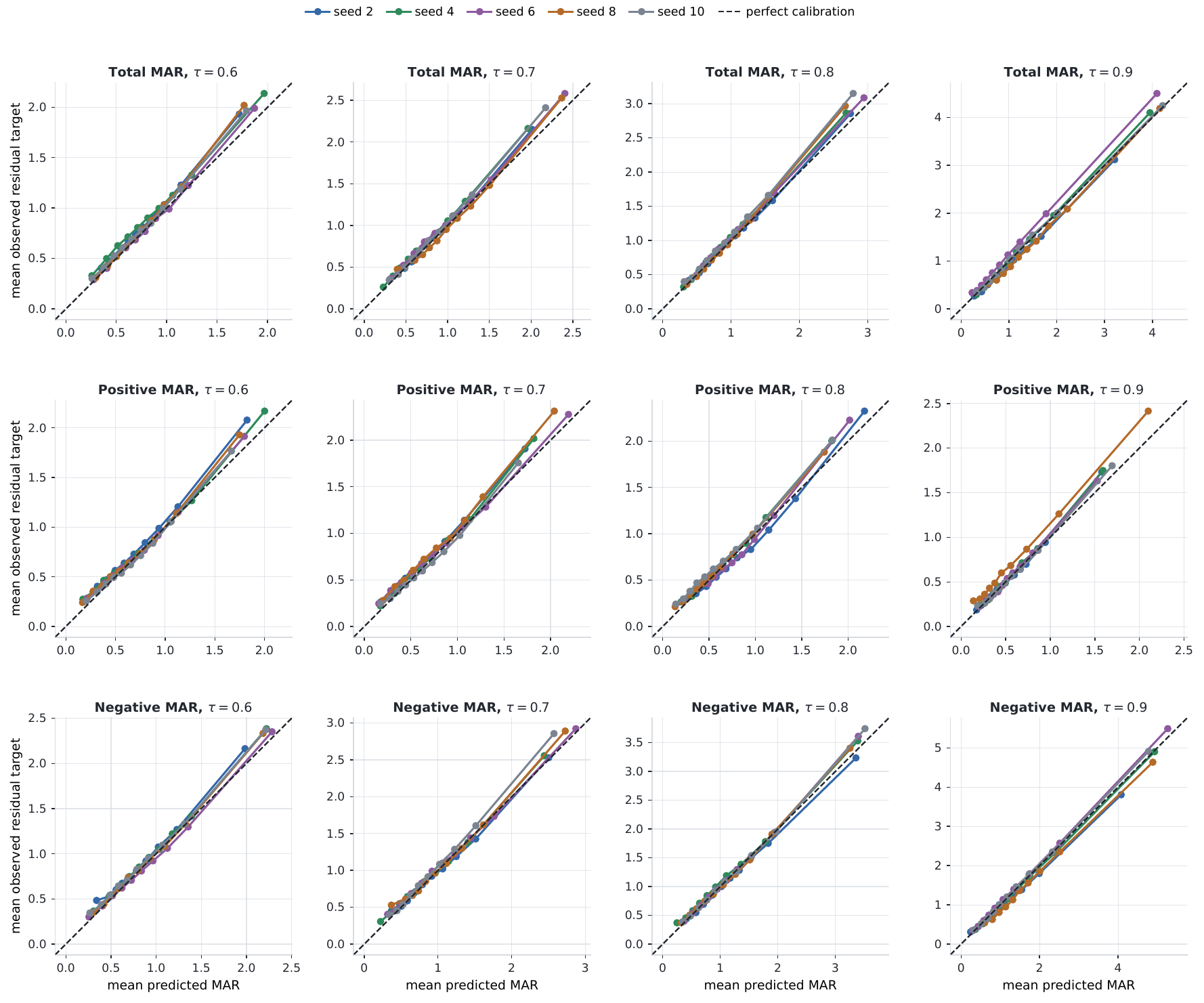}
    \caption{Held-out MAR calibration on walker2d-medium-replay.}
    \label{fig:mar_calibration_walker2d}
\end{figure}

Figures~\ref{fig:mar_calibration_halfcheetah}--\ref{fig:mar_calibration_walker2d} show that the predicted MARs generally agree with the observed residual magnitudes across the three datasets and four expectile levels. The calibration patterns are also consistent across random seeds, although deviations become more noticeable at larger predicted values, particularly for the positive MAR head. These results provide empirical support for the quality of MAR estimation at the group level across different datasets and expectile configurations.

\subsubsection{Diagnostic Behavior under Value Refitting}
\label{app:offline_diagnostic_validation3}
\begin{figure}
    \centering
    \includegraphics[width=0.93\linewidth]{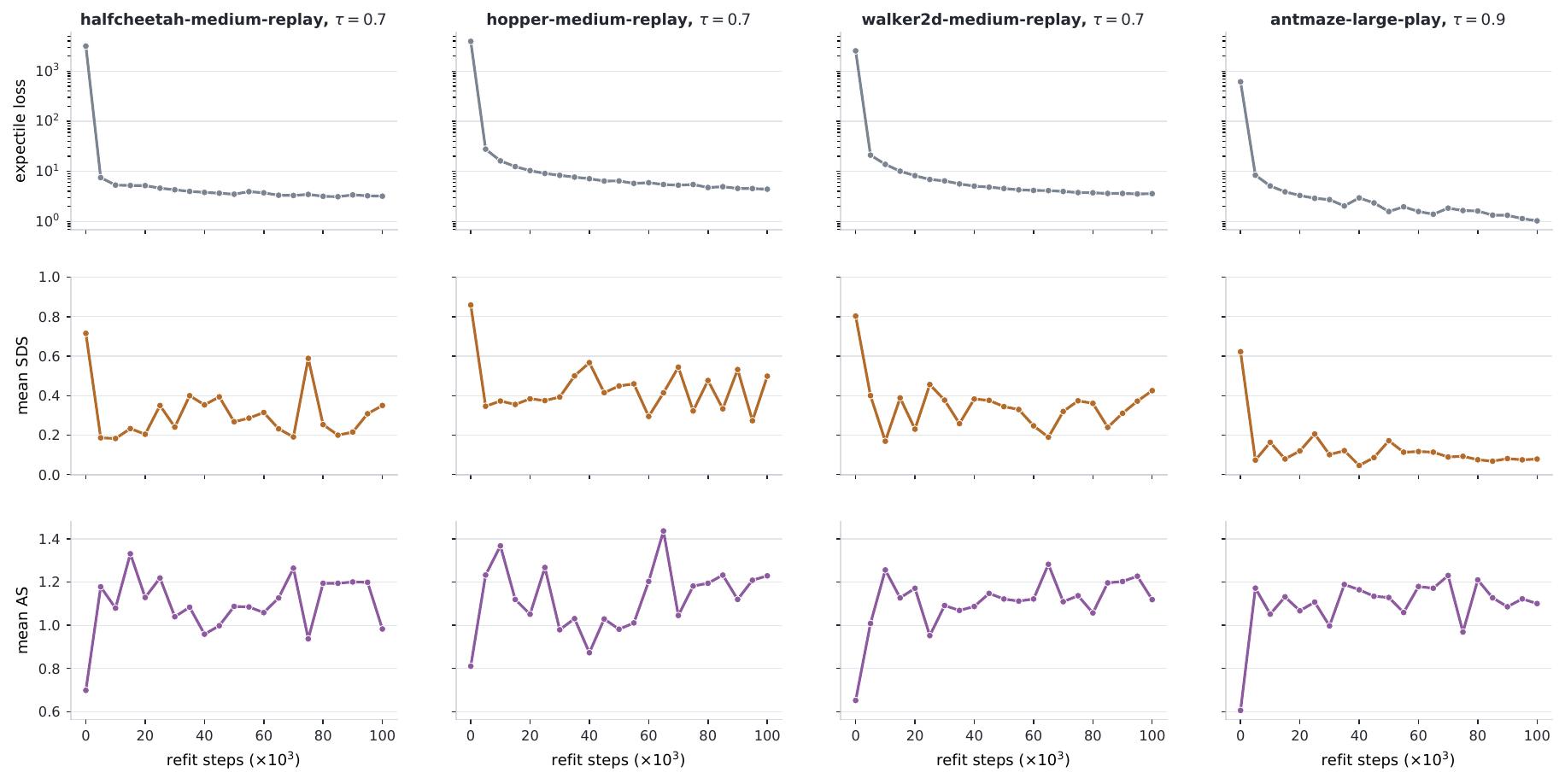}
    \caption{Diagnostic behavior during value-function refitting with frozen Q-functions.}
    \label{fig:value_refitting}
\end{figure}

We further investigate the evolution of AS and SDS during value-function optimization. To isolate the effect of value fitting, the pretrained Q-function is frozen, while the value function and SPA module are updated jointly. Specifically, the SPA module is updated synchronously with the value function, using residual targets $Q_\theta(s,a)-V_\psi(s)$ and state representations $h_\psi(s)$ obtained from the current value network. The held-out expectile loss, mean AS, and mean SDS are tracked throughout training to examine how the diagnostic statistics evolve as the value function becomes better aligned with the fixed Q-function. Experiments are conducted on three MuJoCo medium-replay tasks ($\tau=0.7$) and AntMaze large-play ($\tau=0.9$).

As shown in Figure~\ref{fig:value_refitting}, the held-out expectile loss decreases and gradually converges, accompanied by a pronounced initial reduction in SDS. This correspondence suggests that SDS responds to improvements in the structural self-consistency of the estimated residual statistics during value fitting. Meanwhile, AS increases from its initial values and gradually stabilizes around $1.1$--$1.2$, indicating that the estimated residual asymmetry also evolves as the value function approaches a stable expectile fit. Although both AS and SDS exhibit some fluctuations during later training stages, their overall trajectories illustrate complementary behavior: AS characterizes residual asymmetry, whereas SDS reflects deviations from the structural consistency conditions associated with expectile fitting.

\subsection{Additional Baseline Comparison and Diagnostic Analysis}
\label{app:additional_baseline}

\begin{table}[htp]
\centering
\caption{Additional comparison between originally reported results and reproduced results under our unified protocol. ``Reported'' denotes results from the corresponding papers, while ``Reproduced'' denotes results obtained under our unified experimental protocol. Some reported AntMaze results use the v0 datasets, whereas all reproduced AntMaze results use v2. Reproduced scores are reported as mean $\pm$ standard deviation over five seeds. A dash indicates that the corresponding reported result is unavailable.}
\label{tab:additional_baseline_comparison}
\setlength{\tabcolsep}{3.0pt}
\resizebox{\columnwidth}{!}{
\begin{tabular}{lcrcrcrcrcrrr}
\hline
\multirow{2}{*}{Dataset}
& \multicolumn{2}{c}{IQL}
& \multicolumn{2}{c}{ACTIVE}
& \multicolumn{2}{c}{PIQL}
& \multicolumn{2}{c}{UNIQ}
& \multicolumn{2}{c}{IDQL}
& \multirow{2}{*}{IQL-Agg}
& \multirow{2}{*}{IDQL-Agg} \\
\cline{2-11}
& \multicolumn{1}{c}{Reported}
& \multicolumn{1}{c}{Reproduced}
& \multicolumn{1}{c}{Reported}
& \multicolumn{1}{c}{Reproduced}
& \multicolumn{1}{c}{Reported}
& \multicolumn{1}{c}{Reproduced}
& \multicolumn{1}{c}{Reported}
& \multicolumn{1}{c}{Reproduced}
& \multicolumn{1}{c}{Reported}
& \multicolumn{1}{c}{Reproduced}
& & \\
\hline

halfcheetah-medium
& 47.4 & $47.39 \pm 0.17$ & 50.6 & $48.94 \pm 0.29$ & 48.7 & $48.08 \pm 0.16$ & 48.90 & $46.01 \pm 0.19$ & 51.0 & $51.09 \pm 0.19$ & $47.87 \pm 0.22$ & $52.01 \pm 0.19$ \\

hopper-medium
& 66.3 & $71.33 \pm 1.95$ & 80.1 & $71.98 \pm 3.15$ & 68.2 & $63.67 \pm 5.33$ & 75.60 & $72.10 \pm 4.47$ & 65.4 & $60.48 \pm 8.64$ & $72.44 \pm 4.62$ & $70.52 \pm 6.63$ \\

walker2d-medium
& 78.3 & $82.34 \pm 2.10$ & 84.8 & $82.93 \pm 2.25$ & 83.4 & $58.93 \pm 7.21$ & 85.50 & $81.25 \pm 1.75$ & 82.5 & $85.83 \pm 1.28$ & $82.19 \pm 2.35$ & $78.55 \pm 1.09$ \\

halfcheetah-medium-replay
& 44.2 & $43.57 \pm 1.28$ & 51.1 & $44.85 \pm 0.23$ & 45.5 & $40.72 \pm 1.13$ & 46.00 & $43.06 \pm 0.79$ & 45.9 & $45.60 \pm 0.19$ & $44.71 \pm 0.52$ & $45.96 \pm 0.15$ \\

hopper-medium-replay
& 94.7 & $80.63 \pm 15.22$ & 101.1 & $94.23 \pm 5.63$ & 95.7 & $48.10 \pm 6.74$ & 101.60 & $89.46 \pm 4.26$ & 92.1 & $96.24 \pm 1.24$ & $98.51 \pm 2.39$ & $96.24 \pm 4.32$ \\

walker2d-medium-replay
& 73.9 & $73.39 \pm 1.82$ & 85.4 & $77.94 \pm 7.71$ & 89.5 & $66.93 \pm 3.97$ & 89.40 & $74.38 \pm 4.29$ & 85.1 & $84.05 \pm 5.68$ & $76.87 \pm 6.58$ & $85.03 \pm 4.08$ \\

halfcheetah-medium-expert
& 86.7 & $89.06 \pm 3.40$ & 93.2 & $92.84 \pm 0.70$ & 94.2 & $83.92 \pm 1.10$ & 94.80 & $86.68 \pm 1.91$ & 95.9 & $94.17 \pm 0.15$ & $94.94 \pm 1.69$ & $95.52 \pm 1.82$ \\

hopper-medium-expert
& 91.5 & $95.34 \pm 17.88$ & 89.2 & $109.26 \pm 1.71$ & 101.5 & $23.40 \pm 17.48$ & 111.80 & $96.43 \pm 9.23$ & 108.6 & $108.81 \pm 3.18$ & $104.77 \pm 6.89$ & $104.84 \pm 3.28$ \\

walker2d-medium-expert
& 109.6 & $110.05 \pm 1.31$ & 112.0 & $111.74 \pm 0.42$ & 112.2 & $110.30 \pm 0.65$ & 112.90 & $110.29 \pm 0.64$ & 112.7 & $112.39 \pm 0.15$ & $110.54 \pm 0.25$ & $112.67 \pm 0.03$ \\

\hline
MuJoCo-Total
& 692.6 & 693.10 & 747.5 & 734.70 & 738.9 & 544.06 & 766.50 & 699.66 & 739.2 & 738.66 & 732.84 & 741.34 \\
\hline

antmaze-umaze
& 87.5 & $86.30 \pm 3.10$ & 91.6 & $84.74 \pm 3.53$ & 98.0 & $80.26 \pm 5.37$ & -- & $82.83 \pm 1.95$ & 94.0 & $94.25 \pm 2.09$ & $88.76 \pm 3.41$ & $96.90 \pm 2.08$ \\

antmaze-umaze-diverse
& 62.2 & $65.98 \pm 3.42$ & 78.4 & $73.18 \pm 3.86$ & 98.0 & $6.61 \pm 6.81$ & -- & $62.09 \pm 4.24$ & 80.2 & $80.65 \pm 12.12$ & $73.23 \pm 5.72$ & $84.81 \pm 16.05$ \\

antmaze-medium-play
& 71.2 & $70.70 \pm 1.51$ & 74.6 & $79.09 \pm 2.18$ & 94.0 & $1.02 \pm 0.32$ & -- & $3.14 \pm 1.20$ & 84.5 & $82.87 \pm 5.01$ & $73.54 \pm 3.46$ & $81.63 \pm 2.06$ \\

antmaze-medium-diverse
& 70.0 & $67.72 \pm 3.62$ & 75.2 & $77.35 \pm 4.05$ & 88.0 & $0.50 \pm 0.13$ & -- & $3.06 \pm 0.21$ & 84.8 & $81.32 \pm 6.85$ & $76.59 \pm 4.05$ & $84.21 \pm 0.74$ \\

antmaze-large-play
& 39.6 & $43.43 \pm 2.74$ & 51.2 & $45.25 \pm 5.78$ & 72.0 & $0.00 \pm 0.00$ & -- & $0.22 \pm 0.17$ & 63.5 & $68.11 \pm 2.56$ & $45.56 \pm 4.27$ & $71.19 \pm 2.47$ \\

antmaze-large-diverse
& 47.5 & $37.75 \pm 3.96$ & 48.2 & $42.57 \pm 0.92$ & 50.0 & $0.00 \pm 0.00$ & -- & $0.14 \pm 0.24$ & 67.9 & $73.85 \pm 5.09$ & $46.39 \pm 4.10$ & $73.36 \pm 8.82$ \\

\hline
AntMaze-Total
& 378.0 & 371.88 & 419.2 & 402.19 & 500.0 & 88.39 & -- & 151.47 & 474.9 & 481.05 & 404.08 & 492.10 \\
\hline

kitchen-complete
& 62.5 & $68.51 \pm 5.38$ & 66.5 & $69.23 \pm 2.89$ & 87.0 & $48.78 \pm 9.27$ & -- & $59.10 \pm 6.96$ & -- & $70.23 \pm 2.76$ & $63.94 \pm 6.74$ & $70.83 \pm 2.69$ \\

kitchen-partial
& 46.3 & $55.56 \pm 8.85$ & 73.5 & $72.27 \pm 3.35$ & 67.0 & $26.57 \pm 11.30$ & -- & $39.10 \pm 7.01$ & -- & $64.16 \pm 5.64$ & $58.72 \pm 5.30$ & $67.35 \pm 5.44$ \\

kitchen-mixed
& 51.0 & $49.47 \pm 4.67$ & 73.2 & $41.99 \pm 10.65$ & 55.5 & $31.74 \pm 8.21$ & -- & $48.14 \pm 4.47$ & -- & $63.03 \pm 1.75$ & $61.54 \pm 3.45$ & $63.77 \pm 1.42$ \\

\hline
Kitchen-Total
& 159.8 & 173.54 & 213.2 & 183.48 & 209.5 & 107.10 & -- & 146.34 & -- & 197.42 & 184.20 & 201.95 \\
\hline

\end{tabular}
}
\end{table}

Table~\ref{tab:additional_baseline_comparison} provides a detailed comparison between originally reported results and reproduced results under our unified experimental protocol. The reproduced IQL and IDQL results remain close to their reported domain totals, demonstrating the robustness and generalizability of these backbone methods across different task settings. In contrast, ACTIVE, PIQL, and UNIQ, which modify IQL critic learning, exhibit larger discrepancies, suggesting greater sensitivity to implementation details and hyperparameter configurations.

Although the originally reported performance of ACTIVE can be reproduced with task-specific hyperparameter settings, its performance decreases under our unified protocol. This discrepancy reflects its reliance on task-specific adjustments, whereas our evaluation adopts shared configurations within each domain. More substantial reproduction difficulties arise for methods involving adaptive expectile learning. UNIQ, which employs state-adaptive expectile selection, achieves a MuJoCo total of $699.66$ under our reproduction, compared with the reported $766.50$. Its performance also degrades considerably on AntMaze and Kitchen, which were not evaluated in the original study. We conjecture that dynamically varying expectile levels may interact unfavorably with bootstrapped critic updates, potentially affecting learning stability. Similarly, PIQL reports strong performance through adaptive expectile learning, but its official implementation was unavailable at the time of our experiments. Our independent reimplementation did not reproduce the reported results, particularly on AntMaze and Kitchen, and is therefore included only as an approximate reference.

These observations demonstrate the practical importance of evaluation under a unified protocol, where performance differences are less dependent on task-specific tuning. They also highlight the challenges of reliably reproducing training-time critic adaptation mechanisms across diverse tasks. In contrast, our post-hoc aggregation operates directly on pretrained critics without modifying their learning objectives, separating critic selection from the underlying training process.

\subsection{Ablation of Diagnostic Design Choices}
\label{app:diagnostic_design}
In addition to SDS-based credibility weighting, our diagnostic signal incorporates two further design choices. The AS eligibility filter restricts routing to critics satisfying $\AS(s,i)>1$, prioritizing critics with stronger residual selectivity. Meanwhile, SDS normalization in Eq.~\ref{eq:sds_score} aims to improve the comparability of credibility scores across critics. To examine their contributions, we conduct a $2\times2$ ablation by independently enabling or disabling these components while keeping the remaining routing and training settings unchanged.

\begin{table*}[h]
\centering
\caption{
Ablation of AS eligibility filtering and SDS normalization.
Scores are reported as mean $\pm$ standard deviation over five seeds.
``None'' disables both components, while ``Both'' enables AS eligibility filtering and SDS normalization.
The highest mean score in each row is shown in bold.
}
\label{tab:diagnostic_design}
{
\begin{tabular}{lrrrr}
\hline
\multicolumn{1}{c}{Dataset}
& \multicolumn{1}{c}{None}
& \multicolumn{1}{c}{Eligibility Filtering}
& \multicolumn{1}{c}{SDS Norm.}
& \multicolumn{1}{c}{Both} \\
\hline

halfcheetah-medium
& 47.80 $\pm$ 0.10
& \textbf{47.87} $\pm$ 0.09
& 47.76 $\pm$ 0.17
& \textbf{47.87} $\pm$ 0.22 \\

hopper-medium
& 70.62 $\pm$ 4.31
& \textbf{74.89} $\pm$ 6.70
& 74.43 $\pm$ 1.71
& 72.44 $\pm$ 4.62 \\

walker2d-medium
& 79.51 $\pm$ 2.99
& \textbf{82.93} $\pm$ 1.18
& 78.31 $\pm$ 2.21
& 82.19 $\pm$ 2.35 \\

halfcheetah-medium-replay
& 44.08 $\pm$ 2.81
& 44.60 $\pm$ 1.36
& 44.39 $\pm$ 1.48
& \textbf{44.71} $\pm$ 0.52 \\

hopper-medium-replay
& 97.70 $\pm$ 3.95
& \textbf{98.81} $\pm$ 7.94
& 97.60 $\pm$ 2.03
& 98.51 $\pm$ 2.39 \\

walker2d-medium-replay
& 73.34 $\pm$ 11.59
& 75.16 $\pm$ 5.95
& 75.00 $\pm$ 3.68
& \textbf{76.87} $\pm$ 6.58 \\

halfcheetah-medium-expert
& 92.56 $\pm$ 1.82
& 91.18 $\pm$ 2.83
& 89.93 $\pm$ 2.47
& \textbf{94.94} $\pm$ 1.69 \\

hopper-medium-expert
& \textbf{105.07} $\pm$ 6.09
& 104.10 $\pm$ 5.89
& 104.25 $\pm$ 4.08
& 104.77 $\pm$ 6.89 \\

walker2d-medium-expert
& 110.25 $\pm$ 0.53
& \textbf{110.73} $\pm$ 0.21
& 110.45 $\pm$ 0.38
& 110.54 $\pm$ 0.25 \\

\hline
MuJoCo-Total
& 720.93
& 730.27
& 722.12
& \textbf{732.84} \\
\hline

antmaze-umaze
& 88.69 $\pm$ 7.01
& \textbf{88.77} $\pm$ 6.50
& 86.39 $\pm$ 2.10
& 88.76 $\pm$ 3.41 \\

antmaze-umaze-diverse
& 72.97 $\pm$ 7.47
& \textbf{75.84} $\pm$ 7.46
& 71.72 $\pm$ 4.03
& 73.23 $\pm$ 5.72 \\

antmaze-medium-play
& 69.92 $\pm$ 4.64
& 73.30 $\pm$ 5.80
& 70.49 $\pm$ 2.98
& \textbf{73.54} $\pm$ 3.46 \\

antmaze-medium-diverse
& 72.04 $\pm$ 2.37
& 74.60 $\pm$ 2.90
& 73.72 $\pm$ 2.29
& \textbf{76.59} $\pm$ 4.05 \\

antmaze-large-play
& 45.26 $\pm$ 3.52
& \textbf{46.91} $\pm$ 4.48
& 45.97 $\pm$ 4.05
& 45.56 $\pm$ 4.27 \\

antmaze-large-diverse
& 44.71 $\pm$ 2.60
& 43.51 $\pm$ 6.87
& 42.76 $\pm$ 3.69
& \textbf{46.39} $\pm$ 4.10 \\

\hline
AntMaze-Total
& 393.58
& 402.93
& 391.05
& \textbf{404.08} \\
\hline

kitchen-complete
& 56.79 $\pm$ 6.75
& \textbf{66.59} $\pm$ 1.81
& 60.81 $\pm$ 5.33
& 63.94 $\pm$ 6.74 \\

kitchen-partial
& 56.39 $\pm$ 5.02
& 57.34 $\pm$ 3.80
& 57.39 $\pm$ 3.38
& \textbf{58.72} $\pm$ 5.30 \\

kitchen-mixed
& \textbf{62.14} $\pm$ 3.26
& 55.24 $\pm$ 1.76
& 61.99 $\pm$ 4.87
& 61.54 $\pm$ 3.45 \\

\hline
Kitchen-Total
& 175.31
& 179.18
& 180.19
& \textbf{184.20} \\
\hline

\end{tabular}
}
\end{table*}

As shown in Table~\ref{tab:diagnostic_design}, the AS eligibility filter provides more consistent performance improvements across all three domains, increasing the totals from $720.93$ to $730.27$ on MuJoCo, from $393.58$ to $402.93$ on AntMaze, and from $175.31$ to $179.18$ on Kitchen without SDS normalization. Similar improvements are observed when normalization is enabled. These results demonstrate the benefit of prioritizing critics with stronger residual selectivity. Moreover, the eligibility criterion helps mitigate the false-zero degeneracy of SDS, as the degenerate case is excluded under exact population residual statistics.

In contrast, SDS normalization alone provides only marginal improvements on MuJoCo and slightly reduces performance on AntMaze. However, it yields more substantial gains on Kitchen, improving the domain total from $175.31$ to $180.19$ without eligibility filtering and from $179.18$ to $184.20$ when combined with eligibility filtering. This improvement is particularly relevant to Kitchen, where heterogeneous multi-goal behaviors may introduce greater variability in residual statistics and diagnostic estimates. Normalizing SDS therefore facilitates more stable credibility weighting across heterogeneous tasks, improving the general applicability of the diagnostic design. Overall, these results highlight the importance of AS eligibility filtering and the complementary role of SDS normalization, supporting their combined use in the proposed routing mechanism.

\subsection{Implementation of Aggregation Baselines}
\label{app:aggregation-baseline}
This section provides implementation details for the aggregation baselines evaluated in Section~\ref{sec:further_analysis}. All methods use the same pretrained bank $\mathcal{T}=\{0.6,0.7,0.8,0.9\}$.

\paragraph{Oracle Single-Critic Performance (Oracle Single).}
For each task and random seed, we independently extract a policy from each critic in the pretrained bank, following the same policy extraction and evaluation protocol as IQL. We select the highest evaluation return within each seed and report the mean and standard deviation of these seed-wise maxima over five seeds. Oracle Single thus provides a hindsight reference for the best available single critic, using environment feedback that is unavailable to offline critic selection.

\paragraph{Shuffled Aggregation (Shuffled Agg).}
For each state, Shuffled Agg randomly permutes the SPA-module outputs $(z,z^+,z^-)$ as a tuple across critics, together with their corresponding nominal expectile levels, before computing AS, SDS, and Credited AS. Specifically, critic $i$ receives the SPA outputs and expectile level $\tau_j$ of critic $j$. The permutation is resampled at each actor-training step, disrupting the correspondence between critics and their SPA-derived diagnostic information while preserving the original pairing between the shuffled outputs and their nominal expectile levels. Since the shuffling changes only the assignment of SPA-module outputs and their associated expectile levels without modifying their numerical values, the subsequent eligibility filtering, standardization, softmax weighting, and fallback procedures follow the same protocol as IQL-Agg. The resulting weights are used to aggregate the original pretrained critics for policy extraction. This experiment examines whether effective SPA-based aggregation depends on preserving the correspondence between each critic and its diagnostic information.

\paragraph{Uniform Aggregation (Uniform Agg).}
For each state, Uniform Agg assigns equal weights $w_i=1/K$ to all $K$ critics, yielding the aggregated critic pair:
\begin{equation}
    Q_{\mathrm{uniform}}(s,a) = \frac{1}{K}\sum_{i\in\mathcal I} Q_{\theta_i}(s,a), \qquad V_{\mathrm{uniform}}(s) = \frac{1}{K}\sum_{i\in\mathcal I} V_{\psi_i}(s).
    \label{eq:uniform_agg_critic}
\end{equation}
This preserves the same critic bank and aggregation procedure while removing diagnostic-based selective weighting. This comparison examines whether diagnostic-guided critic weighting improves upon uniform aggregation of the same pretrained critics.

\paragraph{Static Aggregation (Static Agg).}
Static Agg constructs fixed global weights by averaging the state-dependent routing weights $w(s,i)$ defined in Eq.~\ref{eq:routing_weight}. For each task and random seed, we compute the final routing weights over the offline dataset using the same eligibility filtering, standardization, softmax, and fallback procedures as IQL-Agg. The resulting weights are averaged across states:
\begin{equation}
    \bar{w}_i = \frac{1}{|\mathcal{D}|} \sum_{(s,a,r,s') \in \mathcal{D}} w(s,i),
    \label{eq:static_weights}
\end{equation}
where $\mathcal{D}$ denotes the offline dataset used for actor training. The averaged weights remain fixed across states and are used to aggregate the pretrained critics for downstream policy extraction:
\begin{equation}
    Q_{\mathrm{static}}(s,a) = \sum_{i\in\mathcal I} \bar{w}_i Q_{\theta_i}(s,a), \qquad V_{\mathrm{static}}(s) = \sum_{i\in\mathcal I} \bar{w}_i V_{\psi_i}(s).
    \label{eq:static_agg_critic}
\end{equation}
We then train a new actor using the same architecture and optimization protocol as IQL-Agg, while keeping the critic bank unchanged. This comparison evaluates the benefit of state-dependent weighting over fixed weights derived from the same SPA routing mechanism.

\paragraph{Performance-Weighted Aggregation (PW-Agg).}
For each task, PW-Agg constructs fixed global weights from the evaluation returns of individual critic-induced policies. Following the same standardization procedure as IQL-Agg, we normalize the returns across critics before applying softmax:
\begin{equation}
    \widetilde{R}_i = \frac{R_i-\mu_R}{\sigma_R+\epsilon}, \qquad w_i^{\mathrm{PW}} = \frac{\exp(\widetilde{R}_i)}{\sum_{j=1}^{K}\exp(\widetilde{R}_j)},
    \label{eq:pw_weights}
\end{equation}
where $R_i$ denotes the evaluation return of the policy associated with critic $i$, and $\mu_R$ and $\sigma_R$ denote the mean and standard deviation of returns across critics. The resulting weights remain fixed across states, yielding the aggregated critic pair:
\begin{equation}
    Q_{\mathrm{PW}}(s,a) = \sum_{i\in\mathcal I} w_i^{\mathrm{PW}} Q_{\theta_i}(s,a), \qquad V_{\mathrm{PW}}(s) = \sum_{i\in\mathcal I} w_i^{\mathrm{PW}} V_{\psi_i}(s).
    \label{eq:pw_agg_critic}
\end{equation}
We then train a new actor using the same architecture and optimization protocol as IQL-Agg. Since PW-Agg relies on evaluation returns unavailable in fully offline settings, it serves as a performance-informed fixed-weight reference for comparing critic aggregation strategies.

\paragraph{Expectile-Balance Mixture of Experts (EB-MoE).}
EB-MoE provides an alternative learnable state-dependent routing strategy based on deviations from the first-order optimality condition of expectile regression. For each pretrained critic $i$, we define the residual $e_i(s,a)=Q_{\theta_i}(s,a)-V_{\psi_i}(s)$. At the expectile optimum, the residual distribution satisfies $(1-\tau_i)\mathbb{E}[(-e_i)_+\mid s]=\tau_i\mathbb{E}[(e_i)_+\mid s]$, where $(x)_+=\max(x,0)$. Deviations from this condition provide a state-dependent signal for critic routing.

Since the conditional residual moments are not directly available at individual states, we train a two-head MLP on the pretrained value representation $h_{\psi_i}(s)$ to estimate the negative and positive partial moments:
\begin{equation}
    L_i(s)=\mathbb{E}[(-e_i)_+\mid s], \qquad U_i(s)=\mathbb{E}[(e_i)_+\mid s].
    \label{eq:moe_moments}
\end{equation}
Following the SPA module architecture, the two heads are trained using the corresponding sample-wise regression targets:
\begin{equation}
    \ell_i(s,a)=\max(-e_i(s,a),0), \qquad u_i(s,a)=\max(e_i(s,a),0).
    \label{eq:moe_targets}
\end{equation}
We jointly minimize the mean-squared errors between the predicted moments and their respective targets. Based on the resulting estimates $\widehat{L}_i(s)$ and $\widehat{U}_i(s)$, we define the expectile-balance deviation as
\begin{equation}
    \widehat{\Delta}_i(s) = \left|(1-\tau_i)\widehat{L}_i(s)-\tau_i\widehat{U}_i(s)\right|.
    \label{eq:moe_deviation}
\end{equation}
Smaller deviations indicate closer agreement with the nominal expectile-balance condition. We standardize the deviations across critics and apply softmax to their negative values to obtain the routing weights:
\begin{equation}
    \widetilde{\Delta}_i(s) = \frac{\widehat{\Delta}_i(s)-\mu_{\Delta}(s)}{\sigma_{\Delta}(s)+\epsilon}, \qquad w_i^{\mathrm{EB}}(s) = \frac{\exp(-\widetilde{\Delta}_i(s))}{\sum_{j=1}^{K}\exp(-\widetilde{\Delta}_j(s))},
    \label{eq:moe_weights}
\end{equation}
where $\mu_{\Delta}(s)$ and $\sigma_{\Delta}(s)$ denote the mean and standard deviation of the deviations across critics. The resulting weights are used to aggregate the pretrained critics:
\begin{equation}
    Q_{\mathrm{EB}}(s,a) = \sum_{i\in\mathcal I} w_i^{\mathrm{EB}}(s) Q_{\theta_i}(s,a), \qquad V_{\mathrm{EB}}(s) = \sum_{i\in\mathcal I} w_i^{\mathrm{EB}}(s) V_{\psi_i}(s).
    \label{eq:moe_agg_critic}
\end{equation}
We then train a new actor using the same architecture and optimization protocol as IQL-Agg. This aggregation strategy serves as a generic learnable MoE routing baseline, in which critic weights are determined by estimated deviations from the expectile-regression optimality condition. The experiment examines whether state-dependent routing guided by these deviations can achieve comparable performance to the proposed SPA-based diagnostic strategy.

\subsection{Diversity and Stability Analysis of Critic Aggregation}
\label{app:aggregation_analysis}
The proposed multi-critic aggregation relies on a bank of critics trained under different expectile regimes and adaptively combines their predictions through state-dependent routing. This raises two questions: whether the critic bank provides sufficiently diverse candidates for aggregation, and whether the resulting routing behavior remains stable across states, trajectories, and training iterations. We first examine the diversity induced by different expectile levels and the corresponding state-dependent routing preferences. We then analyze the temporal evolution of effective expectiles along policy trajectories using fixed post-hoc critic banks. Finally, we investigate the training dynamics of effective expectiles under joint training, where the critics and SPA modules are optimized together.

\subsubsection{Critic Bank Diversity}
\label{app:critic_diversity_analysis}
Although Figure~\ref{fig:app_correlation_detail} demonstrates that critics trained with different expectile levels can yield different downstream policy performance, it remains unclear whether the resulting critic bank provides sufficient diversity to support our mixture-of-experts-style aggregation framework. In particular, differences in policy performance do not necessarily imply distinct action preferences among critics for the same state. If all critics assign similar relative preferences to candidate actions, their complementary contributions to aggregation may be limited. We therefore conduct an additional experiment to examine whether training with different expectile levels induces meaningful diversity in state-conditional action preferences.

Specifically, for each task, we uniformly sample $1{,}000$ states without replacement from the offline dataset. For each state, we construct a shared set of $64$ candidate actions, consisting of the corresponding dataset action, locally perturbed actions, and actions randomly sampled from the offline dataset. For each critic, we standardize its Q-values across the $64$ candidate actions at each state and apply a softmax with temperature $T=1$ to obtain a discrete action-preference distribution. We measure the disagreement between each pair of critics using the Jensen--Shannon divergence (JSD), normalized by $\log 2$ to lie in $[0,1]$. The overall critic-bank diversity is then computed by averaging the pairwise JSD across all critic pairs at each state, followed by averaging over the $1{,}000$ evaluation states.

To distinguish diversity induced by different expectile levels from variability arising from random initialization and stochastic training, we introduce a same-expectile control bank. The different-expectile bank consists of five critics trained with $\tau\in\{0.5,0.6,0.7,0.8,0.9\}$, whereas the control bank consists of five independently trained critics with random seeds $\{2,4,6,8,10\}$ and a fixed expectile level. Specifically, we set $\tau=0.9$ for AntMaze and $\tau=0.7$ for MuJoCo and Kitchen. Both banks contain five critics and are evaluated using identical states, candidate action sets, normalization procedures, and diversity metrics.

\begin{figure}[h]
    \centering
    \includegraphics[width=0.9\linewidth]{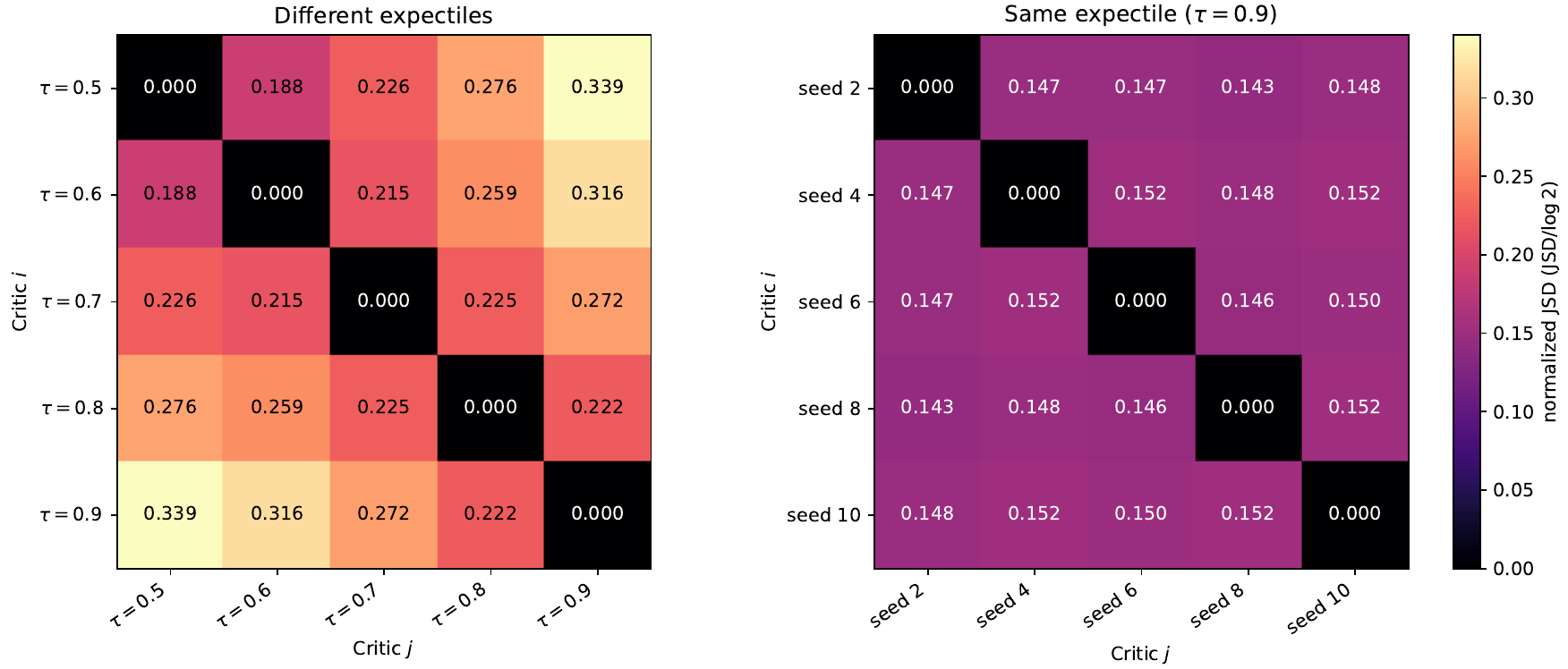}
    \caption{Pairwise action-preference diversity of critic banks on antmaze-large-diverse. The left panel compares five critics trained with different expectile levels, while the right panel compares five independently trained critics with the same expectile level $\tau=0.9$. Each entry reports the Jensen--Shannon divergence (JSD) between the action-preference distributions of two critics.}
    \label{fig:critic_diversity_heatmap}
\end{figure}

Figure~\ref{fig:critic_diversity_heatmap} illustrates the pairwise action-preference divergence on antmaze-large-diverse. Critics trained with similar expectile levels generally exhibit more similar action-preference distributions, while the divergence tends to increase with the separation between their expectile levels. For example, the JSD between the $\tau=0.5$ and $\tau=0.6$ critics is $0.188$, increasing to $0.339$ between the $\tau=0.5$ and $\tau=0.9$ critics. This suggests that different expectile levels induce distinct action preferences, and incorporating multiple expectile regimes can enhance critic-bank diversity. In comparison, the same-expectile control bank also exhibits diversity due to random initialization and stochastic training, but its pairwise divergence is substantially lower and more uniform, with an average JSD of $0.148$ compared with $0.254$ for the different-expectile bank.

Table~\ref{tab:critic_diversity} further reports the average action-preference divergence across all tasks. The different-expectile bank exhibits higher diversity than the same-expectile control on all $18$ tasks, increasing the overall average JSD from $0.102$ to $0.147$. The difference is particularly pronounced on AntMaze, where the average divergence increases from $0.152$ to $0.236$, suggesting that different expectile regimes induce substantially different action preferences in sparse-reward navigation tasks. Meanwhile, critic-bank diversity is also task-dependent. For instance, halfcheetah-medium, halfcheetah-medium-replay, and kitchen-partial exhibit relatively smaller differences between the two banks. Interestingly, these tasks also show relatively limited improvements from IQL-Agg, suggesting that insufficient diversity among candidate critics may partly constrain the benefits of aggregation. When critics exhibit similar action preferences, their complementary contributions may be limited, reducing the potential gains from state-dependent routing. Overall, these findings indicate that varying expectile levels provides additional critic diversity beyond stochastic training variability, while the effectiveness of aggregation may also depend on the degree of diversity available within the critic bank.

\begin{table}[t]
\centering
\caption{
Comparison of critic-bank action-preference diversity across D4RL tasks. Different-Expectile denotes a bank of five critics trained with different expectile levels, while Same-Expectile denotes a control bank of five critics sharing the same expectile level. The higher diversity score in each row is shown in bold.
}
\label{tab:critic_diversity}
\begin{tabular}{lrrr}
\hline
\multicolumn{1}{c}{Dataset}
& \multicolumn{1}{c}{Different-Expectile}
& \multicolumn{1}{c}{Same-Expectile}
& \multicolumn{1}{c}{Difference} \\
\hline
halfcheetah-medium
& \textbf{0.043}
& 0.037
& +0.006 \\

hopper-medium
& \textbf{0.123}
& 0.106
& +0.017 \\

walker2d-medium
& \textbf{0.121}
& 0.052
& +0.069 \\

halfcheetah-medium-replay
& \textbf{0.074}
& 0.068
& +0.006 \\

hopper-medium-replay
& \textbf{0.102}
& 0.072
& +0.030 \\

walker2d-medium-replay
& \textbf{0.111}
& 0.068
& +0.043 \\

halfcheetah-medium-expert
& \textbf{0.070}
& 0.045
& +0.025 \\

hopper-medium-expert
& \textbf{0.112}
& 0.096
& +0.017 \\

walker2d-medium-expert
& \textbf{0.099}
& 0.047
& +0.052 \\
\hline
MuJoCo-Avg
& \textbf{0.095}
& 0.066
& +0.029 \\
\hline

antmaze-umaze
& \textbf{0.194}
& 0.144
& +0.050 \\

antmaze-umaze-diverse
& \textbf{0.250}
& 0.160
& +0.089 \\

antmaze-medium-play
& \textbf{0.233}
& 0.157
& +0.076 \\

antmaze-medium-diverse
& \textbf{0.232}
& 0.152
& +0.080 \\

antmaze-large-play
& \textbf{0.251}
& 0.148
& +0.103 \\

antmaze-large-diverse
& \textbf{0.254}
& 0.148
& +0.105 \\
\hline
AntMaze-Avg
& \textbf{0.236}
& 0.152
& +0.084 \\
\hline

kitchen-complete
& \textbf{0.152}
& 0.136
& +0.016 \\

kitchen-partial
& \textbf{0.118}
& 0.109
& +0.009 \\

kitchen-mixed
& \textbf{0.107}
& 0.093
& +0.015 \\
\hline
Kitchen-Avg
& \textbf{0.126}
& 0.113
& +0.013 \\
\hline
Overall-Avg
& \textbf{0.147}
& 0.102
& +0.045 \\
\hline
\end{tabular}
\end{table}

\subsubsection{Dataset-Averaged Routing Weights}
\begin{figure*}[h]
    \centering
    \includegraphics[width=0.9\linewidth]{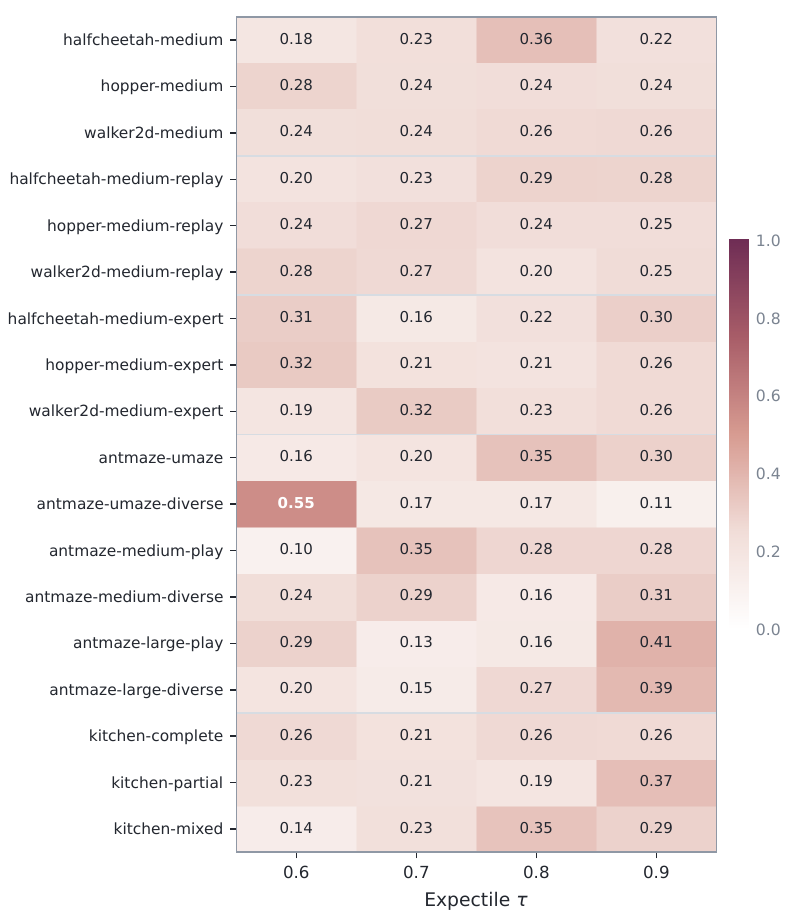}
    \caption{Dataset-averaged routing weights across expectile regimes on D4RL tasks. Each entry reports the mean routing weight assigned to the corresponding expectile critic, $\bar{w}_i = |\mathcal{D}|^{-1}\sum_{s \in \mathcal{D}} w(s,i)$.}
    \label{fig:frequency_heatmap}
\end{figure*}

Having established that critics trained with different expectile levels exhibit distinct action preferences, we further examine how SPA-based routing utilizes this diversity through state-dependent critic weighting. Using the default four-critic bank, we compute the dataset-averaged routing weight assigned to each critic:
\begin{equation}
    \bar{w}_i = \mathbb{E}_{s\sim\mathcal{D}}[w(s,i)],
\end{equation}
where $w(s,i)$ denotes the state-dependent aggregation weight assigned to critic $i$. Figure~\ref{fig:frequency_heatmap} illustrates the resulting average weight distributions across D4RL tasks.

The average routing weights exhibit distinct task-dependent preferences across expectile regimes. In sparse-reward AntMaze tasks, higher-expectile critics receive greater weights on several tasks, consistent with the potential benefits of optimistic value propagation in long-horizon navigation. For example, the $\tau=0.9$ critic receives an average weight of $0.41$ on antmaze-large-play and $0.39$ on antmaze-large-diverse. However, this preference is not universal. On antmaze-umaze-diverse, the lower-expectile critic with $\tau=0.6$ dominates with an average weight of $0.55$, while antmaze-medium-play assigns the largest average weight to $\tau=0.7$. In contrast, MuJoCo tasks generally exhibit more balanced weight distributions, although individual tasks still show moderate preferences for particular expectile regimes. Kitchen tasks similarly exhibit varying preferences, with larger weights assigned to $\tau=0.9$ on kitchen-partial and $\tau=0.8$ on kitchen-mixed.

\subsubsection{Trajectory-Level Effective Expectile}
While the preceding analysis reveals task-dependent average routing preferences, it remains unclear how these preferences evolve during policy execution. To examine the temporal behavior of critic aggregation, we define the effective expectile at each state along a trajectory $\xi=(s_1,\ldots,s_T)$ as
\[
\bar{\tau}(s_t)=\sum_{i\in\mathcal I}w(s_t,i)\tau_i,
\]
where $w(s_t,i)$ is produced by the trained post-hoc SPA module using a fixed critic bank. Here, the effective expectile summarizes routing preferences over the critics' nominal expectile levels; it does not imply that $V_{\mathrm{agg}}(s)$ is the corresponding expectile of
$Q_{\mathrm{agg}}(s,A)$ under $A\sim\mu(\cdot\mid s)$. We track the effective expectile along each episode and report its average over $100$ evaluation rollouts as a function of normalized trajectory progress. This analysis examines whether the routing preferences remain relatively stable or exhibit substantial temporal variation during policy execution.

\begin{figure*}[h]
    \centering
    \includegraphics[width=0.98\linewidth]{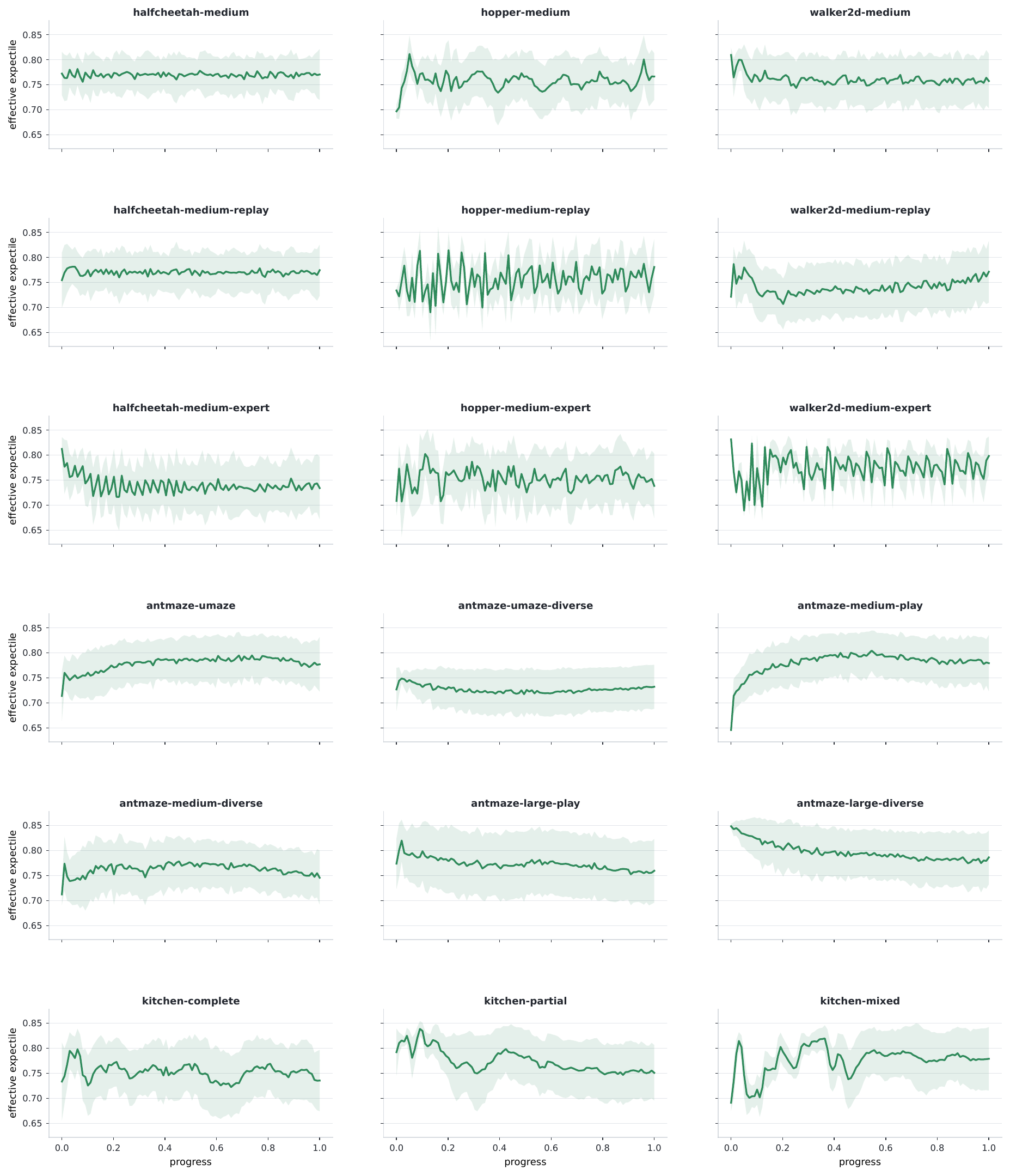}
    \caption{Trajectory-level effective expectile across D4RL tasks. The effective expectile is computed along each episode and averaged over $100$ evaluation rollouts as a function of normalized trajectory progress. Solid curves denote the mean effective expectile, while shaded regions indicate variation across trajectories.}
    \label{fig:trajectory_expectile_curves}
\end{figure*}

Figure~\ref{fig:trajectory_expectile_curves} shows that the average effective expectile generally evolves within relatively narrow, task-dependent ranges. In MuJoCo tasks such as halfcheetah-medium and halfcheetah-medium-replay, the effective expectile remains approximately between $0.7$ and $0.77$, indicating relatively stable routing preferences throughout the episode. Other tasks, including hopper-medium-replay and walker2d-medium-expert, exhibit more noticeable fluctuations, potentially reflecting different critic preferences across phases of cyclic locomotion.

In AntMaze, the effective expectile exhibits more diverse temporal patterns. For instance, antmaze-medium-play and antmaze-umaze gradually shift toward higher expectile levels, suggesting an increasing preference for optimistic critics during later stages of navigation. In contrast, antmaze-umaze-diverse maintains a relatively low effective expectile, while antmaze-large-play and antmaze-large-diverse initially favor higher expectile levels before stabilizing at moderately lower values. These observations further suggest that SPA-based routing adapts to both task characteristics and trajectory progress.

Kitchen tasks exhibit more pronounced temporal variation, particularly on kitchen-partial and kitchen-mixed. This may reflect changing critic preferences across different stages of multi-goal manipulation, where distinct subtasks can require different action preferences. 

Overall, these results suggest that SPA-based routing maintains relatively stable average expectile preferences while adapting to different stages of policy execution. Rather than consistently relying on a fixed critic, the aggregation strategy exhibits task- and trajectory-dependent preferences over expectile regimes. Nevertheless, the averaged curves characterize aggregate temporal behavior and do not necessarily rule out abrupt routing changes within individual trajectories.

\subsubsection{Training Dynamics of Effective Expectiles}

Finally, we investigate the stability of SPA-based routing during joint training of critics and SPA modules, in contrast to the preceding analyses based on fixed, pretrained critic banks. In this setting, both critic estimates and SPA residual statistics evolve throughout optimization, potentially affecting routing preferences. At each training step, we compute the effective expectile $\bar{\tau}(s)$ for each state in the mini-batch and track its batch average alongside the evaluation return.

\begin{figure*}
    \centering
    \includegraphics[width=0.98\linewidth]{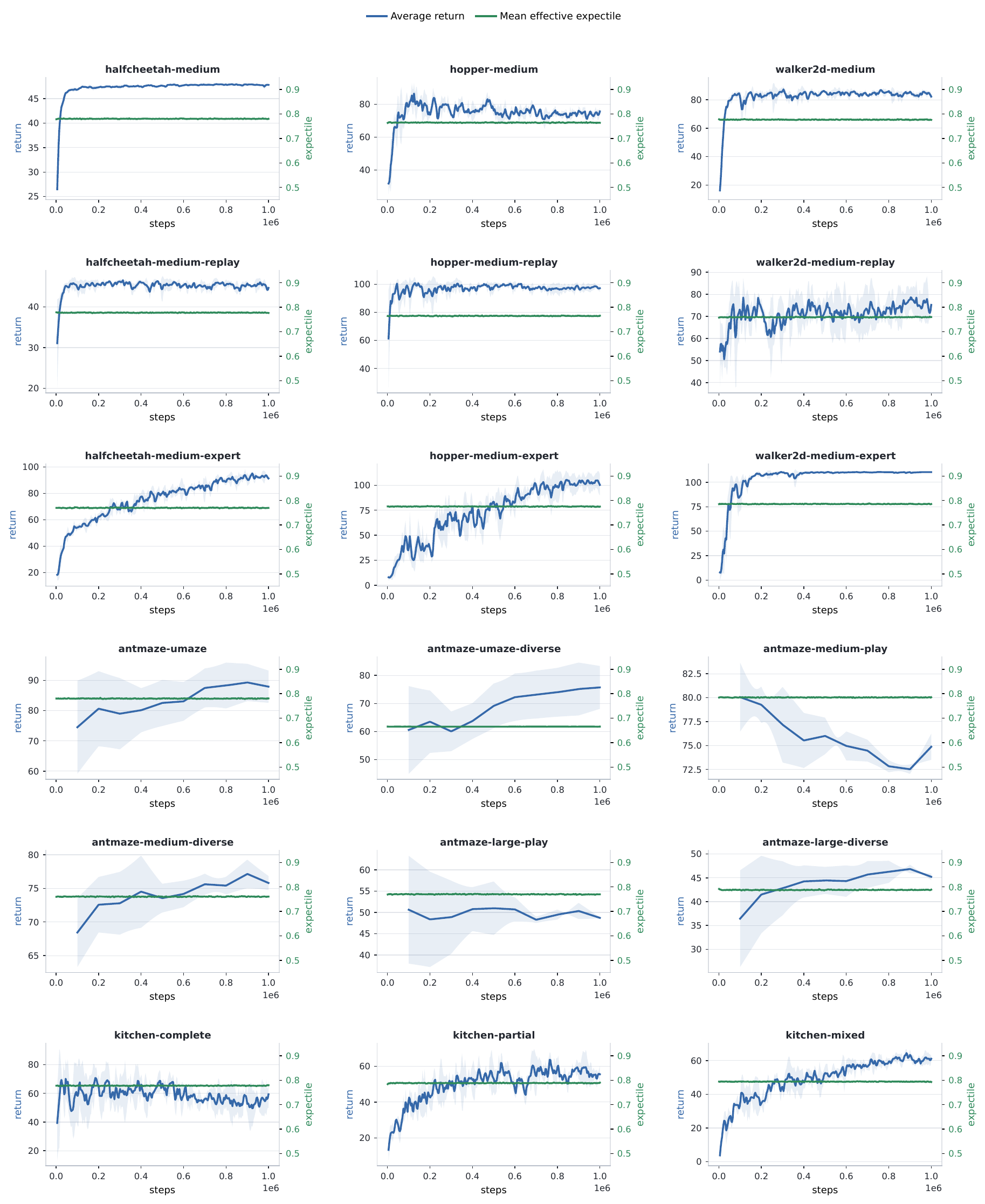}
    \caption{Training dynamics of evaluation return and average effective expectile under joint training. The blue curve denotes the average evaluation return over five seeds, while the green curve denotes the mini-batch average of the routing-induced effective expectile. The effective expectile generally remains within a narrow task-dependent range despite substantial changes in policy performance.}
    \label{fig:return_selected_expectile_dynamics}
\end{figure*}

Figure~\ref{fig:return_selected_expectile_dynamics} illustrates the training dynamics across D4RL tasks. In most MuJoCo environments, evaluation returns improve substantially during training, while the average effective expectile remains within a relatively narrow range. Similar patterns are observed in AntMaze and Kitchen, where evaluation returns exhibit greater fluctuations but the average effective expectile generally shows no large or persistent oscillations. These observations suggest that changes in policy performance do not necessarily correspond to substantial shifts in the overall routing preference.

However, the stability of the average effective expectile does not establish the convergence or stability of individual state-dependent routing weights.

\newpage
\subsection{Sensitivity to Configuration}
\label{app:routing-config}
This section evaluates the sensitivity of the proposed routing mechanism to its configuration choices.
We consider three aspects: the critic-bank configuration, the routing temperature \(\alpha\), and the credibility scale \(\beta\).
All experiments are based on IQL-Agg with the same protocol used in the main experiments.

\subsubsection{Sensitivity to Critic-Bank Configuration}
We first investigate how the composition and size of the expectile critic bank affect aggregation performance. Our default configuration uses $\mathcal{T}=\{0.6,0.7,0.8,0.9\}$, covering a range of relatively conservative to optimistic critics. To examine the effects of critic diversity and bank size, we compare this setting with a reduced bank, $\mathcal{T}=\{0.7,0.9\}$; an extended bank, $\mathcal{T}=\{0.5,0.6,0.7,0.8,0.9,0.95\}$; and a duplicated bank with expectile sequence $(\tau_i)_{i=1}^{12}=(0.6,0.6,0.6,0.7,0.7,0.7,0.8,0.8,0.8,0.9,0.9,0.9)$.

\begin{table*}[h]
\centering
\caption{
Sensitivity to critic-bank configuration.
Scores are reported as mean $\pm$ standard deviation over five seeds.
The highest mean score in each row is shown in bold.
}
\label{tab:sensitivity_critic_group}
\scalebox{1}{
\begin{tabular}{lrrrr}
\hline
\multicolumn{1}{c}{Dataset}
& \multicolumn{1}{c}{Default}
& \multicolumn{1}{c}{Reduced}
& \multicolumn{1}{c}{Extended}
& \multicolumn{1}{c}{Duplicated} \\
\hline

halfcheetah-medium
& \textbf{47.87} $\pm$ 0.22
& 47.33 $\pm$ 0.14
& 47.34 $\pm$ 0.02
& 47.62 $\pm$ 0.22 \\

hopper-medium
& 72.44 $\pm$ 4.62
& 71.17 $\pm$ 0.51
& 69.82 $\pm$ 5.70
& \textbf{75.54} $\pm$ 2.83 \\

walker2d-medium
& \textbf{82.19} $\pm$ 2.35
& 75.21 $\pm$ 1.02
& 79.10 $\pm$ 2.74
& 79.58 $\pm$ 2.95 \\

halfcheetah-medium-replay
& 44.71 $\pm$ 0.52
& \textbf{46.07} $\pm$ 0.17
& 31.18 $\pm$ 14.44
& 45.92 $\pm$ 0.85 \\

hopper-medium-replay
& 98.51 $\pm$ 2.39
& 92.18 $\pm$ 8.25
& 85.55 $\pm$ 7.06
& \textbf{98.73} $\pm$ 3.13 \\

walker2d-medium-replay
& \textbf{76.87} $\pm$ 6.58
& 59.71 $\pm$ 2.39
& 59.72 $\pm$ 13.84
& 73.94 $\pm$ 3.32 \\

halfcheetah-medium-expert
& 94.94 $\pm$ 1.69
& 94.63 $\pm$ 4.28
& 77.79 $\pm$ 27.43
& \textbf{98.35} $\pm$ 2.82 \\

hopper-medium-expert
& 104.77 $\pm$ 6.89
& \textbf{104.82} $\pm$ 10.69
& 95.24 $\pm$ 25.60
& 100.61 $\pm$ 30.08 \\

walker2d-medium-expert
& 110.54 $\pm$ 0.25
& \textbf{110.70} $\pm$ 0.78
& 92.17 $\pm$ 48.76
& 110.34 $\pm$ 0.25 \\

\hline
MuJoCo-Total
& \textbf{732.84}
& 701.82
& 637.89
& 730.63 \\
\hline

antmaze-umaze
& 88.76 $\pm$ 3.41
& \textbf{89.26} $\pm$ 1.56
& 84.24 $\pm$ 5.54
& 88.76 $\pm$ 3.61 \\

antmaze-umaze-diverse
& 73.23 $\pm$ 5.72
& \textbf{76.39} $\pm$ 9.81
& 75.49 $\pm$ 4.26
& 75.04 $\pm$ 3.14 \\

antmaze-medium-play
& 73.54 $\pm$ 3.46
& 75.74 $\pm$ 6.73
& 75.16 $\pm$ 6.11
& \textbf{77.57} $\pm$ 1.71 \\

antmaze-medium-diverse
& 76.59 $\pm$ 4.05
& 75.58 $\pm$ 3.98
& \textbf{78.54} $\pm$ 2.46
& 76.89 $\pm$ 8.86 \\

antmaze-large-play
& 45.56 $\pm$ 4.27
& \textbf{46.29} $\pm$ 7.74
& 34.37 $\pm$ 16.87
& 42.04 $\pm$ 7.37 \\

antmaze-large-diverse
& 46.39 $\pm$ 4.10
& 44.03 $\pm$ 6.65
& \textbf{49.47} $\pm$ 6.87
& 46.40 $\pm$ 4.93 \\

\hline
AntMaze-Total
& 404.08
& \textbf{407.29}
& 397.27
& 406.71 \\
\hline

kitchen-complete
& 63.94 $\pm$ 6.74
& 62.35 $\pm$ 2.93
& 62.49 $\pm$ 2.42
& \textbf{64.05} $\pm$ 5.91 \\

kitchen-partial
& 58.72 $\pm$ 5.30
& \textbf{60.77} $\pm$ 6.98
& 51.17 $\pm$ 2.93
& 59.80 $\pm$ 9.58 \\

kitchen-mixed
& \textbf{61.54} $\pm$ 3.45
& 56.23 $\pm$ 8.95
& 40.72 $\pm$ 18.13
& 58.16 $\pm$ 1.21 \\

\hline
Kitchen-Total
& \textbf{184.20}
& 179.36
& 154.38
& 182.02 \\
\hline

\end{tabular}
}
\end{table*}

As shown in Table~\ref{tab:sensitivity_critic_group}, reducing the critic bank to two expectile levels yields mixed results across domains. The reduced configuration achieves a slightly higher total on AntMaze ($407.29$ vs.\ $404.08$), suggesting that excluding less useful critics may allow aggregation to focus on more favorable expectile regimes. However, performance decreases on MuJoCo ($701.82$ vs.\ $732.84$) and Kitchen ($179.36$ vs.\ $184.20$), indicating that restricted expectile coverage may also limit the complementary critic information available for aggregation. Thus, reducing the critic bank only represents an efficiency-oriented alternative.

In contrast, the extended configuration exhibits substantial performance degradation on MuJoCo ($637.89$ vs.\ $732.84$) and Kitchen ($154.38$ vs.\ $184.20$). Nevertheless, it achieves the highest scores on certain AntMaze tasks, suggesting that more optimistic expectile regimes may benefit some sparse-reward environments. However, the added critics may introduce less reliable value estimates, particularly at the high expectile level $\tau=0.95$, potentially contributing to the observed performance degradation. Although SPA-based routing can suppress the influence of critics exhibiting larger structural discrepancies, biases introduced during critic training may not be fully corrected through post-hoc aggregation. These results highlight the need for caution when expanding the expectile range, particularly when incorporating extreme critics.

The duplicated configuration achieves performance close to the default setting across all three domains, with totals of $730.63$, $406.71$, and $182.02$ on MuJoCo, AntMaze, and Kitchen, respectively. Since the additional critics are independently trained within the same expectile regimes, this configuration increases the number of critic estimates without expanding expectile coverage. Its comparable performance suggests that SPA-based routing can accommodate multiple critics sharing the same expectile level. However, increasing the bank size without introducing additional expectile diversity provides limited performance benefits despite the increased computational cost.

Overall, these results highlight the importance of balancing expectile coverage and critic-bank size. The default four-critic bank offers a favorable trade-off between performance, critic diversity, and computational efficiency, making it a practical general-purpose configuration without requiring task-specific tuning.

\subsubsection{Sensitivity to Routing Temperature $\alpha$}

We next investigate the effect of the routing temperature $\alpha$ in Eq.~\ref{eq:routing_weight} of the main text. This parameter controls the sharpness of the soft routing weights. A smaller $\alpha$ produces more concentrated routing and approaches hard critic selection, whereas a larger $\alpha$ encourages smoother aggregation across critics. This experiment examines the sensitivity of SPA-based routing to different levels of critic selectivity.

\begin{table*}[h]
\centering
\caption{
Sensitivity to routing temperature $\alpha$.
Scores are reported as mean $\pm$ standard deviation over five seeds.
The highest mean score in each row is shown in bold.
}
\label{tab:sensitivity_alpha}
\resizebox{\columnwidth}{!}{
\begin{tabular}{lrrrrr}
\hline
\multicolumn{1}{c}{Dataset}
& \multicolumn{1}{c}{$\alpha=0.25$}
& \multicolumn{1}{c}{$\alpha=0.5$}
& \multicolumn{1}{c}{$\alpha=1.0$}
& \multicolumn{1}{c}{$\alpha=2.0$}
& \multicolumn{1}{c}{$\alpha=5.0$} \\
\hline

halfcheetah-medium
& 47.66 $\pm$ 0.09
& 47.91 $\pm$ 0.26
& 47.87 $\pm$ 0.22
& \textbf{47.96} $\pm$ 0.16
& 47.88 $\pm$ 0.06 \\

hopper-medium
& 67.31 $\pm$ 1.47
& 72.20 $\pm$ 1.91
& 72.44 $\pm$ 4.62
& 71.92 $\pm$ 1.38
& \textbf{73.98} $\pm$ 1.83 \\

walker2d-medium
& 71.82 $\pm$ 9.45
& 76.86 $\pm$ 1.08
& 82.19 $\pm$ 2.35
& 80.87 $\pm$ 2.05
& \textbf{84.26} $\pm$ 0.89 \\

halfcheetah-medium-replay
& 42.84 $\pm$ 1.10
& 44.33 $\pm$ 2.34
& 44.71 $\pm$ 0.52
& 44.69 $\pm$ 1.09
& \textbf{45.70} $\pm$ 0.18 \\

hopper-medium-replay
& 98.09 $\pm$ 5.80
& 96.06 $\pm$ 6.00
& \textbf{98.51} $\pm$ 2.39
& 97.58 $\pm$ 3.40
& 96.04 $\pm$ 2.53 \\

walker2d-medium-replay
& 70.95 $\pm$ 9.11
& 73.68 $\pm$ 12.80
& \textbf{76.87} $\pm$ 6.58
& 73.50 $\pm$ 7.98
& 76.73 $\pm$ 5.65 \\

halfcheetah-medium-expert
& \textbf{95.39} $\pm$ 7.82
& 93.90 $\pm$ 2.80
& 94.94 $\pm$ 1.69
& 93.83 $\pm$ 2.90
& 85.56 $\pm$ 5.23 \\

hopper-medium-expert
& 106.80 $\pm$ 25.01
& \textbf{109.36} $\pm$ 16.69
& 104.77 $\pm$ 6.89
& 102.19 $\pm$ 25.09
& 95.80 $\pm$ 20.56 \\

walker2d-medium-expert
& 110.77 $\pm$ 0.41
& \textbf{110.80} $\pm$ 0.75
& 110.54 $\pm$ 0.25
& 110.28 $\pm$ 0.19
& 110.41 $\pm$ 0.13 \\

\hline
MuJoCo-Total
& 711.65
& 725.10
& \textbf{732.84}
& 722.84
& 716.36 \\
\hline

antmaze-umaze
& \textbf{91.76} $\pm$ 1.76
& 90.34 $\pm$ 2.23
& 88.76 $\pm$ 3.41
& 90.26 $\pm$ 0.38
& 85.89 $\pm$ 6.78 \\

antmaze-umaze-diverse
& 70.53 $\pm$ 5.23
& 72.50 $\pm$ 2.54
& 73.23 $\pm$ 5.72
& \textbf{75.45} $\pm$ 7.03
& 67.18 $\pm$ 2.62 \\

antmaze-medium-play
& 66.55 $\pm$ 5.03
& 70.20 $\pm$ 4.99
& \textbf{73.54} $\pm$ 3.46
& 71.26 $\pm$ 4.84
& 73.48 $\pm$ 5.00 \\

antmaze-medium-diverse
& 70.24 $\pm$ 0.16
& 74.34 $\pm$ 4.60
& \textbf{76.59} $\pm$ 4.05
& 75.54 $\pm$ 5.14
& 72.58 $\pm$ 1.46 \\

antmaze-large-play
& 43.12 $\pm$ 1.65
& 46.26 $\pm$ 2.18
& 45.56 $\pm$ 4.27
& 48.41 $\pm$ 4.63
& \textbf{49.13} $\pm$ 6.79 \\

antmaze-large-diverse
& 43.22 $\pm$ 5.94
& 45.13 $\pm$ 0.92
& 46.39 $\pm$ 4.10
& 46.06 $\pm$ 1.32
& \textbf{48.72} $\pm$ 2.51 \\

\hline
AntMaze-Total
& 385.41
& 398.77
& 404.08
& \textbf{406.98}
& 396.97 \\
\hline

kitchen-complete
& 57.84 $\pm$ 16.64
& 63.99 $\pm$ 8.46
& 63.94 $\pm$ 6.74
& \textbf{64.37} $\pm$ 7.65
& 54.82 $\pm$ 18.70 \\

kitchen-partial
& 58.57 $\pm$ 9.23
& 56.70 $\pm$ 5.35
& 58.72 $\pm$ 5.30
& \textbf{61.42} $\pm$ 1.80
& 59.82 $\pm$ 7.79 \\

kitchen-mixed
& 61.22 $\pm$ 1.31
& \textbf{62.79} $\pm$ 2.67
& 61.54 $\pm$ 3.45
& 60.45 $\pm$ 5.57
& 61.25 $\pm$ 3.21 \\

\hline
Kitchen-Total
& 177.62
& 183.49
& 184.20
& \textbf{186.24}
& 175.90 \\
\hline

\end{tabular}
}
\end{table*}

As shown in Table~\ref{tab:sensitivity_alpha}, SPA-based aggregation maintains competitive performance across moderate routing temperatures. The default setting $\alpha=1.0$ achieves the highest MuJoCo total ($732.84$), while $\alpha=2.0$ yields slightly higher totals on AntMaze ($406.98$ vs.\ $404.08$) and Kitchen ($186.24$ vs.\ $184.20$). These results suggest that the proposed routing mechanism does not require precise temperature tuning to achieve strong performance across domains.

At the task level, the preferred routing temperature varies across environments. Sharper routing with $\alpha=0.25$ substantially reduces the domain totals, particularly on AntMaze ($385.41$) and MuJoCo ($711.65$), suggesting that overly concentrated weights may increase sensitivity to individual critic estimates. Conversely, smoother routing with $\alpha=5.0$ achieves the highest scores on several tasks, including AntMaze-large-play and AntMaze-large-diverse, but reduces the overall MuJoCo and Kitchen totals. These results indicate that neither highly concentrated nor excessively smooth routing consistently benefits aggregation, as the appropriate degree of critic selectivity varies across tasks.

Overall, moderate temperatures within $\alpha\in[0.5,2.0]$ provide relatively stable domain-level performance, reflecting a balance between selective critic weighting and the smoothing effects of aggregation. We therefore retain $\alpha=1.0$ as the default configuration, which achieves competitive performance across all three domains without requiring task-specific temperature tuning.

\subsubsection{Sensitivity to Credibility Scale $\beta$}

Finally, we investigate the effect of the credibility scale $\beta$ in Eq.~\ref{eq:credited-AS} of the main text. This parameter controls the strength of the SDS-based credibility penalty. A smaller $\beta$ imposes a stronger penalty on critics with larger self-consistency deviations, whereas a larger $\beta$ weakens the penalty and makes the routing score closer to AS alone. This experiment examines whether the proposed aggregation mechanism requires careful tuning of structural credibility weighting.

\begin{table*}[h]
\centering
\caption{
Sensitivity to credibility scale $\beta$.
Scores are reported as mean $\pm$ standard deviation over five seeds.
The highest mean score in each row is shown in bold.
}
\label{tab:sensitivity_beta}
\resizebox{\columnwidth}{!}{
\begin{tabular}{lrrrrr}
\hline
\multicolumn{1}{c}{Dataset}
& \multicolumn{1}{c}{$\beta=0.25$}
& \multicolumn{1}{c}{$\beta=0.5$}
& \multicolumn{1}{c}{$\beta=1.0$}
& \multicolumn{1}{c}{$\beta=2.0$}
& \multicolumn{1}{c}{$\beta=5.0$} \\
\hline

halfcheetah-medium
& 47.82 $\pm$ 0.11
& 47.76 $\pm$ 0.17
& \textbf{47.87} $\pm$ 0.22
& 47.82 $\pm$ 0.08
& 47.21 $\pm$ 0.83 \\

hopper-medium
& 71.30 $\pm$ 5.36
& 73.04 $\pm$ 2.56
& 72.44 $\pm$ 4.62
& 74.93 $\pm$ 3.07
& \textbf{75.07} $\pm$ 6.62 \\

walker2d-medium
& \textbf{83.77} $\pm$ 2.33
& 81.74 $\pm$ 2.81
& 82.19 $\pm$ 2.35
& 80.31 $\pm$ 5.00
& 81.91 $\pm$ 4.25 \\

halfcheetah-medium-replay
& \textbf{45.20} $\pm$ 0.21
& 43.85 $\pm$ 1.26
& 44.71 $\pm$ 0.52
& 44.28 $\pm$ 1.11
& 43.71 $\pm$ 1.31 \\

hopper-medium-replay
& 94.68 $\pm$ 3.37
& 95.66 $\pm$ 4.53
& 98.51 $\pm$ 2.39
& 98.10 $\pm$ 5.05
& \textbf{99.61} $\pm$ 1.62 \\

walker2d-medium-replay
& 71.23 $\pm$ 11.24
& 74.36 $\pm$ 8.89
& \textbf{76.87} $\pm$ 6.58
& 76.11 $\pm$ 9.27
& 73.45 $\pm$ 11.82 \\

halfcheetah-medium-expert
& 94.16 $\pm$ 1.41
& \textbf{97.16} $\pm$ 4.01
& 94.94 $\pm$ 1.69
& 93.17 $\pm$ 4.53
& 96.59 $\pm$ 1.88 \\

hopper-medium-expert
& 106.48 $\pm$ 6.32
& 102.50 $\pm$ 12.35
& 104.77 $\pm$ 6.89
& \textbf{109.70} $\pm$ 5.69
& 94.94 $\pm$ 24.56 \\

walker2d-medium-expert
& 110.16 $\pm$ 0.48
& \textbf{110.54} $\pm$ 0.19
& \textbf{110.54} $\pm$ 0.25
& 109.93 $\pm$ 1.07
& 110.15 $\pm$ 0.35 \\

\hline
MuJoCo-Total
& 724.79
& 726.61
& 732.84
& \textbf{734.34}
& 722.62 \\
\hline

antmaze-umaze
& \textbf{92.79} $\pm$ 3.78
& 91.51 $\pm$ 5.66
& 88.76 $\pm$ 3.41
& 86.72 $\pm$ 1.95
& 88.94 $\pm$ 3.09 \\

antmaze-umaze-diverse
& 71.17 $\pm$ 4.82
& 71.70 $\pm$ 1.63
& \textbf{73.23} $\pm$ 5.72
& 71.15 $\pm$ 3.09
& 68.32 $\pm$ 7.58 \\

antmaze-medium-play
& 71.98 $\pm$ 3.58
& 72.41 $\pm$ 5.10
& \textbf{73.54} $\pm$ 3.46
& 66.30 $\pm$ 4.23
& 70.54 $\pm$ 3.63 \\

antmaze-medium-diverse
& 74.40 $\pm$ 4.97
& 73.82 $\pm$ 2.60
& \textbf{76.59} $\pm$ 4.05
& 74.63 $\pm$ 2.76
& 70.39 $\pm$ 1.91 \\

antmaze-large-play
& 47.42 $\pm$ 2.70
& 48.84 $\pm$ 4.46
& 45.56 $\pm$ 4.27
& \textbf{51.55} $\pm$ 2.08
& 48.97 $\pm$ 4.22 \\

antmaze-large-diverse
& 43.66 $\pm$ 2.83
& \textbf{47.47} $\pm$ 2.00
& 46.39 $\pm$ 4.10
& 47.42 $\pm$ 5.33
& 47.40 $\pm$ 3.20 \\

\hline
AntMaze-Total
& 401.43
& \textbf{405.74}
& 404.08
& 397.76
& 394.56 \\
\hline

kitchen-complete
& 62.45 $\pm$ 5.05
& 58.30 $\pm$ 6.73
& 63.94 $\pm$ 6.74
& \textbf{64.65} $\pm$ 8.98
& 62.43 $\pm$ 9.53 \\

kitchen-partial
& \textbf{64.08} $\pm$ 6.36
& 57.26 $\pm$ 1.70
& 58.72 $\pm$ 5.30
& 58.17 $\pm$ 2.32
& 60.35 $\pm$ 2.99 \\

kitchen-mixed
& 59.62 $\pm$ 1.57
& \textbf{64.78} $\pm$ 7.45
& 61.54 $\pm$ 3.45
& 60.48 $\pm$ 5.02
& 62.16 $\pm$ 3.03 \\

\hline
Kitchen-Total
& \textbf{186.15}
& 180.34
& 184.20
& 183.31
& 184.94 \\
\hline

\end{tabular}
}
\end{table*}

As shown in Table~\ref{tab:sensitivity_beta}, SPA-based aggregation maintains competitive performance across a range of credibility scales, although the preferred penalty strength varies across domains. Specifically, $\beta=2.0$ achieves the highest MuJoCo total ($734.34$), $\beta=0.5$ performs best on AntMaze ($405.74$), and $\beta=0.25$ yields the highest Kitchen total ($186.15$). Nevertheless, the default setting $\beta=1.0$ remains close to the best domain-specific configurations, suggesting that competitive performance can be achieved without task-specific tuning.

At the task level, the effects of SDS-based credibility weighting are non-monotonic. Stronger penalties with $\beta=0.25$ benefit certain tasks, particularly on Kitchen, whereas weakening the penalty with $\beta=5.0$ leads to a noticeable performance decline on AntMaze. These results suggest that the appropriate strength of credibility weighting depends on the task, with neither stronger nor weaker penalties consistently dominating. Overall, moderate credibility scales generally provide a reasonable balance between structural credibility and critic selectivity. We therefore adopt $\beta=1.0$ as a robust default across all three domains.

\subsubsection{Post-hoc versus Joint SPA Training}
\label{app:joint_training}

We further investigate whether the auxiliary SPA module can be trained jointly with the critics, rather than relying on a separate post-hoc training stage. In the default post-hoc setting, the Q- and V-functions are first trained using the original IQL objectives and subsequently frozen, after which the SPA module is optimized to estimate the three conditional MAR statistics from fixed residual targets. In contrast, joint training initializes the SPA module alongside the critics and optimizes it throughout critic training. Specifically, the MAR regression targets are continuously updated according to the evolving residuals $e_t(s,a)=Q_{\theta_t}(s,a)-V_{\psi_t}(s)$, while the original critic-learning objectives and SPA estimation losses are retained. This eliminates the separate SPA training stage but introduces non-stationary residual targets during optimization. We compare the two strategies using the same critic-bank configuration and downstream policy extraction protocol.

\begin{table*}[h]
\centering
\caption{
Comparison of post-hoc and joint SPA training.
Scores are reported as mean $\pm$ standard deviation over five seeds.
The highest mean score in each row is shown in bold.
}
\label{tab:spa_training_comparison}
{
\begin{tabular}{lrr}
\hline
\multicolumn{1}{c}{Dataset}
& \multicolumn{1}{c}{Post-hoc}
& \multicolumn{1}{c}{Joint} \\
\hline

halfcheetah-medium
& \textbf{47.87} $\pm$ 0.22
& 47.81 $\pm$ 0.14 \\

hopper-medium
& 72.44 $\pm$ 4.62
& \textbf{75.59} $\pm$ 3.16 \\

walker2d-medium
& 82.19 $\pm$ 2.35
& \textbf{82.27} $\pm$ 2.71 \\

halfcheetah-medium-replay
& \textbf{44.71} $\pm$ 0.52
& \textbf{44.71} $\pm$ 0.85 \\

hopper-medium-replay
& \textbf{98.51} $\pm$ 2.39
& 97.12 $\pm$ 3.03 \\

walker2d-medium-replay
& \textbf{76.87} $\pm$ 6.58
& 75.22 $\pm$ 9.95 \\

halfcheetah-medium-expert
& \textbf{94.94} $\pm$ 1.69
& 91.28 $\pm$ 4.78 \\

hopper-medium-expert
& \textbf{104.77} $\pm$ 6.89
& 100.34 $\pm$ 16.32 \\

walker2d-medium-expert
& \textbf{110.54} $\pm$ 0.25
& 110.38 $\pm$ 0.23 \\

\hline
MuJoCo-Total
& \textbf{732.84}
& 724.74 \\
\hline

antmaze-umaze
& \textbf{88.76} $\pm$ 3.41
& 87.93 $\pm$ 6.72 \\

antmaze-umaze-diverse
& 73.23 $\pm$ 5.72
& \textbf{75.72} $\pm$ 11.11 \\

antmaze-medium-play
& 73.54 $\pm$ 3.46
& \textbf{74.87} $\pm$ 2.86 \\

antmaze-medium-diverse
& \textbf{76.59} $\pm$ 4.05
& 75.80 $\pm$ 2.02 \\

antmaze-large-play
& 45.56 $\pm$ 4.27
& \textbf{48.73} $\pm$ 2.71 \\

antmaze-large-diverse
& \textbf{46.39} $\pm$ 4.10
& 45.22 $\pm$ 4.58 \\

\hline
AntMaze-Total
& 404.08
& \textbf{408.28} \\
\hline

kitchen-complete
& \textbf{63.94} $\pm$ 6.74
& 59.34 $\pm$ 11.83 \\

kitchen-partial
& \textbf{58.72} $\pm$ 5.30
& 55.53 $\pm$ 7.15 \\

kitchen-mixed
& \textbf{61.54} $\pm$ 3.45
& 61.09 $\pm$ 2.66 \\

\hline
Kitchen-Total
& \textbf{184.20}
& 175.95 \\
\hline

\end{tabular}
}
\end{table*}

As shown in Table~\ref{tab:spa_training_comparison}, joint training achieves performance comparable to post-hoc training, with a slightly higher domain total on AntMaze ($408.28$ vs.\ $404.08$). The improvements suggest that jointly learned residual statistics can provide effective routing signals even as the critics evolve. However, joint training underperforms on MuJoCo ($724.74$ vs.\ $732.84$) and Kitchen ($175.95$ vs.\ $184.20$).

One possible explanation is the non-stationarity of MAR regression targets during joint optimization. Since the SPA module estimates conditional residual statistics derived from the current Q- and V-functions, changes in the critics continuously alter the target residual distributions. This may introduce additional optimization variability, particularly during early critic training. In contrast, post-hoc training operates on fixed critic representations and residual targets, providing a more stationary estimation problem. Nevertheless, the competitive performance of joint training, particularly on AntMaze, suggests that SPA estimation does not necessarily require fully converged critics.

Overall, these results demonstrate the flexibility of SPA training. Joint optimization provides a practical alternative when training critics from initialization, avoiding a separate post-hoc estimation stage. However, post-hoc training remains the preferred configuration due to its more consistent performance and its independence from the evolving critic-training process.

\end{document}